\documentclass{article}

\usepackage[T1]{fontenc}
\usepackage{microtype}
\usepackage{graphicx}
\usepackage{subcaption}
\usepackage{booktabs}
\usepackage{longtable}
\usepackage{xurl}
\usepackage{hyperref}

\usepackage[preprint]{icml2026}
\hypersetup{pdfsubject={Machine unlearning}}

\usepackage{amsmath}
\usepackage{amssymb}
\usepackage{mathtools}
\usepackage{amsthm}
\usepackage{aliascnt}
\usepackage[capitalize,noabbrev]{cleveref}
\usepackage{xcolor}

\theoremstyle{plain}
\newtheorem{theorem}{Theorem}
\newaliascnt{proposition}{theorem}
\newtheorem{proposition}[proposition]{Proposition}
\aliascntresetthe{proposition}
\newaliascnt{lemma}{theorem}
\newtheorem{lemma}[lemma]{Lemma}
\aliascntresetthe{lemma}
\newaliascnt{corollary}{theorem}
\newtheorem{corollary}[corollary]{Corollary}
\aliascntresetthe{corollary}
\crefname{proposition}{Proposition}{Propositions}
\crefname{lemma}{Lemma}{Lemmas}
\crefname{corollary}{Corollary}{Corollaries}
\newtheorem{innercustomthm}{Theorem}
\newenvironment{customthm}[1]{\renewcommand\theinnercustomthm{#1}\innercustomthm}{\endinnercustomthm}
\newtheorem{innercustomprop}{Proposition}
\newenvironment{customprop}[1]{\renewcommand\theinnercustomprop{#1}\innercustomprop}{\endinnercustomprop}
\crefname{innercustomthm}{Theorem}{Theorems}
\crefname{innercustomprop}{Proposition}{Propositions}
\theoremstyle{definition}
\newtheorem{definition}{Definition}
\theoremstyle{remark}
\newtheorem{remark}{Remark}

\newcommand{\R}{\mathbb{R}}
\newcommand{\E}{\mathbb{E}}
\newcommand{\tr}{\operatorname{tr}}
\newcommand{\Write}{\textsc{Write}}
\newcommand{\Fit}{\textsc{Fit}}
\newcommand{\Select}{\textsc{Select}}
\newcommand{\Augment}{\textsc{Augment}}
\DeclareMathOperator*{\argmax}{arg\,max}

\icmltitlerunning{Forgetting by Rewriting}

\begin{document}

\twocolumn[
\icmltitle{Forgetting by Rewriting: Exact Unlearning for Written Networks,\\and Why Output Scores Cannot Prove Forgetting}

\begin{icmlauthorlist}
\icmlauthor{Soumik Bhattacharjee}{buet}
\icmlauthor{Musa Tur Farazi}{buet}
\icmlauthor{Md Nasir Uddin}{buet}
\icmlauthor{Tishna Sabrina}{uap}
\icmlauthor{Chowdhury Mohammad Rakin Haider}{buet}
\end{icmlauthorlist}

\icmlaffiliation{buet}{Department of Computer Science and Engineering, Bangladesh University of Engineering and Technology (BUET), Dhaka, Bangladesh}
\icmlaffiliation{uap}{Department of Electrical and Electronic Engineering, University of Asia Pacific (UAP), Dhaka, Bangladesh}
\icmlcorrespondingauthor{Soumik Bhattacharjee}{2005031@cse.buet.ac.bd, soumikbhat330@gmail.com}

\icmlkeywords{machine unlearning, exact deletion, closed-form models, self-training, model selection, membership inference, linear probes}

\vskip 0.3in
]

\printAffiliationsAndNotice{}

\begin{abstract}
When a person asks for their data to be removed from a machine learning model, the safest answer is to retrain the model without that data. Retraining is expensive, so most methods take a shortcut, and it is hard to tell whether the shortcut really removed the data. We look at this problem from two sides. First, we build models whose weights are \emph{written} rather than trained: fixed features designed before any data are seen, followed by steps that each have an exact formula. For these models, a cheap deletion gives exactly the model that retraining would give, even when the pipeline also tunes a hyperparameter and labels unlabelled data for itself. In 2,400 deletion requests on three datasets, the cheap deletion matched retraining every time, up to rounding error. Exact deletion is not privacy, though: comparing the model before and after a deletion reveals which records were deleted (AUC $0.89$--$0.98$). Second, for ordinary networks trained by gradient descent, we prove that no test of the model's outputs can show that the data are gone, so we compare internal layers with retrained models instead. A backdoor case study shows why this matters. With only 10\% of the poisoned images known, a one-layer edit lowers the attack success rate on CIFAR-10 ResNet-18 to $1.5$--$7.8\%$, while retraining on the remaining data keeps it at $76$--$100\%$. Yet the trigger is still stored in earlier layers and returns after one to ten training steps.
\end{abstract}

\section{Introduction}\label{sec:intro}

People can ask a company to delete their data, and the request should also reach the models trained on that data. The clearest way to honour it is to retrain the model from scratch without the deleted records. This defines what a correct deletion is: a deletion is \emph{exact} if it gives the same model that retraining would give \citep{ginart2019making,thudi2022on}. Retraining a large network for every request is too slow and too costly, so most \emph{unlearning} methods take a shortcut, such as a few gradient steps that push the model away from the deleted data \citep{nguyen2025survey}.

\begin{figure*}[t]
\centering
\includegraphics[width=\textwidth]{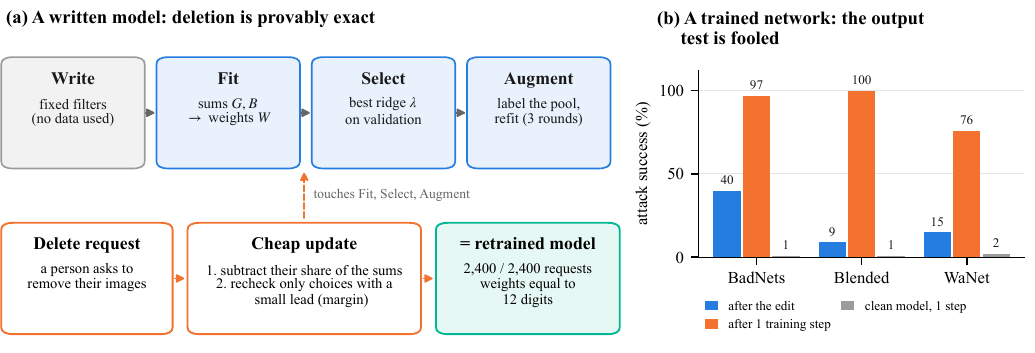}
\caption{The paper in one picture. \textbf{(a)} A written model is built in four steps, and only the last three use data. To delete someone's images, we subtract their share of the stored sums and recheck only the few choices that could change. The result is the same model that retraining would give, in all 2,400 requests we tested. \textbf{(b)} A trained ResNet-18 with a hidden backdoor. An edit of its last layer makes the attack look weak at the output (blue). But one training step on 120 clean and 8 poisoned images brings it back (orange), while a clean model stays near zero (grey). Output tests cannot tell ``removed'' from ``hidden''. (Last-layer edit, 10\% of poisoned images known, validation attack success, mean of 5 seeds.)}
\label{fig:overview}
\end{figure*}

A shortcut raises a hard question: how do we know it worked? The usual tests look only at the model's outputs \citep{hayes2025inexact,triantafillou2024are}. We ask two questions.
\begin{enumerate}
\item Can a cheap deletion be \emph{provably} the same as retraining?
\item Can output tests show that data are really gone from a trained network?
\end{enumerate}
The answers are yes and no (\cref{fig:overview}), and together they give the main lesson of the paper: \emph{forgetting can be proved by how a model is computed, not by what it outputs}.

\paragraph{Yes, for written models.} Our models have \emph{written} weights: fixed hand-designed features, then steps with exact formulas (\cref{fig:overview}a). The fitting step depends on the data only through sums, so deleting is subtracting. This part is known \citep{cao2015towards,guo2020certified}. The new part is model selection and self-training, which make discrete choices that a deletion could flip. We find every choice that could flip and recheck only those.

\paragraph{No, for trained networks.} Deleting a class's output unit passes every output test, yet changes no hidden layer. So we compare hidden layers with retrained models, using small linear \emph{probes}. A backdoor study shows why this matters (\cref{fig:overview}b).

\paragraph{Contributions.}
\begin{enumerate}
\item \textbf{Exact deletion through a whole written pipeline} (\cref{thm:pipeline-inf}), matching retraining in all 2,400 requests we tested. On hand-written features this covers every weight, at low accuracy ($49\%$ on CIFAR-10); on a frozen backbone ($98\%$) it covers the downstream data only.
\item \textbf{Exact is not private} (\cref{sec:x3}): comparing the model before and after a deletion reveals who was deleted (AUC $0.89$--$0.98$).
\item \textbf{Outputs cannot prove forgetting} (\cref{prop:padding-inf}), and a guarantee for a one-layer edit at its own layer (\cref{thm:probe-inf}).
\item \textbf{A backdoor case study with predictions written in advance} (\cref{sec:x7}; 70 of 131 predictions held, \cref{app:prereg}).
\end{enumerate}

\section{Background: what deletion must achieve, and the models we study}\label{sec:setting}

\paragraph{What counts as a correct deletion.} A training algorithm $A$ turns a dataset $D$ into a model $A(D)$. A person asks us to delete a set of records $F\subseteq D$. A deletion method $R$ takes the current model, the request $F$ and some stored information $T(D)$, and returns a new model. We use three labels, and every method in the paper carries one of them (\cref{tab:labels}).
\begin{itemize}
\item \emph{Exact}: $R$ returns the model that retraining on $D\setminus F$ would return \citep[Def.~3.1]{ginart2019making}. When training is random, the two models must have the same distribution; we then say \emph{exact in distribution}.
\item \emph{Certified}: the two models are close in a statistical sense, as in differential privacy \citep{guo2020certified}.
\item \emph{Heuristic}: no guarantee. Most unlearning methods for deep networks, and the layer edit we study in \cref{sec:eval}, are heuristic.
\end{itemize}
An exact method that only reads the request, the model and the stored information is the strongest kind of unlearning \citep{sekhari2021remember}. The practical question is then how much must be stored.

\paragraph{The written pipeline.} Our models are built in four stages (\cref{tab:stages}).
\begin{enumerate}
\item \textbf{Write.} A fixed feature map $\phi$ turns each input into a vector $\phi(x)\in\R^d$. It is designed before any data are seen. Our main example is a bank of 544 hand-designed edge and colour filters on CIFAR-10, followed by pooling and a fixed random projection to $d=1{,}024$ numbers \citep{rahimi2007random}. A weaker variant uses a frozen public backbone such as DINOv2.
\item \textbf{Fit.} A linear read-out $W$ maps features to class scores. It is found by ridge regression on a labelled set $L$ with one-hot labels $y_z$:
\begin{align*}
G_L&=\textstyle\sum_{z\in L}\phi_z\phi_z^\top,\qquad B_L=\textstyle\sum_{z\in L}\phi_zy_z^\top,\\
W_\lambda&=(G_L+\lambda I)^{-1}B_L .
\end{align*}
The data enter only through the two sums $G_L$ and $B_L$.
\item \textbf{Select.} The ridge strength $\lambda$ is chosen from a fixed list $H$ by its score on a validation set $V$. The score is validation accuracy or squared validation error.
\item \textbf{Augment.} The model improves itself on an unlabelled pool $U$ by self-training \citep{scudder1965probability}. In each of $k$ rounds, every pool image gets the label the current model scores highest (a \emph{pseudo-label}). The pool images are then added to the sums with weight $w$, and the read-out is solved again.
\end{enumerate}
The training data are $D=L\cup U\cup V$, and a request may delete points from any of the three sets. We never test on the pool: $U$ is part of the training split.

\begin{figure}[t]
\centering
\includegraphics[width=\columnwidth]{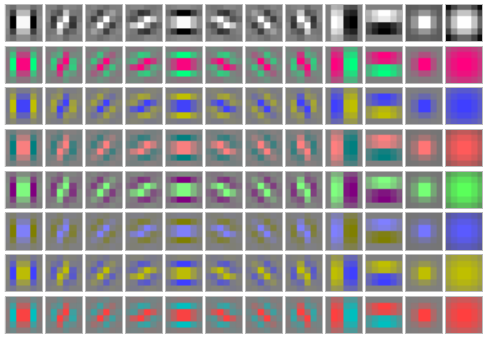}
\caption{Some of the 544 hand-written filters of our written CIFAR-10 model: edges at eight angles, two sizes and two colour blobs (columns), each in eight colour mixes (rows). They are fixed before any image is seen, so no deletion can affect them.}
\label{fig:filters}
\end{figure}

\begin{table}[t]
\centering\footnotesize
\setlength{\tabcolsep}{3pt}
\begin{tabular}{@{}p{1.2cm}p{2.9cm}p{3.6cm}@{}}
\toprule
stage & what it computes & what a deletion must do\\\midrule
Write & fixed features $\phi(x)$ & nothing: no data were used\\
Fit & sums $G,B$; weights $W$ & subtract the deleted points from the sums and solve again\\
Select & best $\lambda$ on validation & find out whether the winner changes\\
Augment & pseudo-labels, then refit & find out which pseudo-labels change, round by round\\
\bottomrule
\end{tabular}
\caption{The four stages of a written pipeline. Fit is easy to undo because it depends on the data only through sums. Select and Augment are harder because they make discrete choices that a small change of the weights can flip.}\label{tab:stages}
\end{table}

\paragraph{Why selection and self-training are the hard part.} Each of these stages picks the highest of several scores: the best $\lambda$, or the top class for a pool image. After a deletion the weights move a little, and a choice can flip if its lead over the runner-up is small. If a pseudo-label flips, the next round fits different targets, and the change can spread. The simplest exact method is to rerun the whole pipeline, which is what we want to avoid. \Cref{sec:exact} shows how to find every choice that might flip, and recheck only those.

\paragraph{The one thing that must not depend on data.} The guarantee covers a weight only if every input to it can be deleted or never saw the data. Our hand-written features use no training image at all. Even the one scale constant that sets their size is computed from random-noise images (\cref{app:exp}). A frozen backbone was trained on other data, so for it exactness covers the downstream data only. Features built from training images, such as cluster centres or patches copied from the data, need a different rule, which we discuss in \cref{sec:exact}.

\section{Exact deletion through the whole pipeline}\label{sec:exact}

This section answers our first question. We first give the idea in two steps, then the result, then what it costs.

\paragraph{Step 1: a deletion moves the weights by a known amount.} Every weight matrix in the pipeline solves a linear system $AW=B$, where $A$ and $B$ are sums over the data. Deleting the points in $F$ removes their terms from both sums. The new weights $W'$ then differ from the old ones by
\begin{equation}\label{eq:smw-main}
W'-W=A'^{-1}\textstyle\sum_{z\in F}w_z\,\phi_z\bigl(\phi_z^\top W-t_z^\top\bigr),
\end{equation}
where $A'$ is the new system matrix, $t_z$ is the target of point $z$ and $w_z$ its weight in the fit. This is the classical Sherman--Morrison--Woodbury formula (\cref{lem:smw}). It tells us two useful things. The change can be computed from the deleted points alone. Its size is small when the deleted points are few and already well fitted.

\begin{figure}[t]
\centering
\includegraphics[width=\columnwidth]{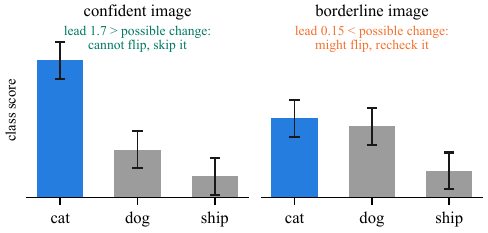}
\caption{The margin rule (illustration). After a deletion, every class score can move by at most a known amount (whiskers). An image whose top class leads by more than that cannot change its label, so we skip it. Only borderline images are rechecked.}\label{fig:margin}
\end{figure}

\paragraph{Step 2: a decision can only flip if its lead is small (\cref{fig:margin}).} Each discrete choice in the pipeline picks the highest of several scores. We call the gap between the top score and the runner-up the \emph{margin}, divided by the length of the feature vector. If the weights move by $\Delta$ (measured by the largest factor by which $\Delta$ can stretch a vector), every score moves by at most $\|\phi\|\cdot\|\Delta\|$, and a gap between two scores by at most $\sqrt2$ times that. So a choice whose margin is larger than $\sqrt2\|\Delta\|$ cannot flip. Only the points with a small margin need to be checked again.

\paragraph{The result.} Putting the two steps together gives an exact deletion rule for the whole pipeline. We state it here in words; \cref{app:formal} gives the full statement with every condition.

\begin{theorem}[Exact deletion through Write--Fit--Select--Augment; informal]\label{thm:pipeline-inf}
Let the features be fixed in advance and let ties be broken in a fixed order. Store the sums of every stage, the correctness of each validation point, and the margin and pseudo-label of each validation and pool point. Then for every deletion request $F$ the following steps return exactly the model that retraining on $D\setminus F$ would return.
\begin{enumerate}
\item[(a)] \emph{Fit:} subtract the deleted points from the sums and solve again.
\item[(b)] \emph{Select:} with squared error as the score, recompute each candidate's score from stored sums. With accuracy as the score, recheck only the validation points whose margin is at most $\sqrt2\|\Delta_h\|$.
\item[(c)] \emph{Augment:} go round by round. Compute the new weights, and relabel only the pool points whose margin is at most $\sqrt2\|\Delta_r\|$.
\end{enumerate}
In addition, a cheap test (a \emph{certificate}) can often prove from the deleted points alone that no choice changes. When it passes, the deletion costs the same whatever the size of the dataset.
\end{theorem}

The fitting step (a) is known \citep{cao2015towards,guo2020certified,ginart2019making}, and selection by squared error from stored sums is classical for ridge regression \citep{allen1974relationship,golub1979generalized}. Bounding how far predicted labels can move under a small change of the data goes back to \citet{okumura2015quick}. Our contribution is to use it for selection and, by induction over the rounds, for self-training, so that the whole pipeline becomes exactly deletable.

\paragraph{What it costs.} The stored information is of two kinds. A compact part holds the $d\times d$ sums and the weights; its size does not depend on the number of training points. A full part also holds the validation and pool features with their margins, which the exact rechecks of (b) and (c) need. For a single ridge fit, the compact sums are as small as any exact method can store: we prove that answering every possible single deletion exactly needs on the order of $d^2$ numbers (\cref{thm:cost}). The time saved has a limit too. Solving the new system costs about $d^3/3$ operations however small the request is, so the speed-up over retraining on $N$ points is about $\min(N/|F|,\,3N/d)$, not $N/|F|$.

\paragraph{Requests that depend on earlier models.} A requester may look at the published model and then choose what to delete. Some exact methods fail in this case, because the published model reveals random choices that the method reuses \citep{gupta2021adaptive}. Our pipeline makes no random choices once its features are written, so it stays exact for any sequence of requests (\cref{prop:adaptive}). This is a reason to \emph{write} features rather than sample them from the data. Features built from training images chosen at random, such as cluster centres, behave differently: if the model reveals the centres, a requester who deletes only other points keeps every centre with certainty, while retraining would draw new centres (\cref{thm:centres}).

\paragraph{Exact is not the same as private.} Exactness compares the new model with a retrained one. It says nothing about an attacker who sees both the old and the new model. By \eqref{eq:smw-main}, their difference is built from the features of the deleted points, so it can point to them. \Cref{sec:x3} measures how strong this leak is.

\section{Checking forgetting in trained networks}\label{sec:eval}

We now turn to our second question. For a network trained by gradient descent there is no cheap exact deletion, so the question becomes how to \emph{check} a heuristic one. The usual tests look at the model's outputs. We show that outputs cannot settle the question, propose a check that looks inside the network, and analyse a simple edit that is useful for testing these ideas.

\subsection{Output scores can be fooled}

Many unlearning methods for class forgetting reach zero accuracy on the forgotten class, including the trivial method of deleting the class's output unit. The next result shows that this is not a coincidence: any score computed from outputs alone can be matched by a model that still holds everything it knew.

\begin{proposition}[Outputs cannot prove deep forgetting; informal]\label{prop:padding-inf}
Take any score computed from a model's outputs, such as accuracy on the forgotten data, a membership attack that queries the model, or a distance to the retrained model's outputs. For any model $g$, we can build a model $\tilde g$ with exactly the same outputs as $g$ on every input, whose hidden layer also contains all features of the original model $f$, connected to the output by zero weights. A linear probe on $\tilde g$ then recovers anything it could recover from $f$. In particular, deleting the output unit of the forgotten class gives zero forget accuracy while leaving every hidden layer exactly as it was.
\end{proposition}

\citet{yong2026erased} build output-identical models with different hidden features; our statement covers any output score and gives the delete-unit example as a corollary (\cref{prop:padding}). The lesson is that a check of deep forgetting has to read the inside of the network.

\subsection{Compare with retraining, layer by layer}

Reading the inside raises a new question: what value should a probe reach? Zero is the wrong target. A retrained model has never seen the forgotten data, yet a probe can still find, for example, the forgotten class, because it is similar to the classes it still knows. So we compare with retrained models. We train a small band of models from scratch without the forgotten data, using different random seeds. At each layer we then measure a probe score, such as how well a linear probe separates forgotten from kept data, or how well a new read-out trained on that layer can relearn the forgotten class. The \emph{ProbeGap} of a method at a layer is the distance between its probe score and the band's average (\cref{def:probegap}). A good method lands inside the band. A method far above the band has hidden the data rather than removed it. A method far below the band has \emph{over-unlearned}: it erased more than retraining would, which is also detectable \citep{ha2026unlearnings}. Because a method could hide information in a layer we do not probe, we report every layer.

\subsection{A one-layer edit with a guarantee}

To test these ideas we need an edit whose effect at one layer we can predict exactly. We use a \emph{projection edit}, proj$(k)$. At a chosen layer, we find the $k$ directions in which the activations of the forgotten data differ most from those of the kept data, measured relative to the natural spread of the kept data. We then remove these $k$ directions and fold the result into the layer's weights. The edit needs no gradient steps and is closely related to known projection methods \citep{kodge2024deep,belrose2023leace}.

\begin{theorem}[What the projection edit guarantees; informal]\label{thm:probe-inf}
At the edited layer, proj$(k)$ removes as much of the forgotten data's extra signal as any $k$ directions can. After the edit, a linear probe that tries to tell forgotten from kept points can explain at most a fraction $t/(1+t)$ of the variance, where $t$ measures the gap between the two groups' averages that survives the edit. This fraction is always below 1, and it is 0 when the edit removes that gap completely. Two warnings come with the guarantee:
\begin{enumerate}
\item[(i)] it covers the edited layer only, and says nothing about earlier or later layers;
\item[(ii)] if the forgotten data are just a random sample of the kept data, the chosen directions are noise. The edit then removes nothing real, and if the forgotten set has at most $k$ points, it makes those points stand out.
\end{enumerate}
\end{theorem}

The full statement is \cref{thm:probe}. Its first part is a classical eigenvalue result \citep{fan1949on}, and the link between the probe and the gap of the two groups' averages is classical in discriminant analysis \citep[Ch.~4]{hastie2009elements}. The bound after the edit and the two warnings are ours. The bound needs the edit to use the exact spread of the kept data; the edits we deploy add a small regulariser for numerical stability, so they carry no formal bound. Warning (i) is exactly what our backdoor study tests. A low attack success rate at the output tells us nothing about the layers below the edit, and probes show that the trigger is still there (\cref{sec:x7}).

\section{Experiments}\label{sec:experiments}

Each subsection answers one question. Three rules hold throughout. \emph{(1)} Every choice is made on a validation split of the training data, and the test set is read once, to score a frozen model. \emph{(2)} Before each run we wrote our prediction in a dated notebook (self-attested; 70 of 131 held, \cref{app:prereg}), and we name the failures. \emph{(3)} The \emph{written} arm uses the hand-written features of \cref{fig:filters}, so the guarantee covers every weight; the \emph{backbone} arm uses frozen DINOv2 or CLIP features, which are accurate but only exact downstream. Runs are numbered X1--X12; \cref{app:exp} has every setting and table.

\subsection{Is the cheap deletion really exact?}\label{sec:x1}\label{sec:x2}

\Cref{thm:pipeline-inf} assumes exact arithmetic, but computers round. So we compared every cheap deletion with a full retrain.

\paragraph{One read-out, 1,000 deletions in a row (X1).} Rounding errors could pile up. They did not: after $1{,}000$ deletions on six feature sets, the weights are within $6.8\cdot10^{-10}$ of retraining, and test predictions are identical in all 12 runs (\cref{tab:x1}).

\paragraph{The whole pipeline (X2, X10, X12).} On CIFAR-10 we run all four stages ($1{,}000$ labels, $44{,}000$ unlabelled images, 8 ridge strengths, 3 self-training rounds) and send 240 requests from all three data sets. Every request gives the same ridge strength, the same $3\times44{,}000$ pseudo-labels, and weights within $6\cdot10^{-12}$ of a rerun (\cref{tab:x2}). Repeating this on CIFAR-100, Tiny-ImageNet and new request orders adds 1,920 more requests, all exact (\cref{tab:x2gen}). X12 checks one detail: the feature scale is now computed from random-noise images instead of training images, and everything stays exact.

\paragraph{The certificates rarely pass, and that is fine.} Exactness comes from the rechecks, not the certificates; a certificate only saves work. The winning ridge strength usually leads by one validation point, and the smallest pool margin is near zero, so the certificates seldom pass. Pseudo-labels really do flip: a few per round on the backbone arm, hundreds on the written arm.

\subsection{Where does the speed come from?}

\begin{figure}[t]
\centering
\includegraphics[width=\columnwidth]{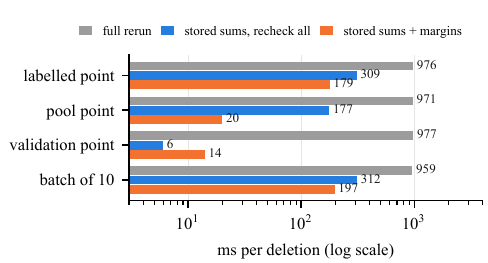}
\caption{Time per deletion on the backbone arm of X2 (median, ms, log scale). Storing the sums (``cached'') gives most of the gain over a rerun; the margin rechecks add most for pool deletions.}\label{fig:speed}
\end{figure}

\paragraph{One read-out.} Solving again from the updated sums is $52$--$70\times$ faster than retraining on $45{,}000$ DINOv2 features, and updating a stored inverse is $158$--$193\times$ faster (X1). Certified removal \citep{guo2020certified} must add noise and loses accuracy ($94.8$ vs $96.2\%$); SISA \citep{bourtoule2021machine} is only $4$--$14\times$ faster (\cref{tab:x1b}).

\paragraph{The whole pipeline.} A simple exact baseline that stores the same sums but rechecks \emph{every} point gets most of the speed-up (\cref{fig:speed}, X8a). The margins help most for pool deletions (177 to 20\,ms). With 100--200 classes, margins are small, and the gain over a rerun drops to about $3\times$ (\cref{tab:x2gen}). In short, \emph{the sums give most of the speed}.

\subsection{Does exact deletion hide who asked?}\label{sec:x3}

An attacker who sees only the new model learns nothing beyond retraining. But a real attacker may also have the old model.

\paragraph{Set-up (X3).} We fit 256 small ridge read-outs that overfit, so there is something to leak. Each deletes a random 10\% of its data. A standard membership attack (U-LiRA, \citealp{hayes2025inexact}) tries to spot the deleted points.

\paragraph{Result.} Against the new model alone, the attack does exactly as well as against retraining: chance (\cref{tab:x3}). Against the old and new model together, the deleted points stand out: AUC $0.953$ for a batch of 250 and $0.9998$ for a single deletion (\cref{tab:x3batch}), and $0.89$--$0.98$ on other backbones and datasets. Serving labels only helps on CIFAR-10 but not CIFAR-100. Exact means ``as if never used'', not ``nobody can tell someone asked''; hiding that needs added noise \citep{dwork2014algorithmic}, which we did not test.

\subsection{Can output scores be fooled in practice?}\label{sec:x4}

\paragraph{Set-up (X4).} We forget the digit 3 from a small CNN on MNIST and Fashion-MNIST with 14 methods, and compare each with five models retrained without it (\cref{tab:x4,fig:x4}).

\begin{figure}[t]
\centering
\includegraphics[width=\columnwidth]{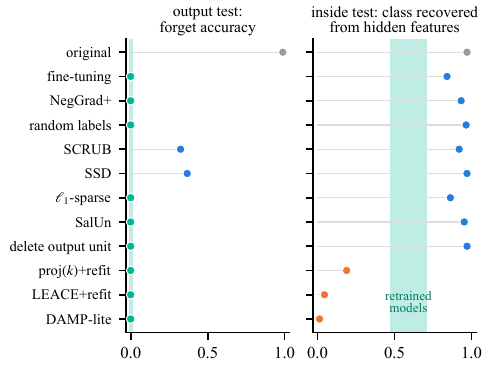}
\caption{Forgetting the digit 3 (MNIST, mean of 3 seeds). Left: the output test calls almost every method perfect. Right: how well the class can be recovered from each model's hidden features by one linear map back to the original's. The shaded band is what retrained models give. Blue: the class is still there; orange: erased more than retraining would.}\label{fig:probes}
\end{figure}

\paragraph{Result.} Almost every method, including deleting the output unit, reaches at most $0.6\%$ forget accuracy, so outputs cannot rank them (\cref{fig:probes}). Probes can. Delete-unit leaves every probe at the original value. Gradient methods let a new read-out relearn the class above the retrained band; projection edits fall below it (over-unlearning). No heuristic method sits in the band on every probe. The edit's bound holds at its layer on every seed, but one layer later a probe recovers the class again ($R^2$ from $0.014$ to $0.46$).

\subsection{Case study: removing a backdoor}\label{sec:x5}\label{sec:x7}

A backdoor attacker relabels some training images and stamps them with a trigger, so the model obeys the trigger. In practice a defender finds only some poisoned images, and retraining without those fails because the rest still teach the trigger \citep{goel2024corrective}. This makes backdoors a good test of whether a low attack success rate (ASR) means the trigger is gone.

\paragraph{Set-up (X5, X7, X9).} We poison 1\% of CIFAR-10 with a patch (BadNets), a blended noise image (Blended) or a warp (WaNet), and train ResNet-18 (clean accuracy $92.7\%$). The defender knows 10\% of the poisons. The edit removes the directions that separate them from clean images, at the last layer or a middle layer, chosen on validation data. We compare with retraining on the remaining data, fine-tuning (FT) and a simplified SalUn \citep{fan2024salun}, over five seeds.

\begin{table}[t]
\centering\footnotesize
\setlength{\tabcolsep}{4pt}
\begin{tabular}{@{}lrrr@{}}
\toprule
ASR at the output (\%) & BadNets & Blended & WaNet\\\midrule
original model & 94.6 & 99.9 & 77.1\\
retrain on the remaining data & 94.8 & 99.9 & 76.1\\
fine-tuning (FT) & 94.5 & 99.9 & 77.4\\
SalUn (simplified) & 0.1 & 0.0 & 1.4\\
projection edit, layer by validation & 1.5 & 7.8 & 5.5\\
\bottomrule
\end{tabular}
\caption{Backdoor attack success at the output on CIFAR-10 ResNet-18, 10\% of poisons known, mean of 5 seeds (intervals in \cref{tab:bd5}). Retraining and FT keep the backdoor; the edit and SalUn remove it from the output, but not from the network (\cref{fig:overview}b).}\label{tab:bd-main}
\end{table}

\paragraph{At the output, the edit works.} Retraining and FT keep the backdoor at $76$--$100\%$; the edit lowers it to $1.5$--$7.8\%$ at a cost of at most $0.2$ points of clean accuracy (\cref{tab:bd-main}). The layer matters, and the validation rule picks the better one in 29 of 30 BadNets and Blended cases and all WaNet cases. Unlike a published report \citep{yu2026instantforget}, this also removes blended and warp triggers from the output.

\paragraph{Inside, the trigger survives.} A new read-out trained on SalUn's features brings the backdoor back to $79$--$100\%$. For the last-layer edit, the features just before the edit still carry it ($94$, $100$, $73\%$, against $1.5$, $58$, $2.4\%$ for a clean model). One training step returns the attack to $97$, $100$ and $76\%$, while a clean model stays at $1$--$2\%$ (\cref{fig:overview}b); the middle-layer edit takes five steps. Finding the unknown poisons and retraining beats the edit on two triggers, at the cost of a full retrain (\cref{tab:defence}), and the edit fails in 1 of 10 further settings (\cref{tab:general}). So the edit is not a defence. It is a clear case of a method that passes the output test and fails the inside test.

\subsection{What if the forgotten data are a random sample?}\label{sec:x6}

A random 10\% of the data is a common benchmark request, and warning (ii) of \cref{thm:probe-inf} says the edit then has nothing real to remove. In simulation the measured quantities are within a few percent of the theory's predictions (\cref{fig:x1x6}), and on ResNet-18 the chosen directions look random (\cref{tab:rand5}). On a CLIP read-out the edit moves the forgotten points \emph{away} from the retrained model (gap $0.0033\to0.13$). On ResNet-18, removing directions found from a random \emph{kept} set changes the forgotten points almost as much: the edit mostly lowers confidence everywhere. For random requests, use exact deletion.

\section{Related work}\label{sec:related}

\paragraph{Exact and certified deletion.} Subtracting a point's contribution from stored sums is the founding idea of unlearning \citep{cao2015towards}. It covers models in summation form, including feature selection, but not model selection or self-training. Exact ridge read-outs on frozen features are now common \citep{huang2025towards,quan2026exact,tavory2026exact,yang2025muso,zhu2019efficient,golatkar2021mixed}, and they leave the feature extractor outside the guarantee. Certified removal adds noise after a Newton step \citep{guo2020certified}, and other methods are approximate \citep{golatkar2020eternal,izzo2021approximate,neel2021descent,mahadevan2021certifiable,chien2024langevin,koloskova2025certified} or exact only in distribution \citep{ullah2021machine}. For deep networks, SISA retrains one shard of the data \citep{bourtoule2021machine}, and DaRE does the same for random forests \citep{brophy2021machine}. Approximate unlearning is known to fail when the model is also selected by cross-validation \citep{suriyakumar2022algorithms}. Local unlearning methods face impossibility results in multi-stage training \citep{yu2025on}. Randomised exact schemes can break when requests depend on the published models \citep{gupta2021adaptive,chourasia2023forget}, and forgetting cannot be verified from the weights alone \citep{thudi2022on,zhang2024verification}. We add exactness for selection and self-training, and the step from frozen to fully written features.

\paragraph{Projection edits and backdoors.} Projection without gradients has been used for class forgetting \citep{kodge2024deep,hatami2026class}. LEACE gives the smallest linear change that hides a concept from linear probes \citep{belrose2023leace}, following INLP \citep{ravfogel2020null}. Similar covariance statistics detect backdoors \citep{tran2018spectral,hayase2021spectre}, and dedicated defences remove backdoors from a few isolated poisons \citep{li2021anti,zeng2022adversarial}. Projection is reported to remove mainly patch triggers \citep{yu2026instantforget}, and unlearning often fails to remove poisons \citep{pawelczyk2025machine}, especially when only some are known \citep{goel2024corrective}. We add a guarantee for the edited layer and evidence that the trigger survives below it.

\paragraph{Evaluating unlearning.} Probes and relearning recover classes that output scores call forgotten \citep{goel2022towards,gao2026illusion,jung2025one,habbati2026gradient,ha2026unlearnings}, and agreeing with the retrained model's outputs does not imply agreeing in the hidden features \citep{yong2026erased}. Per-example membership attacks \citep{hayes2025inexact,carlini2022membership,naderloui2025rectifying} and a competition metric \citep{triantafillou2024are} have replaced earlier population-level tests. Other work studies why unlearning is hard \citep{zhao2024what}, scores it by information \citep{jeon2026information}, or tunes benchmarks against retraining \citep{cadet2025deep,rinberg2026easy}. We add the proof that no output score suffices, and a check that compares each layer with retrained models.

\section{Limitations}\label{sec:limits}
\emph{Accuracy.} The guarantee covers every weight only on fully written features, and these are weak ($49.2$ / $45.5\%$ on CIFAR-10 before and after self-training). On DINOv2 or CLIP features, it covers only the downstream data. \emph{Privacy.} Exactness does not stop an attacker who compares the model before and after a deletion, and we tested no defence against this. \emph{Cost.} The certificates rarely pass on large pools. The exact rechecks need the pool features in storage (up to $1.14$ GB), and with 100--200 classes they are only about $3\times$ faster than a rerun for labelled deletions ($16$--$19\times$ for pool deletions). \emph{The edit.} Its bound covers the edited layer only. Its layer rule was fixed after three seeds, and it fails for Blended at 5\% poisons. It loses to SalUn on ASR and to each find-then-retrain defence on two triggers. \emph{Baselines and scope.} ABL had one schedule, I-BAU none, and the MNIST baselines one round of tuning. We tested no language models.

\section{Conclusion}\label{sec:conclusion}
Whether a deletion can be proved correct depends on what we look at. If a model is computed by exact formulas from features written in advance, a cheap deletion gives exactly the retrained model, through selection and self-training as well as through the fit. This exactness does not hide who asked. If a model is trained by gradient descent, its outputs cannot show that data are gone, and a heuristic edit can pass every output test while the information stays in earlier layers \citep{cooper2025machine}. In short, forgetting can be proved by how a model is computed, not by what it outputs. For trained networks, check every layer against retraining.

\section*{Impact Statement}
This paper is about deleting data from trained models, which relates to the legal right to erasure. Model providers who must honour deletion requests benefit: in a written or frozen-backbone pipeline, exact deletion is cheap and can be checked by a rerun. Defenders who know only part of a poisoning attack get a cheap edit and a warning about how to score it. Two groups stay exposed. People in the pretraining data of a frozen backbone cannot have their data removed by our method. The requesters themselves are exposed to anyone who holds candidate records and compares the model before and after their deletion (\cref{sec:x3}): in our tests, for batches of 250 deletions, this gave AUC $0.89$--$0.98$ and $11$--$36\%$ TPR at $0.1\%$ FPR. Serving labels instead of scores gave near-chance attack AUC on CIFAR-10 but not on CIFAR-100. Hiding them would require a separately specified privacy mechanism, for example noise calibrated to a proven worst-case sensitivity bound for a defined request adjacency over the release transcript; we neither establish nor test one. Our backdoor results are not a safety guarantee: the trigger stays in the earlier layers, and a few fine-tuning steps on poisoned data bring it back (one step after a penultimate-layer edit), so the edit gives no protection against re-poisoning. A projection edit applied to a self-selected forget set may also change predictions for retained people who share its attributes.

\bibliography{forgetting_by_rewriting}

\begin{thebibliography}{80}
\providecommand{\natexlab}[1]{#1}
\providecommand{\url}[1]{\texttt{#1}}
\expandafter\ifx\csname urlstyle\endcsname\relax
  \providecommand{\doi}[1]{doi: #1}\else
  \providecommand{\doi}{doi: \begingroup \urlstyle{rm}\Url}\fi

\bibitem[Allen(1974)]{allen1974relationship}
Allen, D.~M.
\newblock The relationship between variable selection and data augmentation and
  a method for prediction.
\newblock \emph{Technometrics}, 16\penalty0 (1):\penalty0 125--127, 1974.

\bibitem[Bai \& Yin(1993)Bai and Yin]{bai1993limit}
Bai, Z.~D. and Yin, Y.~Q.
\newblock Limit of the smallest eigenvalue of a large dimensional sample
  covariance matrix.
\newblock \emph{The Annals of Probability}, 21\penalty0 (3):\penalty0
  1275--1294, 1993.
\newblock \doi{10.1214/aop/1176989118}.

\bibitem[Barni et~al.(2019)Barni, Kallas, and Tondi]{barni2019new}
Barni, M., Kallas, K., and Tondi, B.
\newblock A new backdoor attack in {CNNs} by training set corruption without
  label poisoning.
\newblock In \emph{2019 IEEE International Conference on Image Processing
  (ICIP)}, pp.\  101--105, 2019.
\newblock \doi{10.1109/ICIP.2019.8802997}.

\bibitem[Baumhauer et~al.(2022)Baumhauer, Sch{\"o}ttle, and
  Zeppelzauer]{baumhauer2022machine}
Baumhauer, T., Sch{\"o}ttle, P., and Zeppelzauer, M.
\newblock Machine unlearning: Linear filtration for logit-based classifiers.
\newblock \emph{Machine Learning}, 111\penalty0 (9):\penalty0 3203--3226, 2022.
\newblock \doi{10.1007/s10994-022-06178-9}.

\bibitem[Belrose et~al.(2023)Belrose, Schneider-Joseph, Ravfogel, Cotterell,
  Raff, and Biderman]{belrose2023leace}
Belrose, N., Schneider-Joseph, D., Ravfogel, S., Cotterell, R., Raff, E., and
  Biderman, S.
\newblock {LEACE}: Perfect linear concept erasure in closed form.
\newblock In \emph{Advances in Neural Information Processing Systems
  (NeurIPS)}, volume~36, 2023.

\bibitem[Bourtoule et~al.(2021)Bourtoule, Chandrasekaran, Choquette-Choo, Jia,
  Travers, Zhang, Lie, and Papernot]{bourtoule2021machine}
Bourtoule, L., Chandrasekaran, V., Choquette-Choo, C.~A., Jia, H., Travers, A.,
  Zhang, B., Lie, D., and Papernot, N.
\newblock Machine unlearning.
\newblock In \emph{2021 IEEE Symposium on Security and Privacy (SP)}, pp.\
  141--159, 2021.
\newblock \doi{10.1109/SP40001.2021.00019}.

\bibitem[Brophy \& Lowd(2021)Brophy and Lowd]{brophy2021machine}
Brophy, J. and Lowd, D.
\newblock Machine unlearning for random forests.
\newblock In \emph{Proceedings of the 38th International Conference on Machine
  Learning (ICML)}, volume 139 of \emph{Proceedings of Machine Learning
  Research}, pp.\  1092--1104, 2021.

\bibitem[Cadet et~al.(2025)Cadet, Borovykh, Malekzadeh, Ahmadi-Abhari, and
  Haddadi]{cadet2025deep}
Cadet, X.~F., Borovykh, A., Malekzadeh, M., Ahmadi-Abhari, S., and Haddadi, H.
\newblock Deep unlearn: Benchmarking machine unlearning for image
  classification.
\newblock In \emph{2025 IEEE 10th European Symposium on Security and Privacy
  (EuroS\&P)}, pp.\  939--962, 2025.
\newblock \doi{10.1109/EuroSP63326.2025.00058}.

\bibitem[Cao \& Yang(2015)Cao and Yang]{cao2015towards}
Cao, Y. and Yang, J.
\newblock Towards making systems forget with machine unlearning.
\newblock In \emph{2015 IEEE Symposium on Security and Privacy (SP)}, pp.\
  463--480. IEEE, 2015.
\newblock \doi{10.1109/SP.2015.35}.

\bibitem[Carlini et~al.(2022)Carlini, Chien, Nasr, Song, Terzis, and
  Tram{\`e}r]{carlini2022membership}
Carlini, N., Chien, S., Nasr, M., Song, S., Terzis, A., and Tram{\`e}r, F.
\newblock Membership inference attacks from first principles.
\newblock In \emph{2022 IEEE Symposium on Security and Privacy (SP)}, pp.\
  1897--1914, 2022.
\newblock \doi{10.1109/SP46214.2022.9833649}.

\bibitem[Chen et~al.(2021)Chen, Zhang, Wang, Backes, Humbert, and
  Zhang]{chen2021when}
Chen, M., Zhang, Z., Wang, T., Backes, M., Humbert, M., and Zhang, Y.
\newblock When machine unlearning jeopardizes privacy.
\newblock In \emph{Proceedings of the 2021 ACM SIGSAC Conference on Computer
  and Communications Security (CCS)}, pp.\  896--911, 2021.
\newblock \doi{10.1145/3460120.3484756}.

\bibitem[Chen et~al.(2017)Chen, Liu, Li, Lu, and Song]{chen2017targeted}
Chen, X., Liu, C., Li, B., Lu, K., and Song, D.
\newblock Targeted backdoor attacks on deep learning systems using data
  poisoning, 2017.

\bibitem[Cherapanamjeri et~al.(2025)Cherapanamjeri, Garg, Rajaraman, Sekhari,
  and Shetty]{cherapanamjeri2025space}
Cherapanamjeri, Y., Garg, S., Rajaraman, N., Sekhari, A., and Shetty, A.
\newblock The space complexity of learning-unlearning algorithms (extended
  abstract).
\newblock In \emph{Proceedings of the Thirty Eighth Conference on Learning
  Theory (COLT)}, volume 291 of \emph{Proceedings of Machine Learning
  Research}, pp.\  1185--1193, 2025.

\bibitem[Chien et~al.(2024)Chien, Wang, Chen, and Li]{chien2024langevin}
Chien, E., Wang, H., Chen, Z., and Li, P.
\newblock Langevin unlearning: A new perspective of noisy gradient descent for
  machine unlearning.
\newblock In \emph{Advances in Neural Information Processing Systems
  (NeurIPS)}, volume~37, 2024.

\bibitem[Chourasia \& Shah(2023)Chourasia and Shah]{chourasia2023forget}
Chourasia, R. and Shah, N.
\newblock Forget unlearning: Towards true data-deletion in machine learning.
\newblock In \emph{Proceedings of the 40th International Conference on Machine
  Learning (ICML)}, volume 202 of \emph{Proceedings of Machine Learning
  Research}, 2023.

\bibitem[Cooper et~al.(2025)Cooper, Choquette-Choo, Bogen, Klyman, Jagielski,
  Filippova, Liu, Chouldechova, Hayes, Huang, Triantafillou, Kairouz, Mitchell,
  Mireshghallah, Jacobs, Grimmelmann, Shmatikov, De~Sa, Shumailov, Terzis,
  Barocas, Wortman~Vaughan, boyd, Choi, Koyejo, Delgado, Liang, Ho, Samuelson,
  Brundage, Bau, Neel, Wallach, Cyphert, Lemley, Papernot, and
  Lee]{cooper2025machine}
Cooper, A.~F., Choquette-Choo, C.~A., Bogen, M., Klyman, K., Jagielski, M.,
  Filippova, K., Liu, K., Chouldechova, A., Hayes, J., Huang, Y.,
  Triantafillou, E., Kairouz, P., Mitchell, N.~E., Mireshghallah, N., Jacobs,
  A.~Z., Grimmelmann, J., Shmatikov, V., De~Sa, C., Shumailov, I., Terzis, A.,
  Barocas, S., Wortman~Vaughan, J., boyd, d., Choi, Y., Koyejo, S., Delgado,
  F., Liang, P., Ho, D.~E., Samuelson, P., Brundage, M., Bau, D., Neel, S.,
  Wallach, H., Cyphert, A.~B., Lemley, M.~A., Papernot, N., and Lee, K.
\newblock Machine unlearning doesn't do what you think: Lessons for generative
  {AI} policy and research.
\newblock In \emph{Advances in Neural Information Processing Systems
  (NeurIPS)}, volume~38, 2025.
\newblock Position paper; oral.

\bibitem[Dwork \& Roth(2014)Dwork and Roth]{dwork2014algorithmic}
Dwork, C. and Roth, A.
\newblock The algorithmic foundations of differential privacy.
\newblock \emph{Foundations and Trends in Theoretical Computer Science},
  9\penalty0 (3--4):\penalty0 211--487, 2014.
\newblock \doi{10.1561/0400000042}.

\bibitem[Fan et~al.(2024)Fan, Liu, Zhang, Wong, Wei, and Liu]{fan2024salun}
Fan, C., Liu, J., Zhang, Y., Wong, E., Wei, D., and Liu, S.
\newblock {SalUn}: Empowering machine unlearning via gradient-based weight
  saliency in both image classification and generation.
\newblock In \emph{International Conference on Learning Representations
  (ICLR)}, 2024.

\bibitem[Fan(1949)]{fan1949on}
Fan, K.
\newblock On a theorem of {W}eyl concerning eigenvalues of linear
  transformations {I}.
\newblock \emph{Proceedings of the National Academy of Sciences}, 35\penalty0
  (11):\penalty0 652--655, 1949.
\newblock \doi{10.1073/pnas.35.11.652}.

\bibitem[Foster et~al.(2024)Foster, Schoepf, and Brintrup]{foster2024fast}
Foster, J., Schoepf, S., and Brintrup, A.
\newblock Fast machine unlearning without retraining through selective synaptic
  dampening.
\newblock In \emph{Proceedings of the AAAI Conference on Artificial
  Intelligence}, volume~38, pp.\  12043--12051, 2024.
\newblock \doi{10.1609/aaai.v38i11.29092}.

\bibitem[Gao et~al.(2026)Gao, Unal, Rangamani, and Zhu]{gao2026illusion}
Gao, Y., Unal, A., Rangamani, A., and Zhu, Z.
\newblock An illusion of unlearning? assessing machine unlearning through
  internal representations.
\newblock In \emph{Proceedings of the 29th International Conference on
  Artificial Intelligence and Statistics (AISTATS)}, volume 300 of
  \emph{Proceedings of Machine Learning Research}, pp.\  4114--4122, 2026.

\bibitem[Ghazi et~al.(2023)Ghazi, Kamath, Kumar, Manurangsi, Sekhari, and
  Zhang]{ghazi2023ticketed}
Ghazi, B., Kamath, P., Kumar, R., Manurangsi, P., Sekhari, A., and Zhang, C.
\newblock Ticketed learning--unlearning schemes.
\newblock In \emph{Proceedings of the 36th Conference on Learning Theory
  (COLT)}, volume 195 of \emph{Proceedings of Machine Learning Research}, 2023.

\bibitem[Ginart et~al.(2019)Ginart, Guan, Valiant, and Zou]{ginart2019making}
Ginart, A., Guan, M.~Y., Valiant, G., and Zou, J.
\newblock Making {AI} forget you: Data deletion in machine learning.
\newblock In \emph{Advances in Neural Information Processing Systems
  (NeurIPS)}, volume~32, 2019.

\bibitem[Goel et~al.(2022)Goel, Prabhu, Sanyal, Lim, Torr, and
  Kumaraguru]{goel2022towards}
Goel, S., Prabhu, A., Sanyal, A., Lim, S.-N., Torr, P., and Kumaraguru, P.
\newblock Towards adversarial evaluations for inexact machine unlearning, 2022.

\bibitem[Goel et~al.(2024)Goel, Prabhu, Torr, Kumaraguru, and
  Sanyal]{goel2024corrective}
Goel, S., Prabhu, A., Torr, P., Kumaraguru, P., and Sanyal, A.
\newblock Corrective machine unlearning.
\newblock \emph{Transactions on Machine Learning Research}, 2024.

\bibitem[Golatkar et~al.(2020)Golatkar, Achille, and
  Soatto]{golatkar2020eternal}
Golatkar, A., Achille, A., and Soatto, S.
\newblock Eternal sunshine of the spotless net: Selective forgetting in deep
  networks.
\newblock In \emph{IEEE/CVF Conference on Computer Vision and Pattern
  Recognition (CVPR)}, 2020.

\bibitem[Golatkar et~al.(2021)Golatkar, Achille, Ravichandran, Polito, and
  Soatto]{golatkar2021mixed}
Golatkar, A., Achille, A., Ravichandran, A., Polito, M., and Soatto, S.
\newblock Mixed-privacy forgetting in deep networks.
\newblock In \emph{IEEE/CVF Conference on Computer Vision and Pattern
  Recognition (CVPR)}, pp.\  792--801, 2021.
\newblock \doi{10.1109/CVPR46437.2021.00085}.

\bibitem[Golub et~al.(1979)Golub, Heath, and Wahba]{golub1979generalized}
Golub, G.~H., Heath, M., and Wahba, G.
\newblock Generalized cross-validation as a method for choosing a good ridge
  parameter.
\newblock \emph{Technometrics}, 21\penalty0 (2):\penalty0 215--223, 1979.

\bibitem[Gu et~al.(2017)Gu, Dolan-Gavitt, and Garg]{gu2017badnets}
Gu, T., Dolan-Gavitt, B., and Garg, S.
\newblock {BadNets}: Identifying vulnerabilities in the machine learning model
  supply chain, 2017.
\newblock Journal version: Gu, Liu, Dolan-Gavitt, Garg, ``BadNets: Evaluating
  Backdooring Attacks on Deep Neural Networks'', IEEE Access 7:47230--47244,
  2019.

\bibitem[Guo et~al.(2020)Guo, Goldstein, Hannun, and van~der
  Maaten]{guo2020certified}
Guo, C., Goldstein, T., Hannun, A., and van~der Maaten, L.
\newblock Certified data removal from machine learning models.
\newblock In \emph{Proceedings of the 37th International Conference on Machine
  Learning (ICML)}, volume 119 of \emph{Proceedings of Machine Learning
  Research}, pp.\  3832--3842, 2020.

\bibitem[Gupta et~al.(2021)Gupta, Jung, Neel, Roth, Sharifi-Malvajerdi, and
  Waites]{gupta2021adaptive}
Gupta, V., Jung, C., Neel, S., Roth, A., Sharifi-Malvajerdi, S., and Waites, C.
\newblock Adaptive machine unlearning.
\newblock In \emph{Advances in Neural Information Processing Systems
  (NeurIPS)}, volume~34, 2021.

\bibitem[Ha et~al.(2026)Ha, Park, and Yoon]{ha2026unlearnings}
Ha, S., Park, S., and Yoon, S.~W.
\newblock Unlearning's blind spots: Over-unlearning and prototypical relearning
  attack.
\newblock In \emph{Proceedings of the 43rd International Conference on Machine
  Learning (ICML)}, volume 306 of \emph{Proceedings of Machine Learning
  Research}, pp.\  39108--39126, 2026.

\bibitem[Habbati et~al.(2026)Habbati, Merlo, Verderame, and
  Guerar]{habbati2026gradient}
Habbati, B., Merlo, A., Verderame, L., and Guerar, M.
\newblock Gradient concentration, not weight saliency, explains
  representation-level class unlearning, 2026.

\bibitem[Hastie et~al.(2009)Hastie, Tibshirani, and
  Friedman]{hastie2009elements}
Hastie, T., Tibshirani, R., and Friedman, J.
\newblock \emph{The Elements of Statistical Learning: Data Mining, Inference,
  and Prediction}.
\newblock Springer, New York, second edition, 2009.

\bibitem[Hatami et~al.(2026)Hatami, Aalishah, and Monosov]{hatami2026class}
Hatami, A., Aalishah, R., and Monosov, I.~E.
\newblock Class unlearning via depth-aware removal of forget-specific
  directions.
\newblock In \emph{Proceedings of the IEEE/CVF Conference on Computer Vision
  and Pattern Recognition (CVPR) Workshops}, pp.\  8516--8525, 2026.

\bibitem[Hayase et~al.(2021)Hayase, Kong, Somani, and Oh]{hayase2021spectre}
Hayase, J., Kong, W., Somani, R., and Oh, S.
\newblock {SPECTRE}: Defending against backdoor attacks using robust
  statistics.
\newblock In \emph{Proceedings of the 38th International Conference on Machine
  Learning (ICML)}, volume 139 of \emph{Proceedings of Machine Learning
  Research}, pp.\  4129--4139, 2021.

\bibitem[Hayes et~al.(2025)Hayes, Shumailov, Triantafillou, Khalifa, and
  Papernot]{hayes2025inexact}
Hayes, J., Shumailov, I., Triantafillou, E., Khalifa, A., and Papernot, N.
\newblock Inexact unlearning needs more careful evaluations to avoid a false
  sense of privacy.
\newblock In \emph{2025 IEEE Conference on Secure and Trustworthy Machine
  Learning (SaTML)}, pp.\  497--519, 2025.
\newblock \doi{10.1109/SaTML64287.2025.00034}.

\bibitem[Higham(2002)]{higham2002accuracy}
Higham, N.~J.
\newblock \emph{Accuracy and Stability of Numerical Algorithms}.
\newblock Society for Industrial and Applied Mathematics (SIAM), Philadelphia,
  PA, second edition, 2002.
\newblock \doi{10.1137/1.9780898718027}.

\bibitem[Huang et~al.(2025)Huang, Tang, Fan, Zhuang, Liu, Wang, Liu, Dong, and
  Song]{huang2025towards}
Huang, Y., Tang, J., Fan, K., Zhuang, H., Liu, A., Wang, T., Liu, Y., Dong, M.,
  and Song, H.~H.
\newblock Towards efficient and exact forgetting services in
  pre-trained-model-based continual learning, 2025.

\bibitem[Izzo et~al.(2021)Izzo, Smart, Chaudhuri, and Zou]{izzo2021approximate}
Izzo, Z., Smart, M.~A., Chaudhuri, K., and Zou, J.
\newblock Approximate data deletion from machine learning models.
\newblock In \emph{Proceedings of the 24th International Conference on
  Artificial Intelligence and Statistics (AISTATS)}, volume 130 of
  \emph{Proceedings of Machine Learning Research}, pp.\  2008--2016, 2021.

\bibitem[Jeon et~al.(2026)Jeon, Jeung, Kim, No, and Choi]{jeon2026information}
Jeon, D., Jeung, W., Kim, T., No, A., and Choi, J.
\newblock An information theoretic evaluation metric for strong unlearning.
\newblock In \emph{Proceedings of the AAAI Conference on Artificial
  Intelligence}, volume~40, pp.\  22173--22181, 2026.
\newblock \doi{10.1609/aaai.v40i26.39373}.

\bibitem[Jia et~al.(2023)Jia, Liu, Ram, Yao, Liu, Liu, Sharma, and
  Liu]{jia2023model}
Jia, J., Liu, J., Ram, P., Yao, Y., Liu, G., Liu, Y., Sharma, P., and Liu, S.
\newblock Model sparsity can simplify machine unlearning.
\newblock In \emph{Advances in Neural Information Processing Systems
  (NeurIPS)}, volume~36, 2023.

\bibitem[Jung et~al.(2025)Jung, Jung, Bae, and Lee]{jung2025one}
Jung, J., Jung, B., Bae, S., and Lee, D.
\newblock One-point contraction: Erasing representational separability toward
  irreversible deep forgetting, 2025.
\newblock Presented at the ICCV 2025 Workshop on Unlearning and Model Editing
  (U\&ME).

\bibitem[Kodge et~al.(2024)Kodge, Saha, and Roy]{kodge2024deep}
Kodge, S., Saha, G., and Roy, K.
\newblock Deep unlearning: Fast and efficient gradient-free class forgetting.
\newblock \emph{Transactions on Machine Learning Research}, 2024.

\bibitem[Kokiopoulou et~al.(2011)Kokiopoulou, Chen, and
  Saad]{kokiopoulou2011trace}
Kokiopoulou, E., Chen, J., and Saad, Y.
\newblock Trace optimization and eigenproblems in dimension reduction methods.
\newblock \emph{Numerical Linear Algebra with Applications}, 18\penalty0
  (3):\penalty0 565--602, 2011.
\newblock \doi{10.1002/nla.743}.

\bibitem[Koloskova et~al.(2025)Koloskova, Allouah, Jha, Guerraoui, and
  Koyejo]{koloskova2025certified}
Koloskova, A., Allouah, Y., Jha, A., Guerraoui, R., and Koyejo, S.
\newblock Certified unlearning for neural networks.
\newblock In \emph{Proceedings of the 42nd International Conference on Machine
  Learning (ICML)}, volume 267 of \emph{Proceedings of Machine Learning
  Research}, pp.\  31275--31298, 2025.

\bibitem[Kurmanji et~al.(2023)Kurmanji, Triantafillou, Hayes, and
  Triantafillou]{kurmanji2023towards}
Kurmanji, M., Triantafillou, P., Hayes, J., and Triantafillou, E.
\newblock Towards unbounded machine unlearning.
\newblock In \emph{Advances in Neural Information Processing Systems
  (NeurIPS)}, volume~36, 2023.

\bibitem[Li et~al.(2021)Li, Lyu, Koren, Lyu, Li, and Ma]{li2021anti}
Li, Y., Lyu, X., Koren, N., Lyu, L., Li, B., and Ma, X.
\newblock Anti-backdoor learning: Training clean models on poisoned data.
\newblock In \emph{Advances in Neural Information Processing Systems
  (NeurIPS)}, volume~34, 2021.

\bibitem[Mahadevan \& Mathioudakis(2021)Mahadevan and
  Mathioudakis]{mahadevan2021certifiable}
Mahadevan, A. and Mathioudakis, M.
\newblock Certifiable machine unlearning for linear models, 2021.

\bibitem[Mar{\v{c}}enko \& Pastur(1967)Mar{\v{c}}enko and
  Pastur]{marchenko1967distribution}
Mar{\v{c}}enko, V.~A. and Pastur, L.~A.
\newblock Distribution of eigenvalues for some sets of random matrices.
\newblock \emph{Mathematics of the USSR-Sbornik}, 1\penalty0 (4):\penalty0
  457--483, 1967.
\newblock \doi{10.1070/SM1967v001n04ABEH001994}.

\bibitem[Naderloui et~al.(2025)Naderloui, Yan, Wang, Fu, Wang, Liu, and
  Hong]{naderloui2025rectifying}
Naderloui, N., Yan, S., Wang, B., Fu, J., Wang, W.~H., Liu, W., and Hong, Y.
\newblock Rectifying privacy and efficacy measurements in machine unlearning: A
  new inference attack perspective.
\newblock In \emph{34th USENIX Security Symposium (USENIX Security 25)}, 2025.

\bibitem[Neel et~al.(2021)Neel, Roth, and Sharifi-Malvajerdi]{neel2021descent}
Neel, S., Roth, A., and Sharifi-Malvajerdi, S.
\newblock Descent-to-delete: Gradient-based methods for machine unlearning.
\newblock In \emph{Proceedings of the 32nd International Conference on
  Algorithmic Learning Theory (ALT)}, volume 132 of \emph{Proceedings of
  Machine Learning Research}, pp.\  931--962, 2021.

\bibitem[Nguyen \& Tran(2021)Nguyen and Tran]{nguyen2021wanet}
Nguyen, A. and Tran, A.
\newblock {WaNet} -- imperceptible warping-based backdoor attack.
\newblock In \emph{International Conference on Learning Representations
  (ICLR)}, 2021.

\bibitem[Nguyen et~al.(2025)Nguyen, Huynh, Ren, Nguyen, Liew, Yin, and
  Nguyen]{nguyen2025survey}
Nguyen, T.~T., Huynh, T.~T., Ren, Z., Nguyen, P.~L., Liew, A. W.-C., Yin, H.,
  and Nguyen, Q. V.~H.
\newblock A survey of machine unlearning.
\newblock \emph{ACM Transactions on Intelligent Systems and Technology},
  16\penalty0 (5):\penalty0 1--46, 2025.
\newblock \doi{10.1145/3749987}.

\bibitem[Okumura et~al.(2015)Okumura, Suzuki, and Takeuchi]{okumura2015quick}
Okumura, S., Suzuki, Y., and Takeuchi, I.
\newblock Quick sensitivity analysis for incremental data modification and its
  application to leave-one-out {CV} in linear classification problems.
\newblock In \emph{Proceedings of the 21th ACM SIGKDD International Conference
  on Knowledge Discovery and Data Mining (KDD)}, pp.\  885--894, 2015.
\newblock \doi{10.1145/2783258.2783347}.

\bibitem[Pawelczyk et~al.(2025)Pawelczyk, Di, Lu, Kamath, Sekhari, and
  Neel]{pawelczyk2025machine}
Pawelczyk, M., Di, J.~Z., Lu, Y., Kamath, G., Sekhari, A., and Neel, S.
\newblock Machine unlearning fails to remove data poisoning attacks.
\newblock In \emph{International Conference on Learning Representations
  (ICLR)}, 2025.

\bibitem[Quan et~al.(2026)Quan, Wu, and Montana]{quan2026exact}
Quan, Y., Wu, W., and Montana, G.
\newblock Exact federated continual unlearning for ridge heads on frozen
  foundation models.
\newblock In \emph{Machine Learning and Knowledge Discovery in Databases.
  Research Track. European Conference, ECML PKDD 2026, Proceedings, Part VI},
  Lecture Notes in Computer Science, pp.\  114--131. Springer, 2026.
\newblock \doi{10.1007/978-3-032-37673-2_7}.

\bibitem[Rahimi \& Recht(2007)Rahimi and Recht]{rahimi2007random}
Rahimi, A. and Recht, B.
\newblock Random features for large-scale kernel machines.
\newblock In \emph{Advances in Neural Information Processing Systems (NIPS)},
  volume~20, pp.\  1177--1184, 2007.

\bibitem[Ravfogel et~al.(2020)Ravfogel, Elazar, Gonen, Twiton, and
  Goldberg]{ravfogel2020null}
Ravfogel, S., Elazar, Y., Gonen, H., Twiton, M., and Goldberg, Y.
\newblock Null it out: Guarding protected attributes by iterative nullspace
  projection.
\newblock In \emph{Proceedings of the 58th Annual Meeting of the Association
  for Computational Linguistics (ACL)}, pp.\  7237--7256, 2020.
\newblock \doi{10.18653/v1/2020.acl-main.647}.

\bibitem[Rinberg et~al.(2026)Rinberg, Puigdemont, Pawelczyk, and
  Cevher]{rinberg2026easy}
Rinberg, R., Puigdemont, P., Pawelczyk, M., and Cevher, V.
\newblock Easy data unlearning bench, 2026.

\bibitem[Scudder(1965)]{scudder1965probability}
Scudder, H.~J.
\newblock Probability of error of some adaptive pattern-recognition machines.
\newblock \emph{IEEE Transactions on Information Theory}, 11\penalty0
  (3):\penalty0 363--371, 1965.
\newblock \doi{10.1109/TIT.1965.1053799}.

\bibitem[Sekhari et~al.(2021)Sekhari, Acharya, Kamath, and
  Suresh]{sekhari2021remember}
Sekhari, A., Acharya, J., Kamath, G., and Suresh, A.~T.
\newblock Remember what you want to forget: Algorithms for machine unlearning.
\newblock In \emph{Advances in Neural Information Processing Systems
  (NeurIPS)}, volume~34, 2021.

\bibitem[Shang et~al.(2026)Shang, Zou, Zhu, and Maleki]{shang2026error}
Shang, C., Zou, H., Zhu, B., and Maleki, A.
\newblock Error accumulation in second-order sequential machine unlearning,
  2026.

\bibitem[Shen et~al.(2024)Shen, Zhang, Zhao, Bialkowski, Chen, and
  Xu]{shen2024label}
Shen, S., Zhang, C., Zhao, Y., Bialkowski, A., Chen, W.~T., and Xu, M.
\newblock Label-agnostic forgetting: A supervision-free unlearning in deep
  models.
\newblock In \emph{International Conference on Learning Representations
  (ICLR)}, 2024.

\bibitem[Sherman \& Morrison(1950)Sherman and Morrison]{sherman1950adjustment}
Sherman, J. and Morrison, W.~J.
\newblock Adjustment of an inverse matrix corresponding to a change in one
  element of a given matrix.
\newblock \emph{The Annals of Mathematical Statistics}, 21\penalty0
  (1):\penalty0 124--127, 1950.
\newblock \doi{10.1214/aoms/1177729893}.

\bibitem[Suriyakumar \& Wilson(2022)Suriyakumar and
  Wilson]{suriyakumar2022algorithms}
Suriyakumar, V.~M. and Wilson, A.~C.
\newblock Algorithms that approximate data removal: New results and
  limitations.
\newblock In \emph{Advances in Neural Information Processing Systems
  (NeurIPS)}, volume~35, 2022.

\bibitem[Tavory et~al.(2026)Tavory, Gade, Sarig, Touitou, and
  Guy]{tavory2026exact}
Tavory, A., Gade, S., Sarig, T., Touitou, N., and Guy, I.
\newblock Exact unlearning via quantized sufficient statistics.
\newblock In \emph{Advances in Neural Information Processing Systems
  (NeurIPS)}, 2026.

\bibitem[Thudi et~al.(2022)Thudi, Jia, Shumailov, and Papernot]{thudi2022on}
Thudi, A., Jia, H., Shumailov, I., and Papernot, N.
\newblock On the necessity of auditable algorithmic definitions for machine
  unlearning.
\newblock In \emph{31st USENIX Security Symposium (USENIX Security 22)}, 2022.

\bibitem[Tran et~al.(2018)Tran, Li, and Madry]{tran2018spectral}
Tran, B., Li, J., and Madry, A.
\newblock Spectral signatures in backdoor attacks.
\newblock In \emph{Advances in Neural Information Processing Systems
  (NeurIPS)}, volume~31, 2018.

\bibitem[Triantafillou et~al.(2024)Triantafillou, Kairouz, Pedregosa, Hayes,
  Kurmanji, Zhao, Dumoulin, Jacques~Junior, Mitliagkas, Wan, Sun~Hosoya,
  Escalera, Dziugaite, Triantafillou, and Guyon]{triantafillou2024are}
Triantafillou, E., Kairouz, P., Pedregosa, F., Hayes, J., Kurmanji, M., Zhao,
  K., Dumoulin, V., Jacques~Junior, J., Mitliagkas, I., Wan, J., Sun~Hosoya,
  L., Escalera, S., Dziugaite, G.~K., Triantafillou, P., and Guyon, I.
\newblock Are we making progress in unlearning? findings from the first
  {NeurIPS} unlearning competition, 2024.

\bibitem[Ullah et~al.(2021)Ullah, Mai, Rao, Rossi, and Arora]{ullah2021machine}
Ullah, E., Mai, T., Rao, A., Rossi, R.~A., and Arora, R.
\newblock Machine unlearning via algorithmic stability.
\newblock In \emph{Proceedings of Thirty Fourth Conference on Learning Theory
  (COLT)}, volume 134 of \emph{Proceedings of Machine Learning Research}, pp.\
  4126--4142, 2021.

\bibitem[Yang et~al.(2025)Yang, He, He, Qiu, and Huang]{yang2025muso}
Yang, R., He, M., He, Z., Qiu, Y., and Huang, X.
\newblock {MUSO}: Achieving exact machine unlearning in over-parameterized
  regimes.
\newblock \emph{Machine Learning}, 114\penalty0 (8), 2025.
\newblock \doi{10.1007/s10994-025-06806-0}.
\newblock Article 176.

\bibitem[Yin et~al.(1988)Yin, Bai, and Krishnaiah]{yin1988limit}
Yin, Y.~Q., Bai, Z.~D., and Krishnaiah, P.~R.
\newblock On the limit of the largest eigenvalue of the large dimensional
  sample covariance matrix.
\newblock \emph{Probability Theory and Related Fields}, 78\penalty0
  (4):\penalty0 509--521, 1988.
\newblock \doi{10.1007/BF00353874}.

\bibitem[Yong et~al.(2026)Yong, Ong, and Chan]{yong2026erased}
Yong, T. P.~Y., Ong, W.~K., and Chan, C.~S.
\newblock Erased, but not gone: Output forgetting is not true forgetting, 2026.

\bibitem[Yu et~al.(2026)Yu, He, Goyal, and Arora]{yu2025on}
Yu, J., He, Y., Goyal, A., and Arora, S.
\newblock On the impossibility of retrain equivalence in machine unlearning.
\newblock In \emph{Conference on Language Modeling (COLM)}, 2026.

\bibitem[Yu(2026)]{yu2026instantforget}
Yu, Z.
\newblock {InstantForget}: Update-free backdoor unlearning with inference-time
  feature reset, 2026.

\bibitem[Zeng et~al.(2022)Zeng, Chen, Park, Mao, Jin, and
  Jia]{zeng2022adversarial}
Zeng, Y., Chen, S., Park, W., Mao, Z.~M., Jin, M., and Jia, R.
\newblock Adversarial unlearning of backdoors via implicit hypergradient.
\newblock In \emph{International Conference on Learning Representations
  (ICLR)}, 2022.

\bibitem[Zhang et~al.(2024)Zhang, Chen, Shen, and Li]{zhang2024verification}
Zhang, B., Chen, Z., Shen, C., and Li, J.
\newblock Verification of machine unlearning is fragile.
\newblock In \emph{Proceedings of the 41st International Conference on Machine
  Learning (ICML)}, volume 235 of \emph{Proceedings of Machine Learning
  Research}, pp.\  58717--58738, 2024.

\bibitem[Zhao et~al.(2024)Zhao, Kurmanji, B{\u{a}}rbulescu, Triantafillou, and
  Triantafillou]{zhao2024what}
Zhao, K., Kurmanji, M., B{\u{a}}rbulescu, G.-O., Triantafillou, E., and
  Triantafillou, P.
\newblock What makes unlearning hard and what to do about it.
\newblock In \emph{Advances in Neural Information Processing Systems
  (NeurIPS)}, volume~37, 2024.

\bibitem[Zhu(2019)]{zhu2019efficient}
Zhu, H.
\newblock Efficient decremental learning algorithms for broad learning system,
  2019.

\end{thebibliography}
\bibliographystyle{icml2026}

\newpage
\appendix
\onecolumn
\section{Formal statements}\label{app:formal}

The main text states each result in words. This appendix gives the precise versions, with every condition. \Cref{app:proofs} proves them. Theorem and proposition numbers match the informal versions in the main text.

\subsection{Notation and the pipeline}\label{app:notation}

\begin{table}[h]
\centering\small
\begin{tabular}{@{}lp{11cm}@{}}
\toprule
symbol & meaning\\\midrule
$L,U,V$; $D$ & labelled set, unlabelled pool, validation set; $D=L\cup U\cup V$, $n=|D|$\\
$F$ & deletion request, $F\subseteq D$\\
$\phi_z\in\R^d$; $R$ & feature vector of point $z$; $R=\max_z\|\phi_z\|$\\
$G_S$, $B_S$ & $\sum_{z\in S}\phi_z\phi_z^\top$ and $\sum_{z\in S}\phi_zy_z^\top$ for a set $S$\\
$H$, $\lambda_h$ & list of candidate ridge strengths and their values\\
$W_h$; $h^\star$ & ridge fit for $\lambda_h$; the selected candidate\\
$k$, $w$ & number of self-training rounds, weight of the pool in the fit\\
$\hat y^{(r)}_u$, $W_r$ & round-$r$ pseudo-label and weights\\
$\mu_W(u)$ & normalised top-1/top-2 margin of $u$ under $W$\\
$\delta_h$, $\delta^{(r)}$ & bounds on the weight change, \cref{eq:deltah,eq:deltar}\\
$\tilde m_r$ & stored lower bound on the round-$r$ pool margins\\
$T(D)$, $T^+(D)$ & compact and full stored state\\
$\|\cdot\|$ & Euclidean norm of a vector; operator norm of a matrix\\
\bottomrule
\end{tabular}
\caption{Notation.}\label{tab:notation}
\end{table}

The pipeline is
\begin{equation}\label{eq:pipeline}
A=\Augment\circ\Select\circ\Fit\circ\Write .
\end{equation}
\Write{} is a feature map $\phi:\mathcal X\to\R^d$ fixed before any data are seen. \Fit{} computes $W_\lambda=(G_L+\lambda I)^{-1}B_L$ with one-hot targets $y_z\in\{0,1\}^c$. \Select{} picks $h^\star=\argmax_{h\in H}s_h(V)$, where $s_h$ is validation accuracy or the negative squared validation loss, and ties are broken by a fixed order. \Augment{} runs $k$ rounds of hard-label self-training from $W_0=W_{h^\star}$:
\begin{align}
\hat y^{(r)}_u&=\argmax\nolimits_j\,(W_{r-1}^\top\phi_u)_j,\nonumber\\
W_r&=\bar A^{-1}\bigl(B_L+w\textstyle\sum_{u\in U}\phi_u e_{\hat y^{(r)}_u}^\top\bigr),\label{eq:st}\\
\bar A&=G_L+wG_U+\lambda_{h^\star}I .\nonumber
\end{align}
The roles of the points are fixed: a deleted point leaves its set and nothing is re-split. Every $\lambda_h>0$, and if $V\setminus F=\emptyset$ all scores tie and the fixed order decides. The normalised margin of $u$ under $W$ is
\[\mu_W(u)=\frac{\text{top-1}-\text{top-2 of }W^\top\phi_u}{\|\phi_u\|}\]
($\infty$ if $\phi_u=0$, whose label the tie order fixes). A stage is a \emph{data touchpoint} if its output depends on $D$. A data-free \Write{} is not one. Standardising or whitening the features also has additive statistics, but it changes every feature of every point, so it lies outside \cref{thm:pipeline}.

\subsection{Exact deletion through the pipeline}

\paragraph{What is stored.} The \emph{compact state} $T(D)$ holds the sums $G_L,B_L,G_U,G_V,B_V$; the label sums $B_U^{(r)}=\sum_{u\in U}\phi_ue_{\hat y_u^{(r)}}^\top$ for $r\le k$; the weights $W_h$ ($h\in H$) and $W_r$ ($r\le k$); and the margin bounds $\tilde m_r$ of part (c) below. This is $O(d^2+(|H|+k)dc)$ numbers, independent of $n$. The \emph{full state} $T^+(D)$ adds the validation features and labels, one correctness bit and margin per validation point and $h$, and every pool feature with its pseudo-label and margin per round. This is $O((|U|+|V|)(d+|H|+k))$ numbers. Like any cache of features, these stores hold per-point data, so a deletion must also purge the requester's entries from them.

\begin{customthm}{\ref*{thm:pipeline-inf}}[Exact deletion through Write--Fit--Select--Augment]\label{thm:pipeline}
Let $\phi$ be data-free and let every tie-break be fixed. For a request $F\subseteq D$, write $\Delta_h=W'_h-W_h$ and $\Delta_r=W'_r-W_r$ for the changes of the stored weights.
\begin{enumerate}
\item[(a)] \emph{Fit.} The downdated solution $W'_h=(G_L-G_{F\cap L}+\lambda_hI)^{-1}(B_L-B_{F\cap L})$ equals the ridge fit on $L\setminus F$. With one eigendecomposition the whole list costs $O(|F|d^2+d^3+|H|d^2c)$.
\item[(b)] \emph{Select.}
\begin{itemize}
\item \emph{Squared validation loss.} The score $s_h=-\tr(W_h^\top G_VW_h)+2\tr(W_h^\top B_V)-\sum_{v}\|y_v\|^2$ is a function of stored sums (the last term equals $|V|$ and cancels in the argmax). Re-selection with $W'_h$ and the downdated $G_{V'},B_{V'}$ is exact, at cost $O(|H|d^2c)$, independent of $|V|$.
\item \emph{0-1 validation accuracy.} A validation point $v$ can change its prediction under $W_h\mapsto W'_h$ only if $\mu_{W_h}(v)\le\sqrt2\|\Delta_h\|$, where
\begin{equation}\label{eq:deltah}
\|\Delta_h\|\le\delta_h:=\frac{\sum_{z\in F\cap L}\|\phi_z\|\,\|W_h^\top\phi_z-y_z\|}{\lambda_{\min}(G_{L\setminus F})+\lambda_h}
\end{equation}
(or the exact $\|\Delta_h\|$ once (a) has run). Let $V'=V\setminus F$, let $a_h$ be the stored accuracy of $W_h$ on $V'$, and let $c_h=\#\{v\in V':\mu_{W_h}(v)\le \sqrt2\delta_h\}$ count the uncertain points.
\item \emph{Certificate.} If $a_{h^\star}-c_{h^\star}/|V'|>\max_{h\ne h^\star}\bigl(a_h+c_h/|V'|\bigr)$, then $h^\star$ is unchanged.
\item \emph{Exact rule.} Otherwise, re-scoring only the $c_h$ uncertain points of each $h$ against their stored labels is exact.
\end{itemize}
\item[(c)] \emph{Augment.} The system matrix $\bar A'=\bar A-G_{F\cap L}-wG_{F\cap U}$ (plus $(\lambda_{h'}-\lambda_{h^\star})I$ if the winner changes to $h'$) does not depend on the pseudo-labels, so one factorisation serves all rounds.
\begin{itemize}
\item \emph{Worst-case certificate.} Suppose $h^\star$ is unchanged. Let $\delta^{(0)}=\delta_{h^\star}$ and, for $r\ge1$,
\begin{equation}\label{eq:deltar}
\delta^{(r)}=\frac{\sum_{z\in F\cap(L\cup U)}w_z\|\phi_z\|\,\|W_r^\top\phi_z-t^{(r)}_z\|}{\max(\lambda_{h^\star},\,\beta)},\qquad \beta=\lambda_{\min}(\bar A)-\textstyle\sum_{z\in F\cap(L\cup U)}w_z\|\phi_z\|^2,
\end{equation}
with $w_z=1$, $t^{(r)}_z=y_z$ for $z\in L$, and $w_z=w$, $t^{(r)}_z=e_{\hat y^{(r)}_z}$ for $z\in U$. Let $\tilde m_{r}\le\min_{u\in U}(\text{top-1}-\text{top-2 of }W_{r}^\top\phi_u)$ be stored lower bounds. If $\tilde m_{r-1}>\sqrt2R\,\delta^{(r-1)}$ for every round $r=1,\dots,k$, then no pseudo-label changes. Deletion then costs $O(|F|d^2+d^3+kd^2c+k|F|dc)$, independent of $n$, $|U|$ and $|V|$, and $\tilde m_{r-1}-\sqrt2R\,\delta^{(r-1)}$ are valid bounds for the next request.
\item \emph{Per-point rule (exact).} In general, go round by round: compute $W'_{r-1}$ exactly, relabel exactly the points with $\mu_{W_{r-1}}(u)\le\sqrt2\|\Delta_{r-1}\|$, and apply rank-one corrections to $B_U^{(r)}$ for those that flip. This reproduces $A(D\setminus F)$ exactly, also when $h^\star$ changes.
\end{itemize}
\item[(d)] \emph{Consequence.} With the full state $T^+(D)$, the pipeline admits an exact deletion operation in the sense of \citet{ginart2019making} for every $F$, with equality rather than equality in distribution. It is $(0,0)$-unlearning in the sense of \citet{sekhari2021remember}, with $|T^+(D)|=O\bigl(d^2+(|H|+k)dc+(|U|+|V|)(d+|H|+k)\bigr)$. With squared-loss selection, a request whose worst-case certificates pass reads only $F$, the published model and the compact $T(D)$.
\end{enumerate}
\end{customthm}

In practice the worst-case certificate in (c) takes a minimum over $10^4$--$10^5$ points, so it is near zero and rarely passes (\cref{sec:x2}); exactness then comes from the per-point rule.

\begin{proposition}[Deterministic pipelines are safe under adaptive requests]\label{prop:adaptive}
If $A$ and $R$ are deterministic and $R(D,A(D),F)=A(D\setminus F)$ for every $F$, then for any adaptive sequence $F_1,F_2,\dots$ (where $F_t$ may depend on all previously published models and on external randomness), the $t$-th published model equals $A(D\setminus\cup_{s\le t}F_s)$. Its conditional law given the transcript is the same point mass that retraining would produce.
\end{proposition}

\Cref{prop:adaptive} states pathwise what \citet[Sec.~1.2]{gupta2021adaptive} note for deterministic methods and for randomness drawn after the request. Features built from $m$ centres sampled from $D$ do have reusable randomness; \cref{thm:centres} in \cref{app:proofs} makes the failure precise. A sampled pipeline must either redraw all $m$ centres at every deletion, which is safe but costs a retrain, or protect its centres, for example with differential privacy \citep{gupta2021adaptive}.

\begin{remark}[Privacy and numerics]\label{rem:privacy}
All three points are known (details in \cref{app:proofs}).
\begin{enumerate}
\item[(i)] An exact mechanism gives an attacker who sees only the post-deletion model exactly its advantage against retraining. It does not protect against an attacker who also sees the model before: $W'-W$ in \eqref{eq:smw-main} points at the forgotten features \citep{chen2021when} (\cref{sec:x3}).
\item[(ii)] A stage left approximate, within a worst-case weight distance $\Delta_w$, becomes $(\varepsilon,\delta)$-indistinguishable only if the same Gaussian noise is added to the release and to the retraining reference \citep{guo2020certified,dwork2014algorithmic}.
\item[(iii)] \Cref{thm:pipeline} holds in exact arithmetic. A float64 near-tie could flip a label, which we do not rule out, but all 2,400 requests of \cref{sec:x2} matched. Chained Sherman--Morrison downdates can compound error through the condition number of $A$ \citep{higham2002accuracy,shang2026error}; in float64 this barely matters (\cref{sec:x1}).
\end{enumerate}
\end{remark}

\subsection{Evaluation of deep forgetting}

\begin{customprop}{\ref*{prop:padding-inf}}[Output metrics cannot certify deep forgetting]\label{prop:padding}
Let $\mathcal M(f)=\Psi(f|_{\mathcal X})$ be any functional of the input-to-logit map: forget, retain or test accuracy, the advantage of any attacker with query access to the logits (including U-LiRA), or any divergence to the retrained model's outputs. Let the model class be closed under concatenating an admissible hidden-feature branch with zero read-out weights. Let $f$ be the original model with penultimate features $h_f$, and let $g(x)=V_gh_g(x)+b_g$ be any model.
\begin{enumerate}
\item[(i)] The padded model $\tilde g$ with features $[h_g;h_f]$ and read-out $[V_g,\,0]$ satisfies $\tilde g|_{\mathcal X}=g|_{\mathcal X}$, hence $\mathcal M(\tilde g)=\mathcal M(g)$.
\item[(ii)] Yet the best linear probe on the penultimate features of $\tilde g$ fits any target, in sample or in population, at least as well as the best probe on $h_f$.
\item[(iii)] In particular, deleting the forgotten class's output unit (removing it from the argmax) is output-optimal: forget accuracy is $0$, and retain accuracy can only rise. As the hidden features are unchanged, it leaves every probe of every hidden layer exactly at its original value.
\end{enumerate}
\end{customprop}

The corollary (iii) also covers any edit of the logits alone, such as linear filtration \citep{baumhauer2022machine}. Without padding, outputs can suffice: a ridge head on known features is fixed by its outputs on $d$ independent inputs.

\begin{definition}[ProbeGap]\label{def:probegap}
Fix a probe family $\mathcal P$ in advance and a layer $\ell$. Let $\mathcal R_\ell(f)$ be the probe score of $f$ at $\ell$ (least-squares $R^2$ of the forget indicator, relearned forget accuracy, \dots), and let $\bar{\mathcal R}_\ell$ be its mean over a band of retrained seeds. Then $\mathrm{ProbeGap}_\ell(f)=\bigl|\mathcal R_\ell(f)-\bar{\mathcal R}_\ell\bigr|$, reported with the band's range.
\end{definition}

For deterministic exact deletion, ProbeGap equals that of retraining by identity. For a method that is exact in distribution, it is distributed like the gap of a fresh retrained seed. A small ProbeGap is a diagnostic for the fixed $\mathcal P$, not a certificate against all probes.

\paragraph{The projection edit.} At a layer with activations $x\in\R^p$, let the retain set have mean $\mu_R$ and covariance $\Sigma_R\succ0$, and let the forget set have mean $\mu_F$. Whiten with the retain moments: $z=\Sigma_R^{-1/2}(x-\mu_R)$, $\delta=\Sigma_R^{-1/2}(\mu_F-\mu_R)$, and $\tilde M_F=\E_F[zz^\top]=\tilde\Sigma_F+\delta\delta^\top$. Let $\gamma_1\ge\dots\ge\gamma_p$ be the eigenvalues of $\tilde M_F$. For $1\le k<p$, proj$(k)$ replaces $x$ by $x'=\mu_R+\Sigma_R^{1/2}Q\Sigma_R^{-1/2}(x-\mu_R)$, where $Q$ removes the top-$k$ eigenvectors of $\tilde M_F$. As a weight edit it is $W\leftarrow W\Sigma_R^{1/2}Q\Sigma_R^{-1/2}$, with the bias corrected at $\mu_R$. Part (b) below needs this unregularised whitening: with $\Sigma_R+\alpha I$ in its place the bound can fail (\cref{app:proofs}). Our theorem checks therefore whiten by the pseudo-inverse on the range of $\Sigma_R$; the deployed edits add $\alpha=10^{-2}$ times the mean diagonal and carry no bound.

\begin{customthm}{\ref*{thm:probe-inf}}[What proj$(k)$ guarantees]\label{thm:probe}
\begin{enumerate}
\item[(a)] \emph{Optimality.} Among rank-$k$ removals with $V^\top\Sigma_RV=I_k$, proj$(k)$ maximises the removed forget energy $\tr(V^\top M_FV)=\sum_{i\le k}\gamma_i$, and it removes exactly $k$ whitened units of retain energy.
\item[(b)] \emph{Probe bound.} Let the probe population be a mixture with forget fraction $0<\pi<1$ and finite second moments, and let $R^2$ be the population least-squares $R^2$ of the forget indicator from the layer's activations. Then $R^2=\pi(1-\pi)\,\Delta\mu^\top\Sigma_x^{-1}\Delta\mu\le t/(1+t)$ with $t=\pi\|\delta\|^2$, with equality if $\Sigma_F=0$. After the edit,
\[R^{2\prime}\;\le\;\frac{t_Q}{1+t_Q}\;\le\;\frac{\pi\gamma_{k+1}}{1+\pi\gamma_{k+1}}\;<\;1,\qquad t_Q=\pi\|Q\delta\|^2 .\]
Removing the single direction $\delta/\|\delta\|$ gives $R^{2\prime}=0$, which is linear guarding in the sense of LEACE \citep{belrose2023leace}. The same holds in-sample, with empirical moments in place of population ones.
\item[(c)] \emph{Random forget sets.} Suppose $F$ consists of $m\ge p$ i.i.d.\ draws from $\mathcal N(\mu_R,\Sigma_R)$ and the whitening uses population moments. Then the eigenvectors of $\tilde M_F$ are Haar-distributed ($\E(u^\top v_1)^2=1/p$ for every unit $u$); $\gamma_1-\gamma_p\to4\sqrt{p/m}$ almost surely as $p,m\to\infty$ with $p/m$ fixed \citep{marchenko1967distribution,bai1993limit}; and $\E\,\pi\|\delta\|^2=\pi p/m$. So the edit removes $k$ directions unrelated to the population, at an expected cost of $k$ of $p$ whitened units of retain variance.
\item[(d)] \emph{Small forget sets.} If $m\le k<p$, every forget point is mapped exactly to $\mu_R$, while under population whitening a fresh retain point keeps $\E\|Qz\|^2=p-k$. The forget set becomes distinguishable from data never seen.
\end{enumerate}
\end{customthm}

Part (b) is about squared loss, not 0-1 accuracy: a class at $\pm a$ against one at $0$ has equal means, yet a threshold reaches $75\%$. It holds at the edited layer only, and an $R^2$ below the retrained model's value is over-unlearning, not success. Real requests are often self-selected and may share attributes such as age or region; then (b) applies, and the edit removes that direction for every retained person who shares them.

\section{Proofs}\label{app:proofs}

Throughout, $\|\cdot\|$ is the Euclidean norm on vectors and the operator norm on matrices. We give full proofs, including those of standard facts; where a step is known, we cite its source in the main text, in \cref{app:formal} and in \cref{tab:novelty}.

\subsection{The downdate identity}
This standard identity is the Sherman--Morrison--Woodbury formula \citep{sherman1950adjustment,higham2002accuracy}.
\begin{lemma}\label{lem:smw}
Let $AW=B$ with $A=A_0+\sum_{z\in D}w_z\phi_z\phi_z^\top$ and $B=B_0+\sum_{z\in D}w_z\phi_zt_z^\top$, and let $A'=A-\sum_{z\in F}w_z\phi_z\phi_z^\top\succ0$, $B'=B-\sum_{z\in F}w_z\phi_zt_z^\top$, $W'=A'^{-1}B'$. Then $W'-W=A'^{-1}\sum_{z\in F}w_z\phi_z(\phi_z^\top W-t_z^\top)$ and
$\|W'-W\|\le\|A'^{-1}\|\sum_{z\in F}w_z\|\phi_z\|\,\|W^\top\phi_z-t_z\|$.
For any two matrices $W,W'$ and every $x$, $\|(W'-W)^\top\phi(x)\|_\infty\le\|\phi(x)\|\,\|W'-W\|$, and if $\arg\max_j(W'^\top\phi(x))_j\ne\arg\max_j(W^\top\phi(x))_j$ then the top-1/top-2 margin of $W^\top\phi(x)$ is at most $\sqrt2\|\phi(x)\|\,\|W'-W\|$. If the argmax does not change, the margin of $W'^\top\phi(x)$ is at least that of $W^\top\phi(x)$ minus $\sqrt2\|\phi(x)\|\,\|W'-W\|$.
\end{lemma}
\begin{proof}
$A'W=AW-\sum_Fw_z\phi_z\phi_z^\top W=B-\sum_Fw_z\phi_z\phi_z^\top W$ and $A'W'=B-\sum_Fw_z\phi_zt_z^\top$; subtracting gives $A'(W'-W)=\sum_Fw_z\phi_z(\phi_z^\top W-t_z^\top)$. The norm bound is the triangle inequality with $\|\phi r^\top\|=\|\phi\|\|r\|$. For the logits, $|e_j^\top(W'-W)^\top\phi|\le\|(W'-W)e_j\|\|\phi\|\le\|W'-W\|\|\phi\|$. For two classes $j\ne j'$, $(s'_{j'}-s_{j'})-(s'_j-s_j)=\phi^\top(W'-W)(e_{j'}-e_j)$ with $\|e_{j'}-e_j\|=\sqrt2$, so its size is at most $\sqrt2\|\phi\|\|W'-W\|$. If the argmax moves from $j$ to $j'$, then $s_j-s_{j'}\ge$ margin before and $s'_{j'}-s'_j\ge0$ after, so the margin is at most $(s'_{j'}-s_{j'})-(s'_j-s_j)\le\sqrt2\|\phi\|\|W'-W\|$. If it stays at $j$, each gap $s'_j-s'_{j'}$ is at least $s_j-s_{j'}-\sqrt2\|\phi\|\|W'-W\|$. The constant $\sqrt2$ is attained for $c=2$.
\end{proof}

\subsection{Proof of Theorem~\ref{thm:pipeline}}
\emph{(a)} Apply \cref{lem:smw} with $A_0=\lambda_hI$, $w_z=1$, $t_z=y_z$: $W'_h$ is by definition the ridge solution on $L\setminus F$. With $G=U\Lambda U^\top$, $W_\lambda=U(\Lambda+\lambda I)^{-1}U^\top B$ costs $O(d^2c)$ per grid point after the $O(d^3)$ decomposition; the downdate costs $O(|F|d^2)$.

\emph{(b)} Squared loss: $\sum_{v\in V}\|W^\top\phi_v-y_v\|^2=\tr(W^\top G_VW)-2\tr(W^\top B_V)+\sum_v\|y_v\|^2=-s_h$, and $(G_V,B_V,\sum\|y_v\|^2)$ are downdated by $F\cap V$ in $O(|F|d^2)$; for one-hot targets $\sum_v\|y_v\|^2=|V|$ is the same for every $h$, so it matters only for the value of the loss. For 0-1 accuracy, first remove $F\cap V$ from the stored correctness bits (exact); $a_h$ is then the accuracy on $V'=V\setminus F$. By \cref{lem:smw}, with $\|A'^{-1}_h\|=1/(\lambda_{\min}(G_{L\setminus F})+\lambda_h)$, where $\lambda_{\min}$ comes from the eigendecomposition of (a), $\|\Delta_h\|\le\delta_h$, and a point $v$ can change its prediction only if $\mu_{W_h}(v)\le\sqrt2\|\Delta_h\|\le\sqrt2\delta_h$ ($\mu=\infty$ for $\phi_v=0$, whose prediction is fixed by the tie order). Hence the new accuracy $a'_h$ lies in $[a_h-c_h/|V'|,\,a_h+c_h/|V'|]$, and under the stated strict inequality $a'_{h^\star}>a'_h$ for all $h\ne h^\star$, whatever the tie-break. Re-scoring the uncertain points of each $h$ gives $a'_h$ exactly, since all other points provably keep their prediction. Re-scoring needs the stored label $y_v$, not only the correctness bit: for $c\ge3$, a point that was wrong and changes its prediction may become right or stay wrong. Since (a) computes $W'_h$ anyway, the same argument holds with the exact $\|\Delta_h\|$ in place of $\delta_h$.

\emph{(c)} Write $t^{(r)}_z=y_z$ for $z\in L$ and $t^{(r)}_z=e_{\hat y^{(r)}_z}$ for $z\in U$, and $W'_r$ for the weights of the rerun on $D\setminus F$. Assume $h^\star$ is unchanged. We show by induction on $r$ that the rerun assigns every $u\in U\setminus F$ the stored label $\hat y^{(r)}_u$. Round $1$ uses $W_0=W_{h^\star}$ and $W'_0=W'_{h^\star}$, with $\|W'_0-W_0\|\le\delta^{(0)}=\delta_{h^\star}$ by (b). If $u$ flipped, \cref{lem:smw} would give $\tilde m_0\le$ margin of $u\le\sqrt2\|\phi_u\|\delta^{(0)}\le\sqrt2R\delta^{(0)}$, contradicting the certificate. (A minimum over $U\setminus F$ cannot be recomputed from $O(k)$ numbers once its minimiser is deleted; the stored bound over all of $U$ is a valid lower bound for it.) Suppose the labels of rounds $1,\dots,r$ agree. Then the rerun's round-$r$ system is $\bar A'W'_r=B_L-B_{F\cap L}+w(B_U^{(r)}-\sum_{z\in F\cap U}\phi_ze_{\hat y^{(r)}_z}^\top)$, which is exactly the downdate of the stored round-$r$ system by the rows of $F$ with their round-$r$ targets. \Cref{lem:smw} with $A_0=\lambda_{h^\star}I$ gives $\|W'_r-W_r\|\le\|\bar A'^{-1}\|\sum_{F\cap(L\cup U)}w_z\|\phi_z\|\|W_r^\top\phi_z-t^{(r)}_z\|$ (validation points do not enter $\bar A$, so $w_z=0$ for $z\in V$), and Weyl's inequality gives $\lambda_{\min}(\bar A')\ge\max(\lambda_{h^\star},\lambda_{\min}(\bar A)-\sum_{F\cap(L\cup U)}w_z\|\phi_z\|^2)$, so $\|W'_r-W_r\|\le\delta^{(r)}$. The round-$(r+1)$ labels come from $W_r$ and $W'_r$, and the same margin argument with $\tilde m_r>\sqrt2R\delta^{(r)}$ rules out a flip. Hence all labels agree, the round-specific label sums and fitted weights equal the rerun's, and the outputs coincide. Since no label changed, \cref{lem:smw} also shows that every margin under $W'_r$ is at least its margin under $W_r$ minus $\sqrt2R\delta^{(r)}$, so $\tilde m_r-\sqrt2R\delta^{(r)}$ is a valid stored bound for the next request. The cost consists of the downdate ($O(|F|d^2)$), one factorisation and one $\lambda_{\min}$ ($O(d^3)$), $k$ solves ($O(kd^2c)$) and the residuals of the forgotten points ($O(k|F|dc)$, absorbed by the other terms when $|F|\le d$ or $kc\le d$); none depends on $|U|$, $|V|$ or $n$.

For the per-point certificate, let $h'$ be the new winner (possibly $h^\star$) and factorise $\bar A'=\bar A-G_{F\cap L}-wG_{F\cap U}+(\lambda_{h'}-\lambda_{h^\star})I$ once. Start from $W'_0=W'_{h'}$, computed exactly in (a). In round $r$, the margin part of \cref{lem:smw} holds for any two matrices, so a point with $\mu_{W_{r-1}}(u)>\sqrt2\|W'_{r-1}-W_{r-1}\|$ keeps its stored label; every other point is relabelled with $W'_{r-1}$, and each flip $\hat y\to\hat y'$ is applied as $B_U^{(r)}\leftarrow B_U^{(r)}+\phi_u(e_{\hat y'}-e_{\hat y})^\top$. After removing the rows of $F$ with their round-$r$ targets, the statistics of round $r$ equal the rerun's, and $W'_r=\bar A'^{-1}(B_L-B_{F\cap L}+wB_U^{(r)\prime})$ is the rerun's round-$r$ solution. Induction over $r=1,\dots,k$ finishes the proof; when $h^\star$ changes, only the margins test more points. In a sequence of requests, a stored margin $\mu$ that was not recomputed is replaced by the valid lower bound $\mu-\sqrt2\|W'-W\|$. The published output then equals the rerun's, but these stored bounds depend on the history of requests, so the internal state need not equal $T^+(D\setminus F)$.

\emph{(d)} By (a)--(c) the output equals $A(D\setminus F)$ for every $F$; with deterministic $A$ this is Definition~3.1 of \citet{ginart2019making} with equality. The definition of \citet[Def.~2]{sekhari2021remember} fixes the state before the request, for all requests. With $T^+(D)$, which holds the features, labels, correctness bits and margins of $V$ and $U$ (size $O((|U|+|V|)(d+|H|+k))$) besides the compact part, the mechanism is $(0,0)$-unlearning for every $F$. The compact part has size $O(d^2)$ for $(G_L,G_U,G_V)$ plus $O((|H|+k)dc)$ for $B_V$, the weights and the label sums, plus $O(k)$ stored minima. With squared-loss selection and passing worst-case certificates, the mechanism reads only $F$ (whose features it computes with the data-free $\phi$), the published model and this compact part; that is a statement about access cost on such requests, not a smaller state. \qed

\subsection{Proof of Proposition~\ref{prop:adaptive}}
By induction on $t$: the model published after request $t-1$ is $A(D_{t-1})$ with $D_{t-1}=D\setminus\cup_{s<t}F_s$, a deterministic function of $D$ and the requests. Whatever $F_t$ the requester chooses, exactness gives $R(D_{t-1},A(D_{t-1}),F_t)=A(D_{t-1}\setminus F_t)$. Conditionally on the whole transcript the output is therefore the point mass at $A(D\setminus\cup_{s\le t}F_s)$, which is also what retraining on the same remaining data outputs. The same argument covers a randomised $A$ when every release retrains with randomness drawn after the request and independent of the transcript: given the transcript, the release then has exactly the law of $A(D\setminus\cup_{s\le t}F_s)$ \citep[Sec.~1.2]{gupta2021adaptive}. \qed

\subsection{Sampled centres}
\begin{theorem}[Sampled centres: exact in distribution, not adaptive-safe]\label{thm:centres}
Let the \Write{} stage use a centre set $C$ drawn uniformly among $m$-subsets of $D$ (Nystr\"om landmarks or patch filters copied from training points), and everything downstream be a deterministic function of $(C,D)$. On a request $F$ with $|D\setminus F|\ge m$, keep $C\setminus F$ and replace each removed centre by a uniform draw without replacement from $(D\setminus F)\setminus C$; then recompute the affected feature columns and statistics.
\begin{enumerate}
\item[(a)] For any $F$ fixed in advance, the new centre set is uniform over $m$-subsets of $D\setminus F$, so the mechanism is an exact deletion operation in distribution.
\item[(b)] If the published model reveals $C$ (a white-box release), there is an adaptive requester that sees it, deletes $|F|$ non-centres, and for whom the post-deletion centre set equals the published one with probability $1$, whereas retraining on $D\setminus F$ gives that set with probability $\binom{n-|F|}{m}^{-1}$. The total-variation distance between the two conditional laws is $1-\binom{n-|F|}{m}^{-1}$.
\end{enumerate}
\end{theorem}

\begin{corollary}[Cost of resampling]\label{cor:centrecost}
Suppose (i) the raw inputs and the feature column of every centre on every point are stored, (ii) the features are unnormalised per-centre features $\phi_j(x)=\kappa(x,c_j)$ (for whitened Nystr\"om features $K_{CC}^{-1/2}k_C(x)$, one centre swap changes all $m$ coordinates; storing the unwhitened moments and applying $K_{CC}^{-1/2}$ afterwards adds $O(m^3)$), and (iii) a single $c$-class ridge head is fitted on top (selection and self-training downstream may need every pool point rescored). Then the expected cost of a request under \cref{thm:centres}(a) is $O(|F|\,m\,(d_{\mathrm{in}}+m+c))$ plus one $O(m^3+m^2c)$ solve, against $O(n\,m(d_{\mathrm{in}}+m+c)+m^3+m^2c)$ for retraining.
\end{corollary}

\paragraph{Proof.}
\emph{\Cref{thm:centres}(a).} Let $C$ be uniform over $m$-subsets of $D$, $|D|=n$, and condition on $j=|C\cap F|$. By exchangeability of the points of $D\setminus F$, given $j$ the set $C\setminus F$ is uniform over $(m-j)$-subsets of $D\setminus F$. The rule adds $j$ points drawn uniformly without replacement from the remaining $|D\setminus F|-(m-j)$ points of $D\setminus F$. For a fixed $m$-subset $S\subseteq D\setminus F$, the probability of producing $S$ is $\binom{m}{m-j}\big/\bigl[\binom{n-|F|}{m-j}\binom{n-|F|-m+j}{j}\bigr]=\binom{n-|F|}{m}^{-1}$, the same for every $S$. So the conditional law is uniform for each $j$, and so is the mixture. Everything downstream is a deterministic function of $(C,D\setminus F)$, so the output has the law of $A(D\setminus F)$.

\emph{\Cref{cor:centrecost}.} A forgotten point is a centre with probability $m/n$, so $\E|C\cap F|=m|F|/n$. Each replacement needs the new centre's feature column on the $\le n$ remaining raw inputs ($O(nd_{\mathrm{in}})$) and its inner products with the $m-1$ stored columns, a rank-two update of the Gram matrix ($O(nm)$; these cross-terms cannot be recovered from the Gram matrix alone, hence (i)), and the new row of $B=\Phi^\top Y$ ($O(nc)$); under (ii) no other column changes, giving $O(|F|m(d_{\mathrm{in}}+m+c))$ in expectation; the rows of $F$ are downdated in $O(|F|m(d_{\mathrm{in}}+m+c))$. Retraining recomputes all $nm$ kernel entries, the Gram matrix, $B$ and the solve. Under (iii) the head is re-solved once in $O(m^3+m^2c)$; a downstream \Select{} or \Augment{} would see the new centre's feature change on every point, so \cref{thm:pipeline}'s certificates do not apply to it.

\emph{\Cref{thm:centres}(b).} The published model reveals $C$ (copied patches are the filters; Nystr\"om landmarks determine the feature map). Let the requester delete $|F|\le n-2m$ points of $D\setminus C$. No centre is removed, so the rule keeps $C$ with probability $1$. Retraining on $D\setminus F\supseteq C$ draws a uniform $m$-subset, which equals $C$ with probability $\binom{n-|F|}{m}^{-1}$. The total-variation distance between a point mass and a law that gives it probability $q$ is $1-q$. Redrawing all $m$ centres uniformly from $D\setminus F$ at every request avoids this, since the new centres are then independent of the transcript, but it costs a retrain. \qed

\subsection{Cost and state of exact ridge deletion}
\begin{theorem}[Cost and state of exact ridge deletion]\label{thm:cost}
\begin{enumerate}
\item[(a)] Downdating the summed statistics and re-solving costs $O(|F|d^2+d^3/3+d^2c)$; Woodbury with a stored inverse costs $O(|F|d^2+|F|^2d+|F|^3+d^2c)$; retraining costs $O(Nd^2+d^3/3+d^2c)$ for $N$ training rows. The speed-up of the refactorisation path is therefore $(Nd^2+d^3/3)/(|F|d^2+d^3/3)$ when the $O(d^2c)$ solves are negligible; for $N\ge d/3$ this is within a factor $2$ of $\min(N/|F|,\,3N/d)$, not $N/|F|$ (for $N\ll d$ it tends to $1$).
\item[(b)] If the state is the pair $(G,B)$ of summed statistics, an exact algorithm must in the worst case read all $|F|d$ feature coordinates of the request and write $dc$ outputs.
\item[(c)] Fix $\lambda>0$, $c\ge1$ and $b=\Omega(\log d+|\log\lambda|)$. Let the published model and the mechanism's outputs be rounded to $b'$-bit fixed-point numbers, and the state be a bit string; the mechanism sees only the request, the published model and the state, with no other access to the retained data. There is a family of $2^{b\,d(d+1)/2}$ datasets in $\R^d$ such that, if $b'\ge b+C_{d,\lambda}$ with $C_{d,\lambda}=O(\log d+|\log\lambda|)$, any mechanism that outputs $W(D\setminus\{z\})$ for every single point $z\in D$ must use at least $b\,d(d+1)/2-b'\,dc$ bits of state. For $c=o(d)$ and $b'=O(b)$ this is $\Omega(d^2b)$ bits; the $O(d^2)$ sufficient statistics take $O(d^2b'')$ bits at $b''$ bits per entry, so they are optimal up to constants whenever $b''=O(b)$.
\end{enumerate}
\end{theorem}
\paragraph{Proof.}
\emph{(a)} Forming $G_F$ costs $|F|d^2$, the Cholesky factorisation $d^3/3$ and the two triangular solves $2d^2c$. Woodbury with $A^{-1}$ stored: $(A-\Phi_F^\top\Phi_F)^{-1}=A^{-1}+A^{-1}\Phi_F^\top(I-\Phi_FA^{-1}\Phi_F^\top)^{-1}\Phi_FA^{-1}$ costs $O(|F|d^2+|F|^2d+|F|^3)$. Retraining forms $G$ from $N$ rows in $Nd^2$. The ratio of the dominant terms is $(Nd^2+d^3/3)/(|F|d^2+d^3/3)$; for $|F|\ll d$ it equals $3N/d+1$, and for $|F|\gg d$ it equals $N/|F|$ up to lower-order terms. In general, with $a=N$, $b=|F|\le a$ and $e=d/3$, $(a+e)/(b+e)$ lies between $a/(2\max(b,e))$ and $(a+e)/\max(b,e)$, which is at most $2a/\max(b,e)$ when $a\ge e$; so for $N\ge d/3$ the ratio is within a factor $2$ of $\min(N/|F|,3N/d)$.

\emph{(b)} With state $(G,B)$ the exact output is $W'=(G-\sum_F\phi_z\phi_z^\top+\lambda I)^{-1}(B-\sum_F\phi_zy_z^\top)$, a rational function of every coordinate $\phi_{z,i}$ that is non-constant for generic $(G,B)$ (differentiating $A'W'=B'$, its derivative in $\phi_{z,i}$ is $A'^{-1}(e_i(\phi_z^\top W'-y_z^\top)+\phi_ze_i^\top W')$, nonzero for generic data). An algorithm that never reads $\phi_{z,i}$ returns the same output when $\phi_{z,i}$ is changed, so it errs on some input. Each of the $dc$ outputs can be changed independently by the data, so each must be written.

\emph{(c)} Take $c\ge2$, $0<\varepsilon^2<\lambda/\sqrt d$ and $\alpha>0$. Let $\mathcal E_b$ be the set of symmetric matrices whose entries on and above the diagonal lie on a grid of $2^b$ values in $[-\alpha/(2d),\alpha/(2d)]$, and for $E\in\mathcal E_b$ let $M=\alpha I+E$, which is positive definite because $\|E\|\le\alpha/2$. Dataset $D_E$ consists of (i) $d$ probe points $\phi=\varepsilon e_i$ with target $e_1$, and (ii) $d$ points $\sqrt{\sigma_j}u_j$ with target $e_2$, where $M=\sum_j\sigma_ju_ju_j^\top$. Then $A=M+\varepsilon^2I+\lambda I$ and the first column of $W=A^{-1}B$ is $\varepsilon A^{-1}\mathbf 1$. The class-1 residual of probe $i$ is $r_i=\varepsilon^2(A^{-1}\mathbf 1)_i-1$, and $|\varepsilon^2(A^{-1}\mathbf 1)_i|\le\varepsilon^2\sqrt d/\lambda<1$, so $r_i\ne0$. Leverage $h_i=\varepsilon^2a_{ii}<1$, where $a_i=A^{-1}e_i$. The Sherman--Morrison form of \cref{lem:smw} for a single point gives $W(D\setminus z_i)-W(D)=A^{-1}\phi_i(\phi_i^\top W-y_i^\top)/(1-h_i)$, whose first column is $s_i\,\varepsilon r_i$ with $s_i=a_i/(1-\varepsilon^2a_{ii})$. A decoder that knows the published $W$ (hence $r_i=\varepsilon W_{i1}-1$) and the mechanism's output for each probe deletion recovers $s_i$, then $a_{ii}=(s_i)_i/(1+\varepsilon^2(s_i)_i)$, then $a_i=s_i(1-\varepsilon^2a_{ii})$, hence $A^{-1}$, $A$ and $E$. So the map $E\mapsto(\text{published model},\text{state})$ is injective on $\mathcal E_b$, which has $2^{b\,d(d+1)/2}$ elements, and the state must carry at least $b\,d(d+1)/2$ bits minus the $b'dc$ bits of the published model. For $c=1$, where every one-hot target is $1$, replace each point $\sqrt{\sigma_j}u_j$ by the pair $\pm\sqrt{\sigma_j/2}\,u_j$ with target $1$: the pair adds $\sigma_ju_ju_j^\top$ to $G$ and $0$ to $B$, so $A$, $W$ and the decoder are unchanged.

\emph{Precision (sketch).} The decoder above uses exact values. A $b'$-bit fixed-point number here is an integer multiple of $2^{-f}$ in $[-2^{e},2^{e})$; there are $2^{e+f+1}$ of them, so $b'=1+e+f$ (one sign bit), and the mechanism outputs the entries of $W$ rounded to the nearest such number; we take $e=O(\log d+|\log\lambda|)$, enough to hold every entry below, so $f=b'-e-1$. Take $\alpha=\lambda$ and $\varepsilon^2=\lambda/(2\sqrt d)$. Then $\lambda\le\lambda_{\min}(A)\le\|A\|\le3\lambda$, $|r_i|\ge1/2$ and $1-\varepsilon^2a_{ii}\ge1/2$, and every entry of $W$ and of the outputs is bounded by a polynomial in $d$, $\lambda$ and $1/\lambda$. Each decoding step is a rational map whose denominators are bounded away from zero by these bounds, and inverting $A^{-1}$ multiplies errors by at most $2\|A\|^2$. So if $W$ and the outputs are rounded with $f$ fractional bits, the decoded $E$ has entry-wise error at most $2^{-f}P(d,\lambda,1/\lambda)$ for a polynomial $P$. The grid step of $\mathcal E_b$ is at least $\lambda2^{-b}/d$. Hence for $b'\ge b+C_{d,\lambda}$ with $C_{d,\lambda}=O(\log d+|\log\lambda|)$, rounding the decoded $E$ to the grid recovers it exactly, and the counting argument holds with a $b'$-bit published model. Both $b'\ge b+C_{d,\lambda}$ and $b'=O(b)$ can hold only if $b=\Omega(\log d+|\log\lambda|)$, which the statement assumes. The lower bound counts bits; the $O(d^2)$ sufficient statistics take $O(d^2b'')$ bits at $b''$ bits per entry, which matches it up to constants when $b''=O(b)$. \qed

\subsection{Remark~\ref{rem:privacy}}
(i) If the released model is equal (or equal in law) to $A(D\setminus F)$, any statistic of it has the same law under both. (ii) For least squares the Newton removal of \citet[Sec.~3]{guo2020certified} has zero gradient residual. Otherwise, with $\sigma=\Delta_w\sqrt{2\ln(1.25/\delta)}/\varepsilon$ and $\varepsilon<1$, the Gaussian mechanism \citep[Thm.~A.1]{dwork2014algorithmic} applies to the two noisy models $\tilde W_{\mathrm{unl}}+N$ and $W_{\mathrm{retrain}}+N'$ with $N,N'\sim\mathcal N(0,\sigma^2I)$ and $\|\tilde W_{\mathrm{unl}}-W_{\mathrm{retrain}}\|_F\le\Delta_w$. The bound $\Delta_w$ must hold for all $D$ and $F$, so that $\sigma$ does not itself depend on the data. If only the unlearned model is noised, its law has a density while the deterministic reference is a point mass, so the total-variation distance is $1$ and no $\delta<1$ is possible. For an exact mechanism $\Delta_w=0$, and the same noise on both sides keeps it exact in distribution relative to the noisy reference. None of this addresses an observer of the releases before and after a deletion. The observed change $\|W(D)-W(D\setminus F)\|$ is not a sensitivity in the sense of differential privacy, which is a worst-case bound over a specified neighbouring relation. Hiding a deletion from that observer needs a separately specified privacy mechanism: a request-adjacency relation over the whole release transcript, a proven worst-case sensitivity bound for it, and noise calibrated to that bound. We neither establish nor evaluate such a mechanism. Each release above is covered separately; a sequence of public releases needs a composition argument. (iii) is the standard backward-error analysis of summation and Cholesky \citep{higham2002accuracy}; we do not claim a new bound and report measurements instead.

\subsection{Proof of Proposition~\ref{prop:padding}}
For every $x$, $\tilde g(x)=[V_g,0][h_g(x);h_f(x)]+b_g=g(x)$, so $\tilde g|_{\mathcal X}=g|_{\mathcal X}$ and $\Psi$ takes the same value. A linear probe $\beta^\top h_f+\beta_0$ on $h_f$ is the probe $[0;\beta]^\top[h_g;h_f]+\beta_0$ on $\tilde g$'s features, so the best probe risk on $\tilde g$ is at most that on $f$. This compares the best attainable risk (in population, or in sample on the same data); a probe fitted on more features can still score lower on held-out data. For delete-unit, remove unit $c$ from the argmax, i.e.\ set row $c$ of the read-out to $0$ and its bias to $-\infty$ (a bias of $0$ is not enough: if all other logits are negative, class $c$ still wins; our runs use $-10^9$). A forget-class input is never assigned class $c$, so forget accuracy is $0$. A retain input of class $y\ne c$ that was correct stays correct, because removing a losing logit cannot change the argmax; one that was assigned $c$ with $y$ as the runner-up becomes correct. The hidden features are untouched, so every probe of every hidden layer is unchanged. \qed

\subsection{Proof of Theorem~\ref{thm:probe}}
\emph{(a)} Put $\tilde V=\Sigma_R^{1/2}V$, so $\tilde V^\top\tilde V=I_k$ and $\tr(V^\top M_FV)=\tr(\tilde V^\top\tilde M_F\tilde V)$, where $M_F=\E_F[(x-\mu_R)(x-\mu_R)^\top]$. By Ky Fan's maximum principle \citep{fan1949on} the maximum over orthonormal $\tilde V$ is $\sum_{i\le k}\gamma_i$, attained by the top-$k$ eigenvectors. The removed whitened retain energy is $\E_R\|(I-Q)z\|^2=\tr(I-Q)=k$.

\emph{(b)} Let $Z\in\{0,1\}$ indicate the forget component, $\Pr(Z=1)=\pi$. Then $\mathrm{Cov}(x,Z)=\pi(1-\pi)\Delta\mu$ with $\Delta\mu=\mu_F-\mu_R$, $\mathrm{Var}(Z)=\pi(1-\pi)$, and the population least-squares $R^2$ is $\mathrm{Cov}(x,Z)^\top\Sigma_x^{+}\mathrm{Cov}(x,Z)/\mathrm{Var}(Z)=\pi(1-\pi)\Delta\mu^\top\Sigma_x^{+}\Delta\mu$. $R^2$ is invariant under invertible affine maps of $x$, so we work in whitened coordinates $z$, where the retain covariance is $I$ and $\Delta\mu$ becomes $\delta$. The law of total covariance gives $\Sigma_z=(1-\pi)I+\pi\tilde\Sigma_F+\pi(1-\pi)\delta\delta^\top\succeq(1-\pi)(I+\pi\delta\delta^\top)$, so $\Sigma_z^{-1}\preceq(I+\pi\delta\delta^\top)^{-1}/(1-\pi)$, and by Sherman--Morrison
\[R^2\le\pi\,\delta^\top(I+\pi\delta\delta^\top)^{-1}\delta=\frac{\pi\|\delta\|^2}{1+\pi\|\delta\|^2},\]
with equality if $\tilde\Sigma_F=0$. $R^2$ can only decrease under non-invertible affine maps, so after the edit we may work with $z'=Qz$. Its retain covariance is $Q$, and its covariance under the mixture is $\Sigma_{z'}=(1-\pi)Q+\pi Q\tilde\Sigma_FQ+\pi(1-\pi)(Q\delta)(Q\delta)^\top\succeq(1-\pi)(Q+\pi(Q\delta)(Q\delta)^\top)$; both sides have range $\mathrm{range}(Q)$, and for positive semidefinite matrices with a common range the order reverses under the pseudo-inverse. On $\mathrm{range}(Q)$, $Q$ acts as the identity and $Q\delta\in\mathrm{range}(Q)$, so the same Sherman--Morrison step gives $R^{2\prime}=\pi(1-\pi)(Q\delta)^\top\Sigma_{z'}^{+}(Q\delta)\le t_Q/(1+t_Q)$ with $t_Q=\pi\|Q\delta\|^2$. Finally $\tilde M_F=\tilde\Sigma_F+\delta\delta^\top\succeq\delta\delta^\top$, so $Q\delta\delta^\top Q\preceq Q\tilde M_FQ$, whose largest eigenvalue is $\gamma_{k+1}$ because $Q$ is the spectral projector onto the remaining eigenvectors; thus $\|Q\delta\|^2\le\gamma_{k+1}$, and $t\mapsto t/(1+t)$ is increasing. With $Q=I-\delta\delta^\top/\|\delta\|^2$, $Q\delta=0$ and $R^{2\prime}=0$.

\emph{Unregularised whitening is needed.} If $\Sigma_R+\alpha I$ replaces $\Sigma_R$ in the whitening, the retain covariance of $z$ is no longer $I$ and the first step fails. Example in $\R^2$: $\Sigma_R=\Sigma_F=\mathrm{diag}(10^{-3},1)$, $\mu_R=0$, $\mu_F=(0.1,0)$, $\pi=0.2$. Then $R^2=0.615$, below the bound $0.667$, but with $\alpha=10^{-2}$ times the mean diagonal the formula gives $t/(1+t)=0.250$. With a singular $\Sigma_R$, the bound holds for pseudo-inverse whitening onto $\mathrm{range}(\Sigma_R)$ when the probe also sees only that range, which is how X4 checks it. Every step uses only first and second moments, so it applies verbatim to the empirical distribution of a finite sample, which gives the in-sample statement.

\emph{(c)} With population whitening, $z_i\sim\mathcal N(0,I_p)$ i.i.d., and $\tilde M_F=\frac1m\sum_iz_iz_i^\top$. For orthogonal $O$, $(Oz_i)_i$ has the same law as $(z_i)_i$, so $O\tilde M_FO^\top\overset d=\tilde M_F$; the top eigenvector is almost surely unique up to sign, so $Ov_1\overset d=\pm v_1$ and $v_1$ is uniform on the sphere, giving $\E(u^\top v_1)^2=1/p$. For $p/m\to y\in(0,1]$ the empirical spectral distribution converges to the Marchenko--Pastur law with support $[(1-\sqrt y)^2,(1+\sqrt y)^2]$ \citep{marchenko1967distribution} and the extreme eigenvalues converge almost surely to its edges (\citealp{bai1993limit}, smallest; \citealp{yin1988limit}, largest), so $\gamma_1-\gamma_p\to4\sqrt y$. $\delta=\frac1m\sum_iz_i\sim\mathcal N(0,I/m)$, so $\E\|\delta\|^2=p/m$, and by (b) the pre-edit $R^2$ is at most $\pi\|\delta\|^2/(1+\pi\|\delta\|^2)\le\pi\|\delta\|^2$. The removed directions are functions of $(z_i)$ alone, independent of the population; the retain energy removed is $k$ whitened units out of $p$ by (a).

\emph{(d)} This part needs neither $m\ge p$ nor a Gaussian law. $\tilde M_F=\frac1m\sum_iz_iz_i^\top$ has range $\mathrm{span}\{z_i\}$, of dimension at most $m\le k$, so its eigenvectors with nonzero eigenvalue are among the top $k$ and $Q$ annihilates $\mathrm{span}\{z_i\}$. Hence $Qz_i=0$ and every forget point is mapped to $\mu_R$. Under population whitening a fresh retain point has $\E zz^\top=I$, so $\E\|Qz\|^2=\tr Q=p-k$ (for $Q$ independent of $z$). \qed

\section{Novelty audit}\label{app:novelty}
Before writing any claim we ran five targeted searches (arXiv, OpenReview, Semantic Scholar and publisher pages; about a dozen queries each; the search log is included in the code release). Rate limits meant that ``nothing found'' is weak evidence. \Cref{tab:novelty} lists the closest prior art for each result.

\begin{table}[h]
\centering\small
\begin{tabular}{@{}p{3.6cm}p{2.3cm}p{9.8cm}@{}}
\toprule
Result & Source & Closest prior art and what is ours\\\midrule
Thm.~\ref{thm:pipeline}(a) ridge downdate & known (cited) & Summation form \citep{cao2015towards}; zero-residual Newton removal \citep{guo2020certified}; exact ridge heads on frozen features \citep{huang2025towards,quan2026exact,tavory2026exact}; random features \citep{yang2025muso}; broad learning systems \citep{zhu2019efficient}.\\
Thm.~\ref{thm:pipeline}(b) Select & squared loss known; 0-1 certificate ours & Squared-loss selection from stored moments is classical closed-form ridge validation \citep{allen1974relationship,golub1979generalized}. Cao and Yang cover feature selection (a $\chi^2$ test in summation form), not model selection; \citet{suriyakumar2022algorithms} show approximate methods fail under cross-validated selection, without an exact rule. Margin certificate = Ginart et al.'s quantised check-then-retrain \citep{ginart2019making} applied to an argmax over a grid. \citet{okumura2015quick} bound individual predicted labels under incremental data changes without retraining (sensitivity analysis, applied to leave-one-out CV); ours is the use inside grid selection and self-training.\\
Thm.~\ref{thm:pipeline}(c) Augment & ours & No exact deletion for self-training or pseudo-labels found; label-agnostic forgetting \citep{shen2024label} is a different problem.\\
Prop.~\ref{prop:adaptive} & known (cited), elementary & \citet[Sec.~1.2]{gupta2021adaptive} note that deterministic methods, and methods whose randomness is drawn after the request, are robust to adaptive deletion. Ours: the explicit pathwise statement for this pipeline. \citet{chourasia2023forget} define deletion under adaptive requests and use noisy training.\\
Thm.~\ref{thm:centres} & ours & No exact unlearning with data-sampled Nystr\"om landmarks or patch filters found. The adaptive failure follows the mechanism of \citet{gupta2021adaptive}.\\
Thm.~\ref{thm:cost}(a) & known & Standard operation counts; the $\min(N/|F|,3N/d)$ form replaces the naive $N/|F|$.\\
Thm.~\ref{thm:cost}(c) $\Omega(d^2b)$ state & ours & For a single ridge fit only. \citet{cherapanamjeri2025space} bound space for realisability testing (eluder dimension, $\Omega(n)$ bits), not exact ridge; \citet{ghazi2023ticketed} study the central state of ticketed exact unlearning; \citet{sekhari2021remember} give the $O(d^2)$ upper bound.\\
Prop.~\ref{prop:padding} & partly known (cited) & \citet[Thm.~1, Prop.~1]{yong2026erased} build output-identical models with separable representations by shifting features along the head's null space. \citet[Sec.~4]{kodge2024deep} and \citet[Sec.~2, A.8]{hatami2026class} observe, without a theorem, that a large negative forget-class bias passes accuracy metrics while the information persists. Ours: any output functional on all inputs, the zero-weight padding form, and the delete-unit corollary as a statement about every probe. \citet{thudi2022on} make the weight-level analogue.\\
Def.~\ref{def:probegap} ProbeGap & partly known (cited) & Retraining as the reference: \citet{yong2026erased}; \citet{hatami2026class} compare representational dissimilarity matrices with the retrained model's; over-unlearning: \citet{ha2026unlearnings}.\\
Thm.~\ref{thm:probe}(a) & known (cited) & Ky Fan for pencils \citep{fan1949on,kokiopoulou2011trace}.\\
Thm.~\ref{thm:probe}(b) & partly known; bound after the edit ours & The $R^2$--Mahalanobis identity is the classical link between indicator regression and LDA \citep[Ch.~4]{hastie2009elements}. LEACE proves guardedness iff equal means and gives the least-change eraser \citep{belrose2023leace}. The bound $R^{2\prime}\le t_Q/(1+t_Q)$ (and its spectral form) for the generalised-eigenvector edit was not found. The full texts of Kodge et al.\ and DAMP \citep{kodge2024deep,hatami2026class} state no theorem: Kodge et al.\ project out SVD bases of unwhitened activations and evaluate by accuracy, membership attacks and saliency; DAMP projects out unwhitened prototype residuals with a probe-based strength and evaluates probes empirically.\\
Thm.~\ref{thm:probe}(c)(d) & ours & Uses Marchenko--Pastur and Bai--Yin; the consequence for projection unlearning of random subsets was not found.\\
proj$(k)$ estimator & ours (detail) & Spectral signatures and SPECTRE use related statistics for detection only \citep{tran2018spectral,hayase2021spectre}.\\
\bottomrule
\end{tabular}
\caption{Prior art for each result. ``Ours'' means we did not find it stated before. Known results are used with citation; their proofs in \cref{app:proofs} are for completeness.}\label{tab:novelty}
\end{table}

\section{Experiment details}\label{app:exp}

This appendix gives the settings of every experiment, in the order of the main text. Each part starts with the question the experiment answers. All numbers in the main text come from the tables at the end of this appendix.

\subsection{Shared settings}

\paragraph{Compute.} All runs used one workstation. X1--X6, X10 and X12 ran on the CPU only, with at most 14 of 24 threads and float64 for every closed-form computation. X7 ran on a shared RTX 5090 with a 4 GB memory cap and at most 4 CPU threads: 2.4 GPU-hours, 1.4 h wall-clock with two processes at a time, at most 1.5 GB per process. Wall-clock times: X6 1 min, X1 9 min (plus 10 min for the competitors), X2 11 min, X3 8 min, X4 10 min, X5 55 min (plus 11 min for the WaNet follow-up). The follow-up rounds: X8 2.7 GPU-process-hours (three processes at a time, under 2 GB each) and about 9 core-hours on the CPU (at most 12 threads); X9 18.5 GPU-process-hours (four processes at a time, 1.5--1.8 GB each, 4.7 h wall-clock); X10 6.3 core-hours on the CPU (2 threads per process, 1.7 h wall-clock); X12 under 2 core-hours (at most 12 threads in total). The code release (to be linked from a later version of this preprint) contains the code, the raw results of every run, the logs and the prediction notebook. It keeps the original runs, which used a scale or $\lambda$ frozen from training data, apart from the X12 reruns with data-free constants.

\paragraph{Data and splits.} For every dataset the training split is permuted with seed 12345. The first $5{,}000$ images are the validation split and the rest the fit split. Membership pools, shadow models, probes and FM-recovery maps use fit or validation images only. The test split is read once per frozen final model, for accuracy and prediction identity only.

\paragraph{Features.} DINOv2-B (1,536 = class token and mean patch token), DINOv2-L (2,048) and CLIP-B/32 (512) features are computed once from the public checkpoints; a constant 1 is appended. The written CIFAR-10 features are 544 hand-written $6\times6$ Gabor and colour filters, ReLU, $4\times4$ average pooling (8,704 numbers), an element-wise square root, and a fixed-seed Gaussian projection to 1,024. No image statistic enters the unscaled map. X2 and X10a multiply these features by one scale constant (see X12 below); X1 uses them unscaled.

\paragraph{Labels of the methods.} \Cref{tab:labels} gives the label (exact, certified or heuristic) of every method we run.

\subsection{X1: one ridge read-out, many deletions}
\emph{Question: do rounding errors build up over many exact deletions, and how fast is each one?}
The two ridge levels are relative: $\lambda=\rho\cdot\mathrm{mean\,diag}\,G(D)$ with $\rho\in\{10^{-6},10^{-3}\}$, where $G(D)$ is the Gram matrix of the fit data. This $\lambda$ is computed once and held fixed, so these runs are exact relative to it; X12 reruns the written arm with a data-free $\lambda$. We delete $1{,}000$ uniformly random fit points one at a time. The refactorisation path downdates $(G,B)$ and solves by Cholesky after every deletion. The chained Sherman--Morrison path updates a stored inverse and the weights with the rank-one formula (Woodbury with $|F|=1$). We retrain from scratch every 100 deletions. Timings are medians over the deletions of one run; X8b repeats them five times (\cref{tab:timing}).

\emph{Certified removal.} One-vs-rest binary logistic regression with $L(w)=\sum_i\ell(w^\top x_i,y_i)+\lambda n\|w\|^2/2+b^\top w$, features scaled to $\|x\|\le1$, $\lambda$ from $\{10^{-6},\dots,10^{-3}\}$ on validation, Newton training, Newton removal, the data-dependent residual bound $\gamma\|X\|\|H^{-1}\Delta\|\|XH^{-1}\Delta\|$ with $\gamma=1/4$, and $\sigma=c\beta/\varepsilon$ with $c=\sqrt{2\ln(1.5/\delta)}$ for the cumulative bound $\beta$, as in \citet{guo2020certified} (\cref{rem:privacy} uses the Gaussian-mechanism constant $1.25$ of \citealp{dwork2014algorithmic}). The bound was accumulated over 100 (50 in the follow-up) simulated sequential removals and extrapolated linearly to 1,000. The accuracy in those rows is that of the noisy model before any removal, so they are a prospective calibration, and $(\varepsilon,\delta)$ holds per one-vs-rest head (the 10-class model is $(10, 10^{-3})$ by basic composition). X8b then runs their Algorithm 2 online with this noise (below): the largest bound after 1,000 removals is $0.81$ of the budget, so the linear extrapolation overestimated it by about $1.24\times$. With $\lambda$ tuned for noiseless accuracy, the noise destroys the model ($\approx10\%$ test accuracy).

\emph{SISA.} Fixed shards and deterministic ridge fits, so it is exact and, by \cref{prop:adaptive}, safe under adaptive requests; with random shard assignment it would be exact only in distribution.

\subsection{X2: the whole pipeline}
\emph{Question: does exactness survive selection and self-training, and how often do the certificates pass?}
$|L|=1{,}000$ random fit images (seed 2027), $|V|=1{,}000$ random validation images, and $U$ the remaining $44{,}000$ fit images; $\lambda\in\{0.3,1,3,10,30,100,300,1000\}$ (absolute; DINOv2 features unscaled; written features multiplied by one constant $s$, see X12); pool weight $w=0.3$; $k=3$. The order of requests is random; batches mix $L$ (30\%) and $U$ (70\%) points. Margins of points that are not recomputed are replaced by their certified lower bounds and never refreshed. The coarse selection check of \cref{tab:x2} compares these decayed bounds with the exact weight change, while the certificate of \cref{thm:pipeline}(b) uses $\delta_h$, which is zero for pool and validation deletions. All runs used the factor $2$ in place of the $\sqrt2$ of \cref{thm:pipeline}; it is also valid, only looser. The worst-case check of \cref{tab:x2} took the minimum over $U\setminus F$ from the stored per-point margins; the compact version with a stored, decaying lower bound $\tilde m_r$ can pass at most as often. Deviation from the plan: the plan had 200 single deletions (100 from $L$, 100 from $U$); the run added 20 from $V$. The DINOv2 arm selects the largest candidate ($\lambda^\star=1000$).

\emph{Memory} in float64 for DINOv2-B ($d=1{,}537$): the compact state $T(D)$ is about 59 MB, mostly the three $d\times d$ Gram matrices (19 MB each), and under 100 MB with 200 classes. The full state $T^+(D)$ adds the pool features: 541 MB for $|U|=44{,}000$ (CIFAR-10, CIFAR-100) and 1.14 GB for $|U|=93{,}000$ (Tiny-ImageNet), before validation features, margins and temporaries.

\emph{Sampled centres} (\cref{thm:centres}). With $n=60$, $m=6$, $|F|=6$ and $10^4$ trials, resampling looks like retraining for a fixed request ($\chi^2$ $p=0.78$), while a requester who deletes non-centres finds all $6.00$ published centres still in place, against $0.67$ under retraining ($p<10^{-300}$).

\subsection{X3: membership attacks}
\emph{Question: does exact deletion hide the deleted points from an attacker?}
Ridge $\lambda=10^{-4}\cdot n\cdot\mathrm{mean}\|x\|^2$ with $n=2{,}500$ and the mean over the whole $5{,}000$-image pool, computed once and held fixed for the original, the exact deletion and the retrained head. The softmax head is trained by full-batch Adam (200 steps, lr $10^{-2}$, zero initialisation). FT, NegGrad+ ($\alpha=0.9$) and SCRUB use grids of three configurations each, tuned on 8 shadow models by distance to the retrained head on validation images. proj$(8)$ uses a retain-covariance regulariser of $10^{-2}$ times the mean diagonal. The U-LiRA-style attack fits per-example Gaussian means (shared variance) on 128 shadow models and scores 128 targets ($31{,}948$ forget positives, $319{,}557$ negatives not fitted in that head). Attack statistic: the logit margin $f_y-\max_{j\ne y}f_j$. Two predictions failed here: the softmax baselines sat $0.014$--$0.022$ above retraining, not the predicted $0.03$, and proj$(8)$ \emph{lowered} the AUC from $0.813$ to $0.668$, against our prediction (with $m=250<p$ the top directions capture sample-specific variation), though it stays far above retraining ($0.501$). The first before/after attack, by margin change, reached AUC $0.79$ against a predicted $0.95$; the squared-loss score of X8c exceeds it.

\subsection{X4: class forgetting and probes}
\emph{Question: do output scores and probes agree on which methods forget?}
The small CNN (two conv layers, a 64-unit layer fc1, read-out fc2) is trained for 3 epochs with Adam (lr $10^{-3}$, batch 128), with three original seeds and a band of five retrained seeds. Our baselines follow \citet{fan2024salun,kurmanji2023towards,foster2024fast,jia2023model}. Gradient baselines use grids of 5--7 configurations, plus $\ell_1$-sparse ($\gamma\in\{10^{-5},10^{-4},10^{-3}\}$, 1 or 3 epochs) and a simplified SalUn (gradient-saliency mask of the top 50\% per tensor, then random-label fine-tuning; 1 or 3 epochs, three learning rates). All methods are tuned on validation images by distance to the band. LEACE is applied as a weight edit at the fc1 input with the eraser of \citet[Thm.~4.3]{belrose2023leace} for the binary forget indicator; DAMP-lite projects out the unwhitened prototype residual $\mu_F-\mu_R$ at the same place. All three edits, and proj$(k)$, are folded into fc1's weights and bias, not inserted as a layer. So fc1's run-time input $x$ is unchanged (its probe still reads the original features), the probe ``at the edited layer'' reads fc1's pre-activation, a linear map of the projected $x'$, and the theorem check reads $x'$ itself. Probe samples: $5{,}000$ retain and $1{,}000$ forget fit images (in-sample), $2{,}000$ retain and all forget validation images (held-out). The theorem check uses pseudo-inverse retain whitening on the range of $\Sigma_R$ (rank 782--784 of 784), with the probe fitted on the whitened features in that range, as \cref{thm:probe}(b) requires. The deployed edits (proj, here and in X3, X5, X7) use the regularised whitening $\Sigma_R+10^{-2}\,\mathrm{mean}(\mathrm{diag}\,\Sigma_R)\,I$, to which the bound does not apply. The padded model of \cref{prop:padding} reproduces the retrained model's logits exactly, while its penultimate probe is at least the original's ($R^2=0.857$ vs $0.837$, MNIST). The spectral form of the bound also holds but is loose ($0.57$--$0.95$ for $k\le32$).

\subsection{X5: backdoors on MNIST}
\emph{Question: does the edit remove a backdoor when only some poisons are known, on small networks?}
1\% of the fit images (550), drawn from non-target classes, are triggered and relabelled 0. BadNets \citep{gu2017badnets}: a $4\times4$ white square at the bottom right. Blended \citep{chen2017targeted}: $0.8x+0.2\eta$ with a fixed uniform noise image $\eta$. WaNet \citep{nguyen2021wanet}: a fixed $4\times4$ random warp field, bicubically upsampled, strength $0.5$, then $2.0$ in the declared follow-up because $0.5$ did not implant; we train without the original's noise mode. SIG \citep{barni2019new}: $x+0.1\sin(2\pi\cdot6j/28)$ along columns, clipped; the original is clean-label, and we use it dirty-label like the others. Our 1\% poison rate is lower than in many benchmark settings. We unlearn with 100, 50, 25 or 10\% of the poisons known. GD configurations were chosen once, on BadNets with all poisons known (seed 0), by clean validation accuracy minus the target-label rate on the known poisons, and reused for every trigger and fraction; this favours the edit elsewhere. $k$ for proj$(k)$ is chosen per cell by the same score. Written head: $\phi(x)=[\mathrm{ReLU}(R^\top x+b);1]$ with fixed random $R\in\R^{784\times2048}$ and $b$, and ridge $\lambda=0.1$ times the mean diagonal of $G$, recomputed on the remaining data by both the exact deletion and the retraining ($\tr(G')/d$ is exact from the stored sums).

\emph{Results} (\cref{tab:x5}). With 10\% known, proj$(k)$ reaches $1.2$--$18.0\%$ ASR on MNIST and $7.7$--$29.9\%$ on Fashion, against $36.5$--$99.7\%$ for retraining on the remainder, FT and SCRUB, at unchanged clean accuracy. NegGrad+ also lowers ASR, but its clean accuracy falls to $69$--$82\%$. On Fashion WaNet ($s=2$, under-implanted, not scored) retraining on the remainder wins at 100, 50 and 25\% known ($2.6$--$16.9$ vs $19.0$--$20.2\%$). The exact written head equals its own retraining on the remainder in all 96 cells ($\le4\cdot10^{-13}$), so it keeps the backdoor exactly as retraining does: an exact method is only as good as its deletion request.

\subsection{X7: backdoors on CIFAR-10 ResNet-18}
\emph{Question: do the MNIST results hold on a standard network, and is the trigger really gone?}
Training: CIFAR ResNet-18 ($3\times3$ stem, no max-pool), 30 epochs, SGD with Nesterov momentum 0.9, peak lr 0.1 (one warm-up epoch, then cosine), weight decay $5\cdot10^{-4}$ (not on BatchNorm or biases), batch 128, random crop and flip, bf16 autocast, channels-last. BadNets: a $3\times3$ white patch at rows and columns 27--29. Blended: $0.8x+0.2\eta$ with a fixed uniform noise image $\eta$. WaNet: a fixed $4\times4$ random warp field, bicubically upsampled to $32\times32$, strength $s=1.0$, no noise mode. All three are dirty-label at 1\%. The trigger is applied before augmentation. proj-pen sits at the penultimate layer and is folded into the read-out, so the pre-edit features still exist at run time; proj-mid is an inserted fixed $1\times1$ layer on the 256 channels after the third stage. FT grid: (1 epoch, lr 0.01), (3, 0.01), (3, 0.03), cosine, no warm-up. SalUn (simplified): the mask is the set of parameters whose forget-loss gradient magnitude is in the top 50\% (global threshold); then SGD (momentum 0.9) on batches of 128 remainder images with true labels and 64 identified poisons with random labels $\ne0$; grid lr $\in\{0.003,0.01,0.03\}$, 100 or 300 steps. The FT and SalUn configurations were chosen on seed 0 for each trigger and known fraction by the defender's score and reused for seeds 1--4. proj-mid uses 2,000 retain images (128,000 position vectors); proj-pen uses 5,000. Retain-covariance regulariser: $10^{-2}$ times the mean diagonal, as in X4 and X5. Head-relearn reads the penultimate features after the edit; for proj-pen this is the edited layer. The control trigger of a new trigger of the same family reaches only $1$--$6\%$ after one step on the edited models, so the fast return is specific to the implanted trigger.

\subsection{X8: follow-ups to X1--X3 and X7}
\emph{Questions: where does the speed come from; does certified removal do better online; how strong is the before/after attack; where is the trigger stored; do dedicated defences beat the edit?}

\emph{(a) Caching baseline.} X2's 240 requests run through four paths on identical state: X2's certified path as released (with its a-priori diagnostic eigendecomposition of $G_L$), the same path without it (lean), a cached exact path (the same stored $(G,B)$ and re-solves, but every validation and pool row rescored in every round whose input weight changed, and no margins), and a full rerun; 6 threads. Anchored margins: one snapshot $W_{\mathrm{ref}}$ per round and $\lambda$; a point counts as certified if its snapshot margin exceeds $c\|W_t-W_{\mathrm{ref}}\|$ and its stored label equals the snapshot label, with $c=2$ and $\sqrt2$. A squared-loss arm selects $h^\star$ from $(G_V,B_V)$; it matched a rerun on all 241 states.

\emph{(b) Certified removal online.} Guo et al.'s Algorithm 2 on X1c's features, noise ($\sigma=6.87$, 3 noise seeds) and $\lambda=10^{-3}$, with $\varepsilon=1$, $\delta=10^{-4}$, $\gamma=1/4$: Newton removals along X1's 1,000-removal order, and a retrain of a head with fresh noise whenever its accumulated data-dependent bound exceeds $\sigma\varepsilon/c$. It ends at $94.76\%$ against $96.18\%$ for the exact head, with no retrain, $6.4\times$ faster than retraining. Timing repeats: X1's six feature sets and paths, 5 repetitions of the first 200 deletions, and the scaling slopes, at 6 threads.

\emph{(c) Before/after sweep.} X3's pool, halves and $\lambda$ on all 256 heads; one exact batch of $B\in\{1,10,50,250\}$ points per head ($B=1$: 10 single requests); margin, squared-loss and label-only (correct before minus correct after) scores; tie-aware AUC; 1,000 cluster resamples over heads. At $0.1\%$ FPR the margin score has the higher TPR ($0.19$ vs $0.11$ for the 250-point batch). For $B=250$, the TPR at $0.1\%$ FPR is $11$--$36\%$ across settings.

\emph{(d) Probes and detect-then-retrain.} X7's data, poisons and checkpoints; the edits, FT and SalUn are re-created with X7's procedure (their validation ASR is within 4.2 points of X7's). Taps and relearn curves as in \cref{tab:probe5}; all models relearn on the same clean remainder (fit set minus all 450 poisons), at steps 0--100, with two relearn seeds. Declared deviation, fixed after a smoke test and before any main run: the pre-registered control triggers were not new to the poisoned model (at step 0 a $3\times3$ checkerboard fired the BadNets implant on $12.7\%$ vs $1.8\%$ of validation images for the clean model, a second noise image the Blended implant on $99.9\%$ vs $0.0\%$, a second warp field the WaNet implant on $18.5\%$ vs $0.8\%$). By a rule fixed before any relearn result we took the first candidate in a listed order whose step-0 rate on the original is within 3 points of the clean model's: a red $3\times3$ patch at the top left, a 0.2 blend with a smooth low-frequency colour image, and a warp field on an $8\times8$ grid. Detect-then-retrain takes the layer and $k$ of X7's validation-selected edit at 10\% known and scores each remaining fit image by $\|U_k^\top A(z-\mu)\|^2$ (proj-mid: maximum over positions) \citep{tran2018spectral}. R1 flags the top 675 (it assumes the poison rate is known); R2 thresholds at the 10th percentile of 5-fold cross-fitted known-poison scores; the model is retrained for 30 epochs with X7's recipe and seed. R1 beats the edit on BadNets ($1.0\%$) and Blended ($0.2\%$) and relearns like the clean model, but loses on WaNet ($17.1\%$, recall $68\%$). R2 (3 seeds) flags 2,060 images on WaNet and reaches $3.0\%$, but fails on Blended ($88.6\%$ at $89\%$ recall; one VGG seed fails alike).

\emph{(e) ABL} \citep{li2021anti}, following the authors' released code: 10 epochs from scratch with the LGA loss ($\gamma=0.5$), isolation of the 1\% lowest-loss images (450), 60 epochs of fine-tuning on the rest (lr 0.1, then 0.01 from epoch 40), and gradient ascent on the isolated images (lr $5\cdot10^{-4}$, 4 passes). Our changes: our ResNet-18, no Cutout, bf16, our normalisation. ABL ends at $21$--$95\%$ ASR, 10--12 points lower in clean accuracy; at its design rate of 10\% poisons it reaches $3.3\%$ (one seed; this check was pre-registered after the main ABL runs).

\subsection{X9: more seeds, more settings, and the probe theorem at scale}
\emph{Questions: are the X7 results stable over seeds and settings, and does \cref{thm:probe} hold on ResNet-18?}
\emph{(a)} Seeds 3 and 4 of X7 with X7's seed-0 FT and SalUn configurations, X8's probes on the same models, and detect-then-retrain R1 with the layer and $k$ of that run's 10\% edits. \emph{(b)} Three seeds per setting, BadNets and Blended: 0.5\% (225) and 5\% (2,250) poisons on CIFAR-10; CIFAR-100 (target class 0, a 100-way read-out); SVHN (683 poisons, 20 epochs, no flip); VGG-16-BN on CIFAR-10 (conv blocks 64-64 / 128-128 / 256$\times$3 / 512$\times$3 / 512$\times$3 with BatchNorm, ReLU and max-pooling, then a linear read-out; proj-pen at the 512-d read-out input, proj-mid at the output of block 3 before its max-pool, fixed before running). CIFAR-100 and SVHN are normalised by fit-split moments; SalUn's random labels range over all classes; R1 flags 1.5 times the number of poisons; the relearn probes keep 45 known poisons at every rate; the control trigger is used on CIFAR-10 at 1\% only. FT and SalUn configurations are chosen on seed 0 per setting and trigger; the layer rule of X7 is pre-registered here. On VGG-16-BN, BadNets needs proj-mid ($1.3\%$; proj-pen $63\%$ even with all poisons known). The one-step return appears in 9 of 10 settings; on CIFAR-100 BadNets the probe before the edit still reads the trigger ($88$ vs $0.3\%$). \emph{(c)} Clean CIFAR-10 ResNet-18 (X7's recipe without poisons): originals with seeds 0--2 on all 45,000 fit images, and a retrain band of seeds 10--14 per forget set (20 models). Forget sets: class 3, and nested random sets of 100, 1,000 and 5,000 fit images; control sets of retained images of the same sizes. proj$(k)$ at the penultimate layer with pseudo-inverse whitening on the range of $\Sigma_R$ (eigenvalues below $10^{-9}$ of the largest dropped), inserted as an explicit fixed $512\times512$ linear layer before the read-out (in fp32), so post-edit probes, head-relearn and $R^2$ read the projected $x'$; $R^2$ in sample on all fit features and held out on validation images. $k^\ast$, FT and SalUn are chosen (on seed 0 for the last two) by distance to the band mean of validation retain and forget accuracy. Per-example AUC: for each forget image the margin $z_y-\mathrm{logsumexp}_{j\ne y}z_j$, minus the mean over the other band members; AUC between the method's and a band member's values over the forget set, averaged over band members. For class 3, proj$(k^\ast)$ over-unlearns (head-relearn $0\%$ against a band of $65$--$67\%$), and FT and SalUn sit above the band (\cref{tab:class5}). For random sets, $\|\delta\|^2$ is $0.90$--$1.08$ times the population-whitened null, $\gamma_1$ is within $7\%$ of the Marchenko--Pastur edge, and the edit moves the forget set (per-example AUC against retraining $0.61\to0.37$ at $m=1{,}000$, $0.69\to0.49$ at $5{,}000$) and costs 1--1.5 test points; the control with directions from a retained set ($0.49$, $0.56$) was added after the results and is not pre-registered. Gaps to the theorem's assumptions: post-ReLU features are not Gaussian, the whitening uses empirical moments, and $m=100<p$ lies outside (c). SCRUB, named in the plan, was not run.

\subsection{X10: X2 and X3 on more data}
\emph{Question: do the X2 and X3 results hold on other datasets, orders and backbones?}
\emph{(a)} X2's pipeline, candidate list, rounds, request mix and certificate code unchanged; every request runs through the certified, lean, cached and rerun paths with a comparison after each. CIFAR-100 DINOv2-B ($|L|=|V|=1{,}000$, $|U|=44{,}000$) and Tiny-ImageNet DINOv2-B ($|L|=|V|=2{,}000$, $|U|=93{,}000$; its labelled validation split serves as the test split), with X2's request seed; CIFAR-10 in both arms with X2's $L$, $U$, $V$ and three new request seeds. Declared deviation: the script first set $|L|$ to 10 per class for every dataset, so the first CIFAR-10 run drew only 100 labels; it was stopped after 21 requests (all exact) and is not scored, and the CIFAR-10 runs use X2's 1,000 labels. The pre-registration had also misstated X2's labels per class (it has 100), so CIFAR-100 and Tiny-ImageNet have ten times fewer labels per class than X2. \emph{(b)} X3 and X8c unchanged except 512 heads, $B\in\{1,250\}$ ($B=1$: 5 single requests per head), and three settings: CIFAR-10 CLIP-B/32 ($d=513$), CIFAR-10 DINOv2-B ($d=1{,}537$) and CIFAR-100 DINOv2-L ($d=2{,}049$); U-LiRA as in X3 on 256 shadow and 256 target heads, which stays at $0.500$--$0.502$ on the deleted head alone. Serving labels only gives AUC $0.50$--$0.51$ on CIFAR-10 and $0.54$ on CIFAR-100 ($8.1\%$ TPR at $0.1\%$ FPR for single deletions); we did not test stronger label-only attacks. No test data are read.

\subsection{X12: a feature scale that uses no data}
\emph{Question: does exactness still hold when even the feature scale is computed without training images?}
The first written runs of X2, X8a and X10a multiplied the written features by $s=1.4/\mathrm{RMS}$ of the features of the first $2{,}000$ CIFAR-10 training images ($s=8.36$), computed once and held fixed. These images can be deleted, so those runs match a rerun with the same frozen $s$, not a rerun that would recompute $s$ on $D\setminus F$; their exactness is conditional on that scalar. X12 replaces $s$ by the same formula on $2{,}000$ synthetic images with i.i.d.\ uniform pixels (NumPy generator seed 20261126): $s_0=4.878$, with no CIFAR image involved. The rule, and the predictions in \cref{tab:prereg2}, were written before any X12 run. Everything else is unchanged. X2 with $s_0$: 240/240 requests exact (same $h^\star$, all pseudo-labels, weights within $5.5\cdot10^{-12}$), $\lambda^\star=100$, test accuracy $49.17\%$ before and $45.50\%$ after self-training (frozen-scale run: $\lambda^\star=300$, $49.11$ / $45.39\%$). X10a's three CIFAR-10 written orders with $s_0$: 720/720 exact on the certified, lean and cached paths, within $5.8\cdot10^{-12}$, final test predictions identical (\cref{tab:x2gen}). A pre-registered sensitivity run with $s=1$ was also exact on 240/240 (within $1.0\cdot10^{-11}$; $\lambda^\star=3$, $49.67$ / $46.22\%$), so the accuracy does not hinge on the choice of scale. X1 uses no scale but sets $\lambda=\rho\cdot\mathrm{mean\,diag}\,G(D)$ once from the fit data, the same kind of frozen constant; X12 reruns X1's written arm with $\lambda=\rho\cdot45{,}000\cdot m_0$, where $m_0$ is the mean squared coordinate of the bias-augmented noise features. It is again within $3.1\cdot10^{-11}$ of retraining after 1,000 deletions with 100\% identical test predictions, at $65.9$ / $66.0\%$ test accuracy for $\rho=10^{-6}$ / $10^{-3}$. X8a's written arm (\cref{tab:x2cache}) was not rerun and keeps the frozen scale.

\subsection{Tables and figures}

\begin{table*}[t]
\centering\footnotesize
\setlength{\tabcolsep}{3pt}
\begin{tabular}{@{}llccccc@{}}
\toprule
trigger & method & ASR, 100\% known & ASR, 25\% known & ASR, 10\% known & clean (10\%) & head-relearn (10\%)\\\midrule
BadNets & original & \multicolumn{3}{c}{94.6 [94.1, 95.0]} & 92.8 & 93.9\\
 & proj-pen$(k)$ & 5.1 [4.2, 6.1] & 21.5 [16.0, 27.0] & 39.0 [32.1, 45.9] & 92.7 & 45.7\\
 & proj-mid$(k)$ & 1.3 [1.1, 1.4] & 1.5 [1.4, 1.7] & 1.5 [1.3, 1.7] & 92.7 & 3.3\\
 & proj, layer by val. (mid 5/5) & 1.3 [1.1, 1.4] & 1.5 [1.4, 1.7] & 1.5 [1.3, 1.7] & 92.7 & 3.3\\
 & retrain rem. & 1.1 [1.0, 1.2] & 94.3 [93.0, 95.5] & 94.8 [93.8, 95.9] & 92.9 & 94.0\\
 & FT & 68.9 [55.7, 82.0] & 94.5 [93.5, 95.5] & 94.5 [92.8, 96.2] & 92.7 & 93.9\\
 & SalUn (simpl.) & 0.0 [0.0, 0.0] & 0.0 [0.0, 0.1] & 0.1 [0.0, 0.1] & 91.9 & 95.0\\
 & detect-then-retrain & & & 1.0 [1.0, 1.1] & 92.8 & --\\
 & clean model (target) & & & & & 1.5\\\midrule
Blended & original & \multicolumn{3}{c}{99.9 [99.8, 100.0]} & 92.7 & 99.9\\
 & proj-pen$(k)$ & 0.1 [0.0, 0.3] & 2.2 [0.6, 3.7] & 7.8 [1.0, 14.6] & 92.7 & 42.1\\
 & proj-mid$(k)$ & 10.7 [0.5, 20.9] & 13.5 [0.0, 27.2] & 14.7 [4.9, 24.5] & 92.7 & 94.1\\
 & proj, layer by val. (pen 5/5) & 0.1 [0.0, 0.3] & 2.2 [0.6, 3.7] & 7.8 [1.0, 14.6] & 92.7 & 42.1\\
 & retrain rem. & 0.1 [0.0, 0.2] & 99.9 [99.8, 100.0] & 99.9 [99.8, 100.0] & 92.9 & 99.9\\
 & FT & 96.3 [91.7, 100.0] & 99.9 [99.8, 100.0] & 99.9 [99.9, 100.0] & 92.6 & 99.9\\
 & SalUn (simpl.) & 0.0 [0.0, 0.0] & 0.0 [0.0, 0.0] & 0.0 [0.0, 0.0] & 91.6 & 99.9\\
 & detect-then-retrain & & & 0.2 [0.0, 0.4] & 92.7 & --\\
 & clean model (target) & & & & & 57.5\\\midrule
WaNet & original & \multicolumn{3}{c}{77.1 [75.7, 78.4]} & 92.6 & 73.2\\
 & proj-pen$(k)$ & 4.0 [2.4, 5.6] & 8.3 [3.1, 13.5] & 16.7 [9.5, 23.9] & 92.6 & 31.0\\
 & proj-mid$(k)$ & 4.4 [2.9, 6.0] & 4.5 [3.2, 5.8] & 5.5 [3.8, 7.3] & 92.6 & 15.7\\
 & proj, layer by val. (mid 5/5) & 4.2 [3.3, 5.1] & 4.5 [3.3, 5.8] & 5.5 [3.8, 7.3] & 92.6 & 15.7\\
 & retrain rem. & 1.3 [1.2, 1.3] & 70.2 [68.6, 71.9] & 76.1 [73.7, 78.6] & 92.7 & 71.8\\
 & FT & 21.9 [19.9, 23.9] & 75.5 [73.7, 77.4] & 77.4 [74.9, 79.9] & 92.3 & 72.8\\
 & SalUn (simpl.) & 0.2 [0.1, 0.4] & 0.7 [0.4, 1.0] & 1.4 [0.5, 2.2] & 91.6 & 79.4\\
 & detect-then-retrain & & & 17.1 [11.4, 22.7] & 92.7 & --\\
 & clean model (target) & & & & & 2.4\\
\bottomrule
\end{tabular}
\caption{X7 at five seeds (X9a; CIFAR-10 ResNet-18, 1\% poisons): test ASR (\%), mean [t-based 95\% interval over seeds 0--4, clipped to $[0,100]$]. Seeds 0--2 are the original X7 runs, seeds 3--4 are new. ``retrain rem.''\ retrains from scratch on everything not known; at 100\% it is the clean model. ``proj, layer by val.'': per seed, the layer (pen or mid) with the higher defender score, a rule fixed after seeds 0--2; in brackets the layer it chose at 10\% known. Detect-then-retrain (X8d, R1): rank the unknown fit images by their spectral score along the known poisons' directions, drop the top 675, retrain. Head-relearn: validation ASR after a ridge refit of the read-out on the model's penultimate features of all 45k poisoned fit images; the last row of each block is the clean model's value.}\label{tab:bd5}
\end{table*}

\begin{table*}[t]
\centering\footnotesize
\setlength{\tabcolsep}{3.2pt}
\begin{tabular}{@{}llrrrrrrrrrrrr@{}}
\toprule
 & & \multicolumn{4}{c}{head-relearn at tap} & \multicolumn{5}{c}{FT-relearn, own trigger, step} & \multicolumn{3}{c}{control trigger, step}\\\cmidrule(lr){3-6}\cmidrule(lr){7-11}\cmidrule(l){12-14}
trigger & model & $\ell_2$ & $\ell_3$ & pen, pre-edit & pen & 0 & 1 & 5 & 10 & 100 & 1 & 5 & 10\\\midrule
BadNets & original & 60 & 81 & 94 & 94 & 94 & 94 & 94 & 95 & 94 & 2 & 60 & 99\\
 & clean retrain & 6 & 4 & 1 & 1 & 1 & 1 & 4 & 20 & 28 & 1 & 5 & 47\\
 & proj-pen, 100\% & 60 & 81 & 94 & 8 & 5 & 91 & 98 & 99 & 95 & 1 & 55 & 95\\
 & proj-mid, 100\% & 60 & 81 & 3 & 3 & 1 & 2 & 98 & 99 & 95 & 1 & 5 & 45\\
 & proj-pen, 10\% & 60 & 81 & 94 & 45 & 40 & 97 & 99 & 99 & 95 & 2 & 74 & 100\\
 & proj-mid, 10\% & 60 & 81 & 3 & 3 & 1 & 3 & 98 & 99 & 95 & 1 & 6 & 54\\
 & SalUn, 10\% & 68 & 86 & 95 & 95 & 0 & 1 & 87 & 91 & 91 & 1 & 33 & 82\\
 & FT, 10\% & 60 & 81 & 94 & 94 & 95 & 95 & 95 & 96 & 93 & 3 & 60 & 100\\\midrule
Blended & original & 99 & 100 & 100 & 100 & 100 & 100 & 100 & 100 & 99 & 1 & 2 & 16\\
 & clean retrain & 80 & 77 & 58 & 58 & 0 & 1 & 88 & 98 & 94 & 1 & 2 & 14\\
 & proj-pen, 100\% & 99 & 100 & 100 & 31 & 0 & 100 & 100 & 100 & 100 & 1 & 5 & 27\\
 & proj-mid, 100\% & 99 & 100 & 96 & 96 & 11 & 93 & 100 & 100 & 100 & 1 & 2 & 13\\
 & proj-pen, 10\% & 99 & 100 & 100 & 39 & 9 & 100 & 100 & 100 & 100 & 1 & 9 & 47\\
 & proj-mid, 10\% & 99 & 100 & 94 & 94 & 15 & 91 & 100 & 100 & 99 & 1 & 2 & 14\\
 & SalUn, 10\% & 100 & 100 & 100 & 100 & 0 & 0 & 50 & 88 & 100 & 1 & 2 & 8\\
 & FT, 10\% & 99 & 100 & 100 & 100 & 100 & 100 & 100 & 100 & 100 & 1 & 2 & 18\\\midrule
WaNet & original & 28 & 44 & 73 & 73 & 76 & 75 & 78 & 84 & 73 & 6 & 37 & 65\\
 & clean retrain & 6 & 4 & 2 & 2 & 1 & 2 & 8 & 49 & 28 & 2 & 8 & 41\\
 & proj-pen, 100\% & 28 & 44 & 73 & 13 & 4 & 55 & 91 & 93 & 70 & 5 & 42 & 59\\
 & proj-mid, 100\% & 28 & 44 & 14 & 14 & 5 & 8 & 59 & 88 & 66 & 3 & 10 & 43\\
 & proj-pen, 10\% & 28 & 44 & 73 & 31 & 15 & 76 & 92 & 92 & 71 & 6 & 48 & 50\\
 & proj-mid, 10\% & 28 & 44 & 16 & 16 & 6 & 10 & 66 & 91 & 66 & 3 & 11 & 47\\
 & SalUn, 10\% & 37 & 59 & 80 & 80 & 2 & 2 & 6 & 38 & 63 & 3 & 2 & 2\\
 & FT, 10\% & 27 & 43 & 72 & 72 & 74 & 75 & 81 & 87 & 75 & 6 & 24 & 57\\
\bottomrule
\end{tabular}
\caption{Where the trigger is (X8d, X9a; CIFAR-10 ResNet-18, mean of 5 seeds, validation ASR \%). Head-relearn: a ridge read-out refitted on frozen features of all 45k poisoned fit images, taken after layer2 ($\ell_2$, pooled), layer3 ($\ell_3$, pooled; before the proj-mid edit), at the penultimate layer before the proj-pen edit (pre-edit) and after it. FT-relearn: SGD (lr 0.01) on batches of 120 clean remainder images and 8 of 45 fixed known poisons; own trigger, mean of two relearn seeds. Control: a new trigger of the same family on the same 45 images, chosen by a rule fixed after a smoke test (declared deviation, \cref{app:exp}). For seeds 0--2 the probed models are re-created with X7's procedure (validation ASR within 4.2 points of X7's). proj-pen is folded into the read-out, so its pre-edit features exist at run time and the pen column reads the projected features; proj-mid is an inserted fixed $1\times1$ layer, so $\ell_3$ is read before it.}\label{tab:probe5}
\end{table*}

\begin{table*}[t]
\centering\footnotesize
\setlength{\tabcolsep}{3.5pt}
\begin{tabular}{@{}lllrrrrcr@{}}
\toprule
trigger & method & seeds & removed & precision & recall & clean test & ASR test & FT-relearn 100\\\midrule
BadNets & detect+retrain R1 & 5 & 675 & 58.4 & 97.4 & 92.8 & 1.0 [1.0, 1.1] & 23.4\\
 & detect+retrain R2 & 3 & 366 & 100.0 & 90.3 & 92.9 & 1.3 [0.6, 1.9] & 35.5\\
 & ABL (blind) & 3 & 450 & 0.2 & -- & 80.5 & 95.1 [91.9, 98.3] & 93.7\\
 & ABL, 10\% poisons & 1 & 450 & 100.0 & -- & 91.3 & 3.3 & 97.0\\
 & clean model (target) & 5 & & & & & & 28.5\\\midrule
Blended & detect+retrain R1 & 5 & 675 & 60.0 & 100.0 & 92.7 & 0.2 [0.0, 0.4] & 90.3\\
 & detect+retrain R2 & 3 & 362 & 99.5 & 89.0 & 92.9 & 88.6 [49.5, 100.0] & 98.7\\
 & ABL (blind) & 3 & 450 & 14.0 & -- & 82.7 & 21.5 [0.0, 100.0] & 99.9\\
 & clean model (target) & 5 & & & & & & 94.1\\\midrule
WaNet & detect+retrain R1 & 5 & 675 & 40.7 & 67.8 & 92.7 & 17.1 [11.4, 22.7] & 41.8\\
 & detect+retrain R2 & 3 & 2060 & 17.9 & 90.0 & 92.7 & 3.0 [2.7, 3.2] & 31.1\\
 & ABL (blind) & 3 & 450 & 0.0 & -- & 80.8 & 78.4 [63.4, 93.4] & 59.3\\
 & clean model (target) & 5 & & & & & & 28.4\\
\bottomrule
\end{tabular}
\caption{Dedicated baselines at 1\% poisons (CIFAR-10 ResNet-18), 10\% of the poisons known except for ABL, which uses none. Test \%, mean [t-based 95\% interval over seeds, clipped]. Detect-then-retrain scores every unknown fit image by its energy along the top directions of the known poisons (spectral signature, \citealp{tran2018spectral}; layer and $k$ from the validation-selected edit), removes the known and the flagged images, and retrains from scratch (about 3 min). R1 flags the top 675 (1.5 times the poison rate, rate treated as known); R2 thresholds at the 10th percentile of cross-fitted known-poison scores (rate unknown). Precision and recall are over the 405 unknown poisons. ABL \citep{li2021anti} follows the authors' released schedule; the 10\%-poison row (one seed, written after the main ABL runs) checks the implementation in ABL's design regime. FT-relearn 100: validation ASR after 100 relearn steps (as in \cref{tab:probe5}).}\label{tab:defence}
\end{table*}

\begin{table*}[t]
\centering\footnotesize
\setlength{\tabcolsep}{2.4pt}
\begin{tabular}{@{}llcc|cccccc|cc|c@{}}
\toprule
 & & & & \multicolumn{6}{c|}{ASR, 10\% known} & \multicolumn{2}{c|}{100\% known} & step 1\\
setting & trigger & orig. & clean & pen & mid & layer by val. & retr.\ / FT & SalUn & detect & pen & mid & pen / clean\\\midrule
C10, 0.5\% & BadNets & 93.1 & 92.8 & 36.1 & 1.3 & 1.3 [0.8, 1.8] 0p3m & 92.0 / 93.0 & 0.1 & 1.1 & 6.9 & 1.4 & 85.3 / 1.0\\
 & Blended & 99.8 & 92.8 & 10.6 & 14.9 & 8.9 [0.0, 28.8] 2p1m & 99.9 / 99.9 & 0.0 & 0.1 & 0.8 & 18.2 & 99.8 / 1.1\\
C10, 5\% & BadNets & 97.3 & 92.6 & 45.3 & 1.3 & 1.3 [1.1, 1.5] 0p3m & 97.4 / 97.4 & 2.1 & 1.2 & 5.0 & 1.3 & 95.4 / 1.4\\
 & Blended & 100.0 & 92.6 & 54.3 & 60.3 & 44.5 [14.2, 74.8] 2p1m & 100.0 / 100.0 & 1.3 & 0.1 & 10.8 & 51.7 & 100.0 / 0.9\\
C100, 1\% & BadNets & 87.7 & 73.2 & 3.1 & 0.1 & 2.1 [0.0, 7.6] 2p1m & 87.7 / 88.4 & 1.5 & 0.2 & 0.8 & 0.1 & 4.5 / 0.1\\
 & Blended & 99.7 & 73.1 & 1.9 & 0.7 & 1.8 [0.0, 5.9] 2p1m & 99.5 / 99.7 & 8.4 & 0.0 & 0.1 & 0.2 & 43.7 / 0.3\\
SVHN, 1\% & BadNets & 95.8 & 96.6 & 6.0 & 0.4 & 2.2 [0.0, 9.9] 1p2m & 95.9 / 95.6 & 0.0 & 0.2 & 0.2 & 0.3 & 97.6 / 0.3\\
 & Blended & 99.9 & 96.5 & 0.7 & 0.2 & 0.2 [0.0, 0.5] 1p2m & 100.0 / 100.0 & 0.0 & 0.2 & 0.1 & 0.0 & 100.0 / 0.1\\
VGG, C10, 1\% & BadNets & 93.3 & 92.0 & 68.1 & 1.3 & 1.3 [0.9, 1.7] 0p3m & 93.5 / 93.1 & 0.1 & 1.1 & 63.0 & 1.3 & 89.9 / 1.4\\
 & Blended & 99.9 & 92.1 & 8.8 & 0.7 & 8.8 [0.0, 35.7] 3p0m & 99.9 / 99.9 & 0.0 & 30.6 & 1.5 & 0.3 & 100.0 / 0.3\\
\bottomrule
\end{tabular}
\caption{Generality (X9b): test ASR (\%), mean over 3 seeds; for the layer-selected edit, the t-based 95\% interval (clipped) and how often it chose pen (p) or mid (m). C10 / C100: CIFAR-10 / CIFAR-100 ResNet-18; VGG: VGG-16-BN. ``retr.'': retraining on the remainder. ``clean'': the original's clean test accuracy; the layer-selected edit changes it by at most 0.7 points (proj-mid on CIFAR-100 costs up to 1.9). ``detect'': detect-then-retrain R1 (1.5 times the number of poisons flagged). VGG-16-BN: pen = the 512-d input of the read-out, mid = the output of conv block 3 before its max-pool; fixed before running. Step 1: validation ASR after one FT-relearn step with 8 known poisons, for proj-pen (100\% known) and the clean model. SVHN uses 20 epochs and no flip; CIFAR-100 a 100-way read-out.}\label{tab:general}
\end{table*}

\begin{table}[t]
\centering\footnotesize
\setlength{\tabcolsep}{3pt}
\begin{tabular}{@{}lccccc@{}}
\toprule
$k$ & $R^2$ & $\frac{t_Q}{1+t_Q}$ & ratio & $\frac{\pi\gamma_{k+1}}{1+\pi\gamma_{k+1}}$ & held-out $R^2$\\\midrule
0 (none) & 0.814 & 0.953 & 1.17 & -- & 0.675\\
1 & 0.1216 & 0.2504 & 2.06 & 0.700 & 0.0875\\
2 & 0.1047 & 0.2152 & 2.05 & 0.670 & 0.0738\\
4 & 0.0409 & 0.0695 & 1.70 & 0.584 & 0.0207\\
8 & 0.0296 & 0.0455 & 1.54 & 0.502 & 0.0127\\
16 & 0.0193 & 0.0269 & 1.40 & 0.380 & 0.0104\\
32 & 0.0068 & 0.0081 & 1.19 & 0.251 & 0.0091\\
64 & 0.0036 & 0.0040 & 1.13 & 0.164 & 0.0072\\
128 & 0.0016 & 0.0018 & 1.08 & 0.108 & 0.0026\\
256 & 0.0007 & 0.0007 & 1.04 & 0.060 & 0.0017\\
$\delta$ only & $<10^{-25}$ & $<10^{-29}$ & -- & -- & -0.0007\\
\bottomrule
\end{tabular}
\caption{\Cref{thm:probe}(b) at ResNet scale (X9c): forgetting class 3 (cat, $\pi=0.099$) at the penultimate layer of CIFAR-10 ResNet-18 ($p=512$, $\Sigma_R$ of full rank), pseudo-inverse whitening, mean of 3 originals. $R^2$: in-sample least-squares $R^2$ of the class indicator after proj$(k)$; ratio = tight bound / measured. Largest $R^2$ minus bound over all $k$ and seeds: $1.4\cdot10^{-26}$ (floating point). The bound is for the in-sample probe; held-out $R^2$ can exceed it ($k\ge32$).}\label{tab:thm5b}
\end{table}

\begin{table}[t]
\centering\footnotesize
\setlength{\tabcolsep}{3pt}
\begin{tabular}{@{}lcccc@{}}
\toprule
method & retain & forget & head-relearn & held-out $R^2$\\\midrule
original & 93.8 & 83.8 & 85.1 & 0.675\\
retrain (5 seeds) & 94.4 & 0.0 & 65.8 [64.6, 67.2] & 0.394\\
proj$(k^\ast)$, $k^\ast=128$ & 94.5 & 3.3 & 0.0 & 0.003\\
proj$(\delta)$ & 94.7 & 5.1 & 0.0 & -0.001\\
FT & 94.7 & 7.2 & 83.3 & 0.607\\
SalUn (simpl.) & 94.1 & 0.0 & 87.6 & 0.628\\
\bottomrule
\end{tabular}
\caption{Class-3 forgetting on CIFAR-10 ResNet-18 (X9c), mean of 3 originals (band: 5 retrained seeds, range in brackets). Test retain and forget-class accuracy (\%); head-relearn: validation forget-class accuracy after a ridge refit of the read-out on all 45k fit images with true labels; held-out $R^2$ of the class indicator at the penultimate layer. proj sits below the band (over-unlearning), FT and SalUn above it.}\label{tab:class5}
\end{table}

\begin{table*}[t]
\centering\footnotesize
\setlength{\tabcolsep}{2.6pt}
\begin{tabular}{@{}r|ccc|cc|c|ccccc|cccc|cc@{}}
\toprule
 & \multicolumn{3}{c|}{$\|\delta\|^2/(r/m)$} & \multicolumn{2}{c|}{$\gamma_1$/edge} & ovl. & \multicolumn{5}{c|}{loss AUC vs non-members} & \multicolumn{4}{c|}{per-example AUC} & \multicolumn{2}{c}{post hoc}\\
$m$ & F & R & N & F & R & top-8 & orig. & band & proj & FT & SalUn & orig. & proj & FT & SalUn & F dirs & R dirs\\\midrule
100 & 0.90 & 1.02 & 1.02 & -- & -- & 0.014 & 0.528 & 0.486 & 0.449 & 0.527 & 0.057 & 0.627 & 0.437 & 0.611 & 0.051 & 0.18 & 0.52\\
1{,}000 & 1.08 & 0.96 & 1.04 & 1.04 & 1.04 & 0.017 & 0.538 & 0.497 & 0.464 & 0.525 & 0.410 & 0.611 & 0.373 & 0.418 & 0.064 & 0.37 & 0.49\\
5{,}000 & 1.08 & 0.90 & 1.28 & 1.06 & 0.97 & 0.018 & 0.552 & 0.513 & 0.521 & 0.540 & 0.503 & 0.686 & 0.493 & 0.482 & 0.068 & 0.49 & 0.56\\
\bottomrule
\end{tabular}
\caption{\Cref{thm:probe}(c) at ResNet scale (X9c): random forget sets of $m$ fit images, CIFAR-10 ResNet-18 penultimate layer ($r=\mathrm{rank}\,\Sigma_R=512$), mean of 3 originals. Left: statistics of the original model's features for the forget set (F), a retained set of the same size (R) and $m$ non-members (N, validation images; for $m=5{,}000$ all of them), relative to the null value $r/m$ of population whitening and the upper Marchenko--Pastur edge $(1+\sqrt{r/m})^2$ (outside the full-rank regime $m\ge r$ of \cref{thm:probe}(c) for $m<r$). Under an independent Gaussian reference model ($n_R>r+2$) with the empirical (maximum-likelihood) retain whitening used here, the expected F and N value is $(r/m)(1+m/n_R)\,n_R/(n_R-r-2)$, about $1.01$, $1.03$ and $1.14$ times $r/m$ for $m=100$, $1{,}000$, $5{,}000$ ($n_R=45{,}000-m$); ovl.\ = $\|U_8^\top U_8'\|_F^2/8$ between the F and R top-8 directions (Haar: $8/512=0.016$). Middle: membership. Loss AUC: forget set vs non-members; per-example AUC: each forget image's margin shift against the retrain band (0.5 = chance-level ranking under this score; band members score 0.41--0.59 at $m=100$). proj uses $k^\ast$ chosen by distance to the band ($4$--$16$ at $m=100$, $32$ otherwise) and costs 1--1.5 test points at $m\ge1{,}000$. Right, post hoc and not pre-registered: per-example AUC after removing the top 32 directions of the forget set (F dirs) or of a retained set of the same size (R dirs).}\label{tab:rand5}
\end{table*}

\begin{table}[t]
\centering\footnotesize
\setlength{\tabcolsep}{3pt}
\begin{tabular}{@{}lcc@{}}
\toprule
 & DINOv2-B & written\\\midrule
test acc.\ round 0 / after 3 rounds & 98.25 / 98.47 & 49.17 / 45.50\\
exact (same $h^\star$, labels, $W$) & 240/240 & 240/240\\
max rel.\ $|W-W_{\mathrm{rerun}}|$ & $7.6\cdot10^{-13}$ & $5.5\cdot10^{-12}$\\
Select cert.\ (\cref{thm:pipeline}b) $L$/$U$/$V$ & 0/100/100\% & 0/100/100\%\\
\quad coarse check, decayed margins & 0/92/90\% & 0/83/85\%\\
$h^\star$ changes ($L$ deletions) & 6\% & 20\%\\
worst-case Augment cert.\ passes & 9.1\% & 9.1\%\\
pool relabelled per round & 18--20\% & 45--84\%\\
pseudo-label flips per round & 6.3/3.7/3.4 & 662/641/684\\
certified path / rerun (median s) & 0.033/0.65 & 0.20/0.94$^\ast$\\
\bottomrule
\end{tabular}
\caption{X2 on CIFAR-10: 220 single deletions, then 20 batches of 10, each checked against a full rerun. Pass rates are over single deletions. The written arm uses the data-free feature scale (X12; $^\ast$3 threads, the DINOv2 arm 14). The earlier written run with a scale computed from 2,000 training images gave the same exactness (240/240, $2.8\cdot10^{-12}$; 49.11 / 45.39\%), but only relative to that frozen scale.}\label{tab:x2}
\end{table}

\begin{table}[t]
\centering\footnotesize
\setlength{\tabcolsep}{2.5pt}
\begin{tabular}{@{}llrrrrr@{}}
\toprule
arm & req. & certified & lean & cached & rerun & U rescored\\\midrule
written & L & 237 & 191 & 171 & 484 & 100 / 100\%\\
 & U & 107 & 81 & 111 & 496 & 46 / 99\%\\
 & V & 8 & 6 & 4 & 499 & 0 / 100\%\\
 & batch & 237 & 190 & 172 & 474 & 98 / 100\%\\\midrule
DINOv2-B & L & 275 & 179 & 309 & 976 & 38 / 85\%\\
 & U & 23 & 20 & 177 & 971 & 3 / 85\%\\
 & V & 17 & 14 & 6 & 977 & 0 / 82\%\\
 & batch & 289 & 197 & 312 & 959 & 42 / 100\%\\
\bottomrule
\end{tabular}
\caption{Matched caching baseline for X2 (X8a, CIFAR-10, the same 240 requests). Median latency (ms, 6 threads) per request type. certified: X2's path as released; lean: the same without its a-priori diagnostics (the path quoted in \cref{sec:x2}); cached: the same stored statistics and re-solves, but every validation and pool row rescored, no margins; rerun: from scratch. All four paths are exact on 240/240 requests in both arms. This written arm still uses the earlier feature scale computed from 2,000 training images, so its exactness is relative to that frozen scale; \cref{tab:x2gen} repeats the same paths with the data-free scale. Last column: mean fraction of the pool rescored per round under the decayed margins of X2 and under margins anchored to one snapshot per round. Anchoring rescores more because a decayed margin is refreshed whenever its point is rescored, while a snapshot margin is compared with all drift since the snapshot.}\label{tab:x2cache}
\end{table}

\begin{table*}[t]
\centering\footnotesize
\setlength{\tabcolsep}{3pt}
\begin{tabular}{@{}lrrrrrrrrrrrrr@{}}
\toprule
 & & & & & & & & \multicolumn{3}{c}{$L$ request (ms)} & \multicolumn{3}{c}{$U$ request (ms)}\\\cmidrule(lr){9-11}\cmidrule(l){12-14}
setting & $C$ & $|L|$ & exact & max rel.\ $\Delta W$ & $\lambda^\star$ & relab. & flips & lean & cached & rerun & lean & cached & rerun\\\midrule
C100 DINOv2-B & 100 & 1{,}000 & 240/240 & $1.4\cdot10^{-13}$ & 1000 & 0.47 & 122 & 1080 & 1005 & 3090 & 190 & 552 & 3081\\
TinyIN DINOv2-B & 200 & 2{,}000 & 240/240 & $3.7\cdot10^{-13}$ & 300 & 0.44 & 87 & 2571 & 2866 & 8399 & 441 & 1758 & 8240\\
C10 DINOv2-B, order 2 & 10 & 1{,}000 & 240/240 & $8.0\cdot10^{-13}$ & 1000 & 0.19 & 4 & 404 & 691 & 2608 & 42 & 346 & 2642\\
C10 DINOv2-B, order 3 & 10 & 1{,}000 & 240/240 & $6.8\cdot10^{-13}$ & 1000 & 0.19 & 4 & 345 & 525 & 2289 & 31 & 256 & 2279\\
C10 DINOv2-B, order 4 & 10 & 1{,}000 & 240/240 & $1.4\cdot10^{-12}$ & 1000 & 0.23 & 9 & 367 & 513 & 2271 & 33 & 249 & 2246\\
C10 written, order 2 & 10 & 1{,}000 & 240/240 & $5.8\cdot10^{-12}$ & 100 & 0.67 & 856 & 462 & 471 & 1361 & 170 & 261 & 1351\\
C10 written, order 3 & 10 & 1{,}000 & 240/240 & $5.2\cdot10^{-12}$ & 100 & 0.67 & 414 & 488 & 468 & 1355 & 187 & 289 & 1338\\
C10 written, order 4 & 10 & 1{,}000 & 240/240 & $4.4\cdot10^{-12}$ & 100 & 0.66 & 663 & 500 & 462 & 1353 & 182 & 284 & 1346\\
\bottomrule
\end{tabular}
\caption{X2 on more data (X10a): the same pipeline and request mix (220 single deletions, then 20 batches of 10) on CIFAR-100 and Tiny-ImageNet (10 labels per class) and on three new deletion orders for both CIFAR-10 arms (X2's $L$, $U$, $V$). Every request is checked against a full rerun: 1920/1{,}920 exact (same $h^\star$, pseudo-labels and weights), and the final test predictions are identical in all 8 runs. The written rows use the data-free feature scale (X12). relab.: mean fraction of $U$ rescored per round; flips: realised pseudo-label changes per round (single deletions). Latency: median, 2 threads per run (not comparable with \cref{tab:x2,tab:x2cache}).}\label{tab:x2gen}
\end{table*}

\begin{table}[t]
\centering\footnotesize
\setlength{\tabcolsep}{1.6pt}
\begin{tabular}{@{}llccc@{}}
\toprule
head & method (label) & AUC [95\% CI] & \multicolumn{2}{c}{TPR at FPR}\\
 & & & 1\% & 0.1\%\\\midrule
ridge & original & 0.813 [0.811, 0.815] & 0.196 & 0.080\\
ridge & exact deletion (exact) & 0.501 [0.498, 0.505] & 0.010 & 0.001\\
ridge & retrain (exact) & 0.501 [0.498, 0.505] & 0.010 & 0.001\\
ridge & proj$(8)$ (heur.) & 0.668 [0.665, 0.671] & 0.116 & 0.047\\
softmax & original & 0.554 [0.551, 0.558] & 0.027 & 0.011\\
softmax & retrain (exact) & 0.501 [0.497, 0.504] & 0.011 & 0.001\\
softmax & FT (heur.) & 0.515 [0.512, 0.518] & 0.016 & 0.004\\
softmax & NegGrad+ (heur.) & 0.523 [0.520, 0.527] & 0.034 & 0.023\\
softmax & SCRUB (heur.) & 0.516 [0.513, 0.520] & 0.025 & 0.014\\\midrule
ridge & before/after, margin & 0.790 [0.788, 0.793] & 0.371 & 0.192\\
ridge & before/after, loss & 0.953 [0.952, 0.954] & 0.333 & 0.107\\
\bottomrule
\end{tabular}
\caption{X3. Per-example membership attack on 128 target heads. Validation accuracy is $99.4$--$99.5\%$ for every head except NegGrad+ ($97.2\%$) and SCRUB ($98.6\%$). Last two rows: an attacker who sees each head before and after the exact deletion of its forget set compares margins or squared losses (X8c; all 256 heads, cluster-bootstrap CI over heads; the margin score gives 0.792 on the 128 targets alone).}\label{tab:x3}
\end{table}

\begin{table*}[t]
\centering\footnotesize
\setlength{\tabcolsep}{3pt}
\begin{tabular}{@{}lrcccccc@{}}
\toprule
 & & & & \multicolumn{2}{c}{TPR at 0.1\% FPR} & & loss AUC\\
setting & $B$ & margin AUC & loss AUC & loss & label & label AUC & vs retained\\\midrule
C10 DINOv2-L (X3) & 1 & 0.831 [0.814, 0.847] & 0.9998 [0.9997, 0.9998] & 0.963 & 0.004 & 0.502 & 1.000\\
\quad U-LiRA 0.813 / 0.501 & 10 & 0.820 [0.805, 0.835] & 0.9974 [0.9970, 0.9977] & 0.615 & 0.005 & 0.503 & 0.999\\
 & 50 & 0.817 [0.811, 0.823] & 0.987 [0.987, 0.988] & 0.278 & 0.005 & 0.502 & 0.997\\
 & 250 & 0.790 [0.788, 0.793] & 0.953 [0.952, 0.954] & 0.107 & 0.006 & 0.503 & 0.987\\\midrule
C10 CLIP-B/32 & 1 & 0.892 [0.881, 0.904] & 0.9990 [0.9989, 0.9992] & 0.798 & 0.020 & 0.510 & 0.999\\
\quad U-LiRA 0.726 / 0.501 & 250 & 0.800 [0.798, 0.801] & 0.887 [0.886, 0.887] & 0.131 & 0.000 & 0.510 & 0.916\\
C10 DINOv2-B & 1 & 0.830 [0.815, 0.845] & 0.9997 [0.9997, 0.9998] & 0.950 & 0.015 & 0.508 & 1.000\\
\quad U-LiRA 0.804 / 0.502 & 250 & 0.790 [0.789, 0.792] & 0.948 [0.947, 0.948] & 0.128 & 0.013 & 0.506 & 0.983\\
C100 DINOv2-L & 1 & 0.981 [0.975, 0.986] & 0.9999 [0.9999, 1.0000] & 0.995 & 0.081 & 0.541 & 1.000\\
\quad U-LiRA 0.948 / 0.500 & 250 & 0.955 [0.955, 0.956] & 0.984 [0.984, 0.985] & 0.358 & 0.000 & 0.541 & 0.999\\
\bottomrule
\end{tabular}
\caption{Before/after attack on exactly deleted ridge heads (X8c, X10b). Each head is fitted on a random half (2,500) of a 5,000-image fit-split pool and deletes $B$ of its points in one exact batch ($B=250$ is X3's forget set; $B=1$: 10 or 5 single-point requests per head). Scores: change of the margin (X3's attack), of the squared loss, or of correctness (label-only output). Positives: deleted points; negatives: all other pool points (last column: retained members). AUC with 95\% cluster-bootstrap CIs over heads (256 for the X3 setting, 512 otherwise). U-LiRA: AUC of X3's per-example attack on the original / on the exactly deleted head, which never sees the earlier release.}\label{tab:x3batch}
\end{table*}

\begin{table*}[t]
\centering\footnotesize
\setlength{\tabcolsep}{3pt}
\begin{tabular}{@{}lcccccc@{}}
\toprule
 & \multicolumn{2}{c}{refactor, $\rho=10^{-6}$ / $10^{-3}$} & \multicolumn{2}{c}{chained SM, $\rho=10^{-6}$ / $10^{-3}$} & batch 10 & batch 100\\\midrule
C10 CLIP-B/32 & 159 [128, 174] & 147 [128, 158] & 662 [476, 693] & 573 [412, 653] & 65 [63, 71] & 83 [27, 100]\\
C10 DINOv2-B & 71 [69, 71] & 70 [67, 71] & 231 [230, 238] & 229 [219, 231] & 63 [56, 69] & 60 [57, 61]\\
C10 DINOv2-L & 53 [52, 54] & 52 [50, 53] & 159 [154, 161] & 154 [150, 156] & 52 [28, 57] & 47 [35, 48]\\
C100 DINOv2-B & 58 [55, 61] & 56 [53, 59] & 228 [214, 232] & 218 [215, 233] & 61 [46, 65] & 51 [33, 59]\\
TinyIN DINOv2-B & 96 [93, 100] & 96 [90, 107] & 438 [413, 499] & 445 [411, 479] & 98 [91, 112] & 77 [58, 93]\\
C10 written & 113 [94, 120] & 100 [92, 102] & 424 [314, 461] & 363 [310, 382] & 73 [39, 103] & 68 [26, 71]\\
\bottomrule
\end{tabular}
\caption{X1 timings repeated 5 times (X8b; 6 threads, first 200 of X1's 1,000 deletions per repetition): speed-up over retraining, median [min, max] over repetitions. Original runs with $\lambda=\rho\cdot\mathrm{mean\,diag}\,G(D)$ frozen from the fit data; X12 reruns the written arm with a data-free $\lambda$. Weights after 200 deletions equal a retrain within $10^{-11}$ in every repetition. Scaling slopes (log time vs log size), median [min, max]: $|F|$ 0.55 [0.52, 0.58], $n$ 0.06 [-0.05, 0.20], $d$ 1.66 [1.29, 1.89]. Absolute speed-ups differ from \cref{tab:x1} (14 threads, one run); the ratios within a run are what is compared.}\label{tab:timing}
\end{table*}

\begin{table}[h]
\centering\small
\begin{tabular}{@{}p{5.2cm}p{4.6cm}@{}}
\toprule
Method & Label\\\midrule
Ridge on written or frozen features; the whole pipeline (Thm.~\ref{thm:pipeline}); SISA with fixed shards (our X1) & exact, deterministic, safe under adaptive requests (Prop.~\ref{prop:adaptive})\\
Retraining a network with a fresh random seed (X4, X7) & exact in distribution, safe under adaptive requests (randomness drawn after the request; \citealp[Sec.~1.2]{gupta2021adaptive})\\
Sampled centres with resampling (Thm.~\ref{thm:centres}) & exact in distribution, non-adaptive requests only\\
Guo et al.\ certified removal & certified $(\varepsilon,\delta)$\\
Detect-then-retrain (X8, X9) & exact for the request it builds (known plus flagged points), heuristic for the unknown poisons\\
proj$(k)$, LEACE edit, DAMP-lite, delete-unit, refit, FT, NegGrad+, RL, SCRUB, SSD, $\ell_1$-sparse, SalUn, ABL & heuristic (proj$(k)$ with unregularised whitening also carries the probe bound of Thm.~\ref{thm:probe})\\
\bottomrule
\end{tabular}
\caption{The label of every method. ``Exact'' for frozen public backbones means exact relative to the downstream data only. SISA with random shard assignment would be exact only in distribution.}\label{tab:labels}
\end{table}

\begin{table}[h]
\centering\small
\setlength{\tabcolsep}{3pt}
\begin{tabular}{@{}lrrrrrr@{}}
\toprule
features & $d$ & test & retrain & refactor & SM & $3N/d$\\\midrule
C10 CLIP-B/32 & 513 & 96.2 & 0.050 s & 88$\times$ & 301$\times$ & 263\\
C10 DINOv2-B & 1537 & 98.7 & 0.343 s & 57$\times$ & 158$\times$ & 88\\
C10 DINOv2-L & 2049 & 99.4 & 0.908 s & 58$\times$ & 149$\times$ & 66\\
C100 DINOv2-B & 1537 & 90.5 & 0.370 s & 55$\times$ & 180$\times$ & 88\\
TinyIN DINOv2-B & 1537 & 86.4 & 0.879 s & 99$\times$ & 342$\times$ & 185\\
C10 written & 1025 & 66.0 & 0.184 s & 101$\times$ & 409$\times$ & 132\\
\bottomrule
\end{tabular}
\caption{X1. Test accuracy (\%) of the ridge head, retraining time, and single-deletion speed-up of the refactorisation and chained Sherman--Morrison (SM) paths ($\rho=10^{-6}$, i.e.\ $\lambda=10^{-6}\cdot\mathrm{mean\,diag}\,G(D)$ frozen from the fit data; $N=45{,}000$; TinyIN $N=95{,}000$, 200 classes). Timings: one run, median over deletions. Every run: 100\% identical test predictions after 1,000 deletions. The ranges quoted in \cref{sec:x1} ($52$--$70\times$, $158$--$193\times$ at $d=1{,}537$, $N=45{,}000$) cover both $\rho=10^{-6}$ and $10^{-3}$. Chained SM drifts only $0.9$--$9.6\times$ more than refactorisation, not the predicted $\ge10\times$. These are the original runs with the frozen $\lambda$; X12 reruns the written row with a data-free $\lambda$.}\label{tab:x1}
\end{table}

\begin{table}[t]
\centering\small
\setlength{\tabcolsep}{3pt}
\begin{tabular}{@{}lcc@{}}
\toprule
method (label) & test acc.\ (\%) & s / deletion\\\midrule
ridge, exact downdate (exact) & 96.18 & 0.00057$^\ast$\\
logistic, no noise (not cert.) & 96.73 & 8.0 (retrain)\\
CR, $T{=}100$, $\lambda$ tuned w/o noise & 9.96 $\pm$ 1.88 & 0.67\\
CR, $T{=}1000$, $\lambda$ tuned w/o noise & 10.16 $\pm$ 1.37 & 0.67\\
CR, $T{=}1000$, $\lambda$ tuned w/ noise & 94.76 $\pm$ 0.05 & 0.66\\
\quad same, Alg.~2 run online (X8b) & 94.76 [94.72, 94.84] & 0.86$^\dagger$\\
\midrule
C10 CLIP: single ridge (exact) & 96.16 & 0.0011\\
\quad SISA $S=5$ (exact) & 96.13 & 0.0114\\
\quad SISA $S=10$ (exact) & 96.12 & 0.0062\\
\quad SISA $S=20$ (exact) & 96.12 & 0.0034\\
C10 DINOv2-B: single ridge (exact) & 98.65 & 0.0068\\
\quad SISA $S=5$ (exact) & 98.67 & 0.0852\\
\quad SISA $S=10$ (exact) & 98.63 & 0.0467\\
\quad SISA $S=20$ (exact) & 98.58 & 0.0276\\
C100 DINOv2-B: single ridge (exact) & 90.46 & 0.0083\\
\quad SISA $S=5$ (exact) & 90.55 & 0.0956\\
\quad SISA $S=10$ (exact) & 90.48 & 0.0500\\
\quad SISA $S=20$ (exact) & 90.34 & 0.0280\\
\bottomrule
\end{tabular}
\caption{Competitors on identical frozen features (CIFAR-10 CLIP-B/32 for certified removal, $\varepsilon=1$, $\delta=10^{-4}$ per one-vs-rest model, first two CR rows: $\lambda=10^{-6}$ chosen by noiseless validation accuracy; third CR row and Alg.~2 row: $\lambda=10^{-3}$ chosen by noisy validation accuracy, a declared follow-up; the noise for budget 1,000 is extrapolated linearly from 100 or 50 simulated removals, and accuracy is that of the noisy model before any removal, so these $T{=}1000$ rows are a prospective calibration). The Alg.~2 row runs the online budget and retrain trigger of \citet{guo2020certified} through all 1,000 removals with that noise (median [min, max] over 3 noise seeds, test accuracy after the last removal): no head reaches its budget, so it never retrains. $^\dagger$Amortised seconds per removal at 6 threads (the other timings: 14 threads). $^\ast$Median refactorisation time of X1 on the same features at $\rho=10^{-6}$. X1's $\rho$ scales the mean Gram diagonal, while CR's $\lambda$ weights $n\|w\|^2/2$; the two are different parameterisations. SISA averages the logits of $S$ ridge shards; its shards are fixed, so it is exact and adaptive-safe (\cref{prop:adaptive}).}\label{tab:x1b}
\end{table}

\begin{figure*}[h]
\centering
\includegraphics[width=\textwidth]{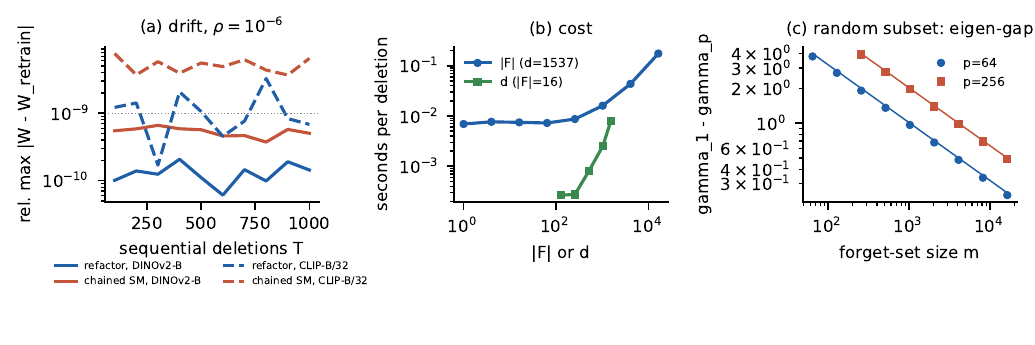}
\caption{(a) X1 drift to retraining over 1,000 sequential deletions at $\rho=10^{-6}$, i.e.\ $\lambda=10^{-6}\cdot\mathrm{mean\,diag}\,G(D)$ frozen from the fit data (original runs; dotted: $10^{-9}$). (b) Seconds per exact deletion against $|F|$ and $d$ (DINOv2-B, $N=45{,}000$). (c) X6: generalised eigen-gap of a random forget set against $4\sqrt{p/m}$ (lines).}\label{fig:x1x6}
\end{figure*}

\begin{table*}[t]
\centering\small
\setlength{\tabcolsep}{3.5pt}
\begin{tabular}{@{}llcccccccc@{}}
\toprule
 & & \multicolumn{4}{c}{MNIST} & \multicolumn{4}{c}{Fashion-MNIST}\\\cmidrule(lr){3-6}\cmidrule(l){7-10}
method & label & FA & relearn & FM-rec. & $R^2_h$ & FA & relearn & FM-rec. & $R^2_h$\\\midrule
original & -- & 0.988 & 0.975 & 0.971 & 0.837 & 0.872 & 0.936 & 0.877 & 0.734\\
retrain$^\ddagger$ & exact (distr.) & 0.000 & \textbf{0.953} & \textbf{0.586} & \textbf{0.729} & 0.000 & \textbf{0.918} & \textbf{0.426} & \textbf{0.627}\\
\quad band range & & -- & [0.94, 0.96] & [0.47, 0.71] & [0.66, 0.75] & -- & [0.90, 0.93] & [0.27, 0.56] & [0.61, 0.64]\\\midrule
FT & heur. & 0.000 & 0.977 & 0.841 & 0.787 & 0.000 & 0.936 & 0.849 & 0.725\\
NegGrad+ & heur. & 0.000 & 0.968 & 0.933 & 0.819 & 0.006 & 0.930 & 0.848 & 0.732\\
RL & heur. & 0.000 & 0.985 & 0.965 & 0.820 & 0.000 & 0.951 & 0.863 & 0.726\\
SCRUB & heur. & 0.324 & 0.972 & 0.920 & 0.796 & 0.259 & \textbf{0.925} & 0.831 & 0.714\\
SSD & heur. & 0.367 & 0.974 & 0.971 & 0.837 & 0.091 & 0.937 & 0.873 & 0.727\\
$\ell_1$-sparse & heur. & 0.000 & \textbf{0.949} & 0.863 & 0.771 & 0.000 & \textbf{0.909} & 0.633 & 0.649\\
SalUn (simpl.) & heur. & 0.000 & 0.976 & 0.953 & 0.799 & 0.000 & 0.936 & 0.865 & 0.726\\
delete-unit & heur. & 0.000 & 0.975 & 0.971 & 0.837 & 0.000 & 0.936 & 0.877 & 0.734\\
refit fc2 & heur. & 0.000 & 0.975 & 0.971 & 0.837 & 0.000 & 0.936 & 0.877 & 0.734\\
proj$(k)$+mask & heur.$^\dagger$ & 0.000 & 0.881 & 0.162 & 0.461 & 0.000 & \textbf{0.929} & 0.182 & 0.588\\
proj$(k)$+refit & heur.$^\dagger$ & 0.000 & 0.786 & 0.189 & 0.384 & 0.000 & \textbf{0.903} & \textbf{0.310} & 0.526\\
proj$(k^\ast)$+refit & heur.$^\dagger$ & 0.000 & 0.966 & 0.111 & 0.570 & 0.000 & \textbf{0.929} & 0.182 & 0.588\\
LEACE edit+refit & heur. & 0.000 & 0.794 & 0.045 & 0.396 & 0.000 & 0.891 & 0.102 & 0.522\\
DAMP-lite+mask & heur. & 0.000 & 0.808 & 0.013 & 0.377 & 0.000 & 0.867 & 0.062 & 0.469\\
\bottomrule
\end{tabular}
\caption{X4, class forgetting (class 3), mean over 3 original seeds; band = range of 5 retrained seeds. $^\ddagger$Mean of the first 3 of the 5 band seeds, one paired with each original seed. FA: test forget accuracy. relearn: forget accuracy after a ridge re-fit of the last layer on all classes. FM-rec.: forget accuracy after one linear map from the model's penultimate features to the original's \citep{jung2025one}. $R^2_h$: held-out least-squares $R^2$ of the forget indicator at the penultimate layer. Bold: inside the retrain band. $^\dagger$The unregularised version of this edit carries the probe bound of \cref{thm:probe}(b) at the edited layer; the regularised edit reported here does not. $k$ selected on validation by output distance to the band ($k=4$/$16$ with mask, $1$/$2$ with refit); $k^\ast$ selected to match the band's relearn probe ($32$/$16$).}\label{tab:x4}
\end{table*}

\begin{table*}[t]
\centering\footnotesize
\setlength{\tabcolsep}{2.4pt}
\begin{tabular}{@{}lccccccccccccccccc@{}}
\toprule
 & & \multicolumn{4}{c}{BadNets} & \multicolumn{4}{c}{Blended} & \multicolumn{4}{c}{WaNet ($s=2$)} & \multicolumn{4}{c}{SIG}\\
 & clean & 100\% & 50\% & 25\% & 10\% & 100\% & 50\% & 25\% & 10\% & 100\% & 50\% & 25\% & 10\% & 100\% & 50\% & 25\% & 10\%\\\midrule
\multicolumn{18}{@{}l}{\emph{MNIST}, original ASR 100, 98, 59, 93\%}\\
proj$(k)$ & 98.2 & 0.6 & 0.9 & 1.6 & 3.7 & 0.4 & 0.5 & 1.4 & 6.1 & 7.0 & 7.9 & 11.4 & 18.0 & 0.4 & 0.5 & 0.7 & 1.2\\
retrain rem. & 98.2 & 0.1 & 97.9 & 98.9 & 98.7 & 0.1 & 92.0 & 96.1 & 95.9 & 0.5 & 42.3 & 50.7 & 63.2 & 0.1 & 54.5 & 90.7 & 93.6\\
FT & 98.8 & 0.6 & 99.7 & 99.9 & 99.7 & 0.3 & 99.6 & 97.1 & 95.8 & 0.9 & 70.4 & 80.8 & 85.5 & 0.2 & 96.5 & 99.2 & 97.6\\
NegGrad+ & 73.1 & 0.0 & 0.0 & 0.0 & 30.6 & 0.0 & 0.0 & 0.0 & 0.0 & 0.0 & 0.0 & 0.0 & 0.0 & 0.0 & 0.0 & 0.0 & 0.0\\
SCRUB & 98.7 & 0.0 & 99.1 & 99.7 & 99.7 & 0.2 & 97.1 & 99.2 & 99.1 & 5.8 & 58.6 & 69.8 & 71.0 & 0.3 & 95.1 & 96.6 & 96.5\\
SSD & 41.6 & 28.3 & 30.5 & 44.7 & 34.9 & 0.0 & 0.0 & 0.5 & 0.6 & 0.0 & 0.0 & 0.0 & 0.0 & 0.0 & 0.0 & 0.0 & 0.1\\
written head, exact & 94.3 & 0.4 & 88.3 & 93.5 & 95.0 & 0.1 & 89.4 & 97.1 & 98.5 & 1.3 & 3.8 & 5.9 & 7.9 & 0.5 & 5.2 & 11.5 & 16.1\\
\midrule
\multicolumn{18}{@{}l}{\emph{Fashion-MNIST}, original ASR 93, 77, 41, 72\%}\\
proj$(k)$ & 86.7 & 3.7 & 5.3 & 8.2 & 20.1 & 6.9 & 8.2 & 9.8 & 15.1 & 19.0 & 19.4 & 20.2 & 29.9 & 3.6 & 4.1 & 4.8 & 7.7\\
retrain rem. & 86.5 & 0.6 & 83.7 & 88.9 & 92.0 & 2.7 & 38.1 & 57.4 & 70.0 & 2.6 & 9.2 & 16.9 & 36.5 & 1.7 & 13.4 & 39.0 & 62.9\\
FT & 89.7 & 9.6 & 95.0 & 96.4 & 94.8 & 5.2 & 93.4 & 98.1 & 97.1 & 4.2 & 64.3 & 77.2 & 81.1 & 4.1 & 93.7 & 93.1 & 94.5\\
NegGrad+ & 72.4 & 3.1 & 3.5 & 3.7 & 4.7 & 0.0 & 0.0 & 0.0 & 0.0 & 0.0 & 0.0 & 0.0 & 0.0 & 0.0 & 0.0 & 0.0 & 0.0\\
SCRUB & 88.3 & 0.2 & 93.4 & 94.4 & 93.8 & 2.2 & 85.9 & 89.1 & 88.2 & 4.9 & 46.9 & 54.2 & 50.5 & 1.9 & 83.7 & 84.5 & 83.6\\
SSD & 74.2 & 43.0 & 43.1 & 49.2 & 53.1 & 32.2 & 26.4 & 30.7 & 24.1 & 41.7 & 41.5 & 34.3 & 37.2 & 38.4 & 33.7 & 26.4 & 40.5\\
written head, exact & 83.8 & 1.7 & 42.0 & 59.1 & 66.6 & 2.1 & 43.8 & 62.7 & 70.9 & 2.6 & 6.9 & 13.5 & 17.8 & 2.5 & 6.9 & 9.8 & 12.1\\
\bottomrule
\end{tabular}
\caption{X5, attack success rate (\%, test, mean of 3 seeds) after unlearning with 100, 50, 25 or 10\% of the 1\% poisons known; ``clean'' is mean clean test accuracy over the 16 cells. WaNet at the planned strength $s=0.5$ did not implant (original ASR $0.7\%$ MNIST, $3.0\%$ Fashion), so its columns show the declared follow-up at $s=2$; on Fashion-MNIST it implanted only to $41\%$, below the pre-registered $50\%$, so those cells are reported but not scored. GD baselines (FT, NegGrad+, SCRUB, SSD) use one configuration tuned on BadNets with all poisons known (seed 0), while $k$ of proj$(k)$ is chosen per cell; this favours the edit outside that cell. ``retrain rem.'' retrains on everything not known; the exact written head (fixed random ReLU features + ridge) equals its own retraining on the remainder in every cell.}\label{tab:x5}
\end{table*}

\begin{figure*}[h]
\centering
\includegraphics[width=\textwidth]{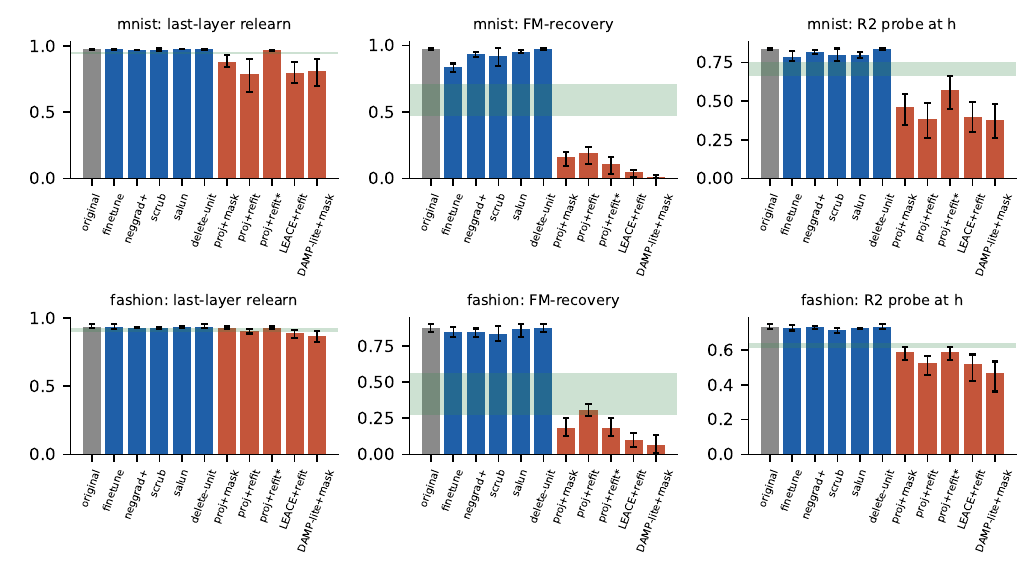}
\caption{X4 probes (mean over 3 seeds; bars: bootstrap over the 3 seeds, about the min--max seed range, not a 95\% CI; green: range of the 5-seed retrain band). Blue: gradient baselines; red: written edits; grey: original. $\ast$: $k$ matched to the band's relearn probe.}\label{fig:x4}
\end{figure*}

\clearpage
\section{Predictions written before each experiment}\label{app:prereg}
Before each experiment we wrote down what we expected, in a dated lab notebook that is part of the code release. The notebook has no per-entry time stamps or hashes, so the record is self-attested. \Cref{tab:prereg,tab:prereg2} list every prediction with its verdict. Of 131 predictions, 70 held, 40 partly failed and 21 failed.

{\small
\begin{longtable}{@{}p{1.0cm}p{6.4cm}p{1.9cm}p{6.4cm}@{}}
\caption{Pre-registration ledger, first round (X1--X7). Every prediction was written in the lab notebook before its run; this is self-attested (the notebook is included in the code release and has no per-entry time stamps). X1c and X5b are declared follow-ups, written after the main run exposed a problem. Verdict rule: \emph{held} if every part held, \emph{failed} if no part held, \emph{partly failed} otherwise; X3.2, X4.7 and X6.5 were regraded under this rule, as noted in their rows. This round: 43 scored predictions, 18 held, 13 partly failed, 12 failed. The X5b row for Fashion-MNIST is reported but not scored. Design change not in the plan: X2 sent 220 single deletions (adding 20 from $V$) instead of 200; no verdict depends on it. X7 (CIFAR-10 ResNet-18, GPU) was added last. The follow-up rounds X8--X10 and X12 (88 predictions) are summarised in the last rows and listed in \cref{tab:prereg2}. All rounds: 131 predictions, 70 held, 40 partly failed, 21 failed.}\label{tab:prereg}\\
\toprule
ID & prediction & verdict & outcome\\\midrule
\endfirsthead
\toprule
ID & prediction & verdict & outcome\\\midrule
\endhead
X1.1 & refactorisation within $10^{-9}$ of retraining after 1,000 deletions; 100\% identical predictions & held & $\le6.8\cdot10^{-10}$ in 12/12 runs, 100\% identical; one intermediate checkpoint reached $3.3\cdot10^{-9}$\\
X1.2 & chained Sherman--Morrison drift $\ge10\times$ refactorisation (small $\lambda$) & failed & $0.9$--$9.6\times$\\
X1.3 & single-deletion speed-up $>100\times$ at $n=45$k, $d=1{,}537$ & failed & refactorisation $52$--$70\times$; saturates near $3N/d$. Graded on refactorisation, the path the pipeline uses; chained Sherman--Morrison reaches $158$--$193\times$\\
X1.4 & certified removal loses $\ge1$ accuracy point at a 1,000-removal budget & held & accuracy collapses to $\approx10\%$ ($\lambda$ tuned without noise)\\
X1.5 & SISA loses $\ge0.5$ point at 20 shards, speed-up $<25\times$ & partly failed & loss $\le0.12$ point (failed); speed-up $13.5$--$13.8\times$ (held)\\
X2.1 & 100\% of pipeline deletions match the rerun to $10^{-9}$ & held & 240/240 in both arms, $\le2.8\cdot10^{-12}$\\
X2.2 & 0-1 Select certificate passes $\ge95\%$ of single deletions & failed & certificate of \cref{thm:pipeline}(b): 55\% in both arms (0\% for labelled-point deletions, 100\% for pool and validation); coarse check on decayed margins: 50\% / 54\%\\
X2.3 & worst-case Augment certificate passes $<20\%$ for $|U|\ge10^4$ & held & 9.1\% in both arms\\
X2.4 & relabel $<5\%$ of $U$ per round; $\le10$ flips per round; certified path $\ge5\times$ faster & failed & 18--84\% relabelled; flips 3--6 (DINOv2) but 393--472 (written); $20\times$ / $3\times$ faster\\
X2.5 & resampled centres indistinguishable non-adaptively, detected adaptively & held & $\chi^2$ $p=0.78$; adaptive overlap 6.00 vs 0.67, $p<10^{-300}$\\
X3.1 & exact head within $\pm0.02$ AUC of retrain baseline & held & identical (0.501)\\
X3.2 & original ridge and logistic heads AUC $\ge0.60$ & partly failed (regraded) & ridge 0.813 (held); softmax 0.554 (failed)\\
X3.3 & some GD baseline $\ge0.03$ AUC above its retrain baseline & failed & $+0.014$ to $+0.022$\\
X3.4 & proj$(k)$ on random 10\% gives no AUC reduction ($\ge$ original $-0.02$) & failed & lowered 0.813 $\to$ 0.668 (retrain 0.501)\\
X3.5 & before/after attack on the exact head AUC $\ge0.95$ & failed & 0.792 (0.812 vs retained members)\\
X4.1 & edited-layer $R^2\le\pi\gamma_{k+1}$ on every seed; held-out within 0.02 & held (vacuous) & always below; $\pi\gamma_{k+1}\ge1$ for $k\le32$; $\pi\|Q\delta\|^2$ informative; the spectral form $\pi\gamma_{k+1}/(1+\pi\gamma_{k+1})$ of \cref{thm:probe}(b) is $0.57$--$0.95$\\
X4.2 & proj along $\delta$ gives in-sample $R^2<0.01$ & held & $0.000$ on all 6 seeds\\
X4.3 & later-layer relearn above edited-layer value & held & $0.014\to0.46\to0.88$ (MNIST)\\
X4.4 & proj+refit below the band; probe-matched $k^\ast\in\{1,2\}$ & failed & below on MNIST, inside on Fashion; $k^\ast=32$ / $16$\\
X4.5 & FM-recovery restores GD baselines to within 5 points; proj inside or below band & partly failed & 4/7 (MNIST), 6/7 (Fashion); proj below/inside: held\\
X4.6 & proj ties DAMP-lite within 2 points of forget accuracy & held & both 0.000\\
X4.7 & padded model reproduces every output metric (max $|$logit diff$|<10^{-5}$) while keeping the original's probe values & partly failed (regraded) & max diff $0$ (held); probe $\ge$ original's, not equal ($0.857$ vs $0.837$ MNIST)\\
X6.1 & eigen-gap within 25\% of $4\sqrt{p/m}$ for $m\ge4p$ & held & ratio 0.955--0.987\\
X6.2 & Haar overlap within 20\% of $1/p$ & partly failed & pooled 1.11 / 1.15; 5 of 17 cells outside (as sampling noise predicts)\\
X6.3 & pre-edit in-sample $R^2\le\pi p/m$ (in expectation, within 25\%) & held & ratio 0.955--1.065, above 1 in some cells but within the stated 25\%\\
X6.4 & proj$(k)$ on random 10\% increases distance to retrain & held & 0.0033 $\to$ 0.022 ($k=8$) $\to$ 0.13 ($k=128$)\\
X6.5 & timing slopes: $\approx1$ in $|F|$, $\approx0$ in $n$, 2--3 in $d$ & partly failed (regraded) & $|F|$: 0.58 (failed); $n$: $-0.02$ (held); $d$: 1.39 (failed)\\
X1c.1 & certified removal with $\lambda$ tuned for the noisy model still loses $\ge1$ point & held & $94.76\%$ vs $96.18\%$\\
X5.1 & BadNets: proj ASR $\le1\%$ (all identified), $\le5\%$ (25\%); retrain on remainder $\ge50\%$ at 25\% & partly failed & MNIST $0.6$ / $1.6\%$ (held); Fashion $3.7$ / $8.2\%$ (failed); remainder $98.9$ / $88.9\%$\\
X5.2 & WaNet and SIG leave proj ASR $\ge50\%$ & failed & WaNet not implanted; SIG $0.4\%$ / $3.6\%$\\
X5.3 & Blended lands between BadNets and WaNet/SIG & failed & no ordering\\
X5.4 & exact written head equals retraining on the remainder in every cell & held & $\le4.0\cdot10^{-13}$ in 96 cells\\
X5b.1 & stronger WaNet: proj ASR $\ge50\%$ with all identified & failed & $7.0\%$ (MNIST)\\
X5b.2 & stronger WaNet: retrain on remainder $\ge50\%$ at 25\% & held (barely) & $50.7\%$\\
X5b.3 & exact head equals retraining on the remainder & held & identical\\
X5b, Fashion & implant check: original ASR $\ge50\%$, else not scored & not scored & implanted to $41.1\%$; proj $19.0$ / $20.2$ / $29.9\%$ vs retrain on remainder $2.6$ / $16.9$ / $36.5\%$ at 100 / 25 / 10\%\\
X7.0 & CIFAR ResNet-18: clean $\ge93\%$; ASR $\ge90\%$ (BadNets), $\ge80\%$ (Blended) & partly failed & clean $92.7\%$; ASR $94.5$ / $99.9\%$\\
X7.1 & retrain on remainder keeps ASR $\ge50\%$ at 25\% and 10\% known (BadNets, Blended) & held & $94.3$--$100\%$ (WaNet, not in the prediction: $69.6$--$76.7\%$)\\
X7.2 & proj-pen, all known: ASR $\le5\%$ (BadNets), $\le10\%$ (Blended); clean within 1 point & partly failed & $5.3\%$ (failed), $0.2\%$, clean $\pm0.1$\\
X7.3 & proj-pen, 10\% known: ASR $\le10\%$ (BadNets), $\le20\%$ (Blended) & partly failed & $42.3\%$ (failed), $9.9\%$\\
X7.4 & FT-relearn after proj-pen $\ge10$ points above the clean model & partly failed & BadNets $94.8$ vs $30.9$; Blended $99.7$ vs $93.9$\\
X7.5 & FT keeps $\ge50\%$ at 10\%; SalUn $\le10\%$ at 100\% but $\ge30\%$ at 10\% & partly failed & FT $93.8$ / $99.9$; SalUn $0.0$ and $\le0.1\%$\\
X7.6 & proj-mid not better than proj-pen at 10\% by $>5$ points & partly failed & Blended held; BadNets $1.4$ vs $42.3\%$\\
X7.7 & WaNet (if implanted): proj-pen ASR $\ge30\%$, all known & failed & implanted ($76.6\%$); $4.3\%$\\\midrule
X8 & 40 follow-ups to X1--X3 and X7 (\cref{tab:prereg2}) & 23 / 10 / 7 & held / partly failed / failed\\
X9 & 26: X7 at 5 seeds, backdoor generality, \cref{thm:probe}(b)(c) on ResNet-18 & 13 / 11 / 2 & held / partly failed / failed\\
X10 & 14: X2 and X3 on more data & 8 / 6 / 0 & held / partly failed / failed\\
X12 & 8: the written arm with a data-free feature scale & 8 / 0 / 0 & held / partly failed / failed\\
\bottomrule
\end{longtable}}

{\small
\begin{longtable}{@{}p{1.0cm}p{6.4cm}p{1.9cm}p{6.4cm}@{}}
\caption{Pre-registration ledger, follow-up rounds X8--X10 and X12, written after a review and before each round's runs (same rules as \cref{tab:prereg}). 88 predictions: 52 held, 27 partly failed, 9 failed (X8: 23 / 10 / 7; X9: 13 / 11 / 2; X10: 8 / 6 / 0; X12: 8 / 0 / 0). X8a.5, X8d.6 and X9c.6 were graded failed in the notebook although one of their parts held; they are regraded partly failed under the rule, as noted in their rows. Declared deviations, each written before the affected runs unless said otherwise: (i) X8d's control triggers were replaced by a rule fixed after a smoke test showed that the listed ones fire the implanted backdoor (\cref{app:exp}); (ii) X8e.5 was written after the main ABL runs, as an implementation check; (iii) X9b pre-registers X7's post hoc layer rule; (iv) X9c's control with directions from a retained set was added after the results and is not pre-registered or scored; (v) X10a's first CIFAR-10 run used the wrong number of labels (100 instead of 1,000), was stopped after 21 requests and is not scored, and the pre-registration misstated X2's labels per class. X8d's models for seeds 0--2 are re-created, not X7's own.}\label{tab:prereg2}\\
\toprule
ID & prediction & verdict & outcome\\\midrule
\endfirsthead
\toprule
ID & prediction & verdict & outcome\\\midrule
\endhead
X8a.1 & caching study: all four paths exact on 240/240 requests in both arms; certified path reproduces X2's flips & held & max rel.\ $W$ diff $2.9\cdot10^{-12}$; flips identical on 100\% of requests\\
X8a.2 & $L$ requests: lean certified within $1.5\times$ of the cached baseline (median) in both arms & partly failed & written 191 vs 171 ms (held); DINOv2 179 vs 309 ms, $1.73\times$ (failed)\\
X8a.3 & $U$ requests: lean $\ge2\times$ faster than cached on DINOv2, $<1.5\times$ on written & held & $8.9\times$ / $1.38\times$\\
X8a.4 & cached baseline $\ge3\times$ faster than the rerun for $L$, $U$, batch, both arms & partly failed & DINOv2 $3.2$ / $5.5$ / $3.1\times$ (held); written $2.8$ / $4.5$ / $2.8\times$\\
X8a.5 & anchored margins rescore $\ge1.5\times$ less of $U$ than decayed ones; anchored augment certificate $<20\%$ & partly failed (regraded) & anchored rescores more (DINOv2 0.84 vs 0.19, written 1.00 vs 0.66; failed); augment certificate 0\% (held)\\
X8a.6 & every scheme sound on every request & held & no flip outside a candidate set\\
X8a.7 & $c=\sqrt2$ lowers the decayed rescored fraction by $\ge20\%$ (DINOv2) & held & 0.209 $\to$ 0.164 ($-22\%$)\\
X8a.8 & anchored Select certificate passes at least as often as decayed on singles, $\le10\%$ on $L$ & partly failed & 0.5\% vs 50--54\% (failed); $L$: both 0\% (held)\\
X8a.9 & squared-loss Select from moments equals the row-by-row rerun on all 241 states & held & same $h^\star$ and labels; losses agree to $3\cdot10^{-12}$\\
X8b.1 & Guo Alg.~2: at most 1 retrain per head over 1,000 removals & held & 0 retrains (largest bound 0.81 of budget)\\
X8b.2 & final accuracy within $\pm0.5$ of X1c, $\ge1$ point below the exact head & held & $94.72$--$94.84\%$ vs $96.18\%$\\
X8b.3 & amortised speed-up $5$--$20\times$, below the exact downdate & held & $6.4\times$ [$5.4$, $6.7$] vs $147$--$159\times$\\
X8b.4 & 5 timing repeats: every per-deletion speed-up within $\pm30\%$ of its median ($n\ge45$k) & held & widest $-28\%$ (CLIP, SM)\\
X8b.5 & DINOv2-B refactor $<100\times$, SM $\ge100\times$ in all repeats & held & $67$--$71\times$; $219$--$238\times$\\
X8b.6 & slopes: $|$slope in $n|<0.2$, slope in $|F|<0.8$, every repeat & partly failed & $|F|$ held; $n$ in 4 of 5 (one 0.201)\\
X8c.1 & $B=250$ on X3's 128 targets reproduces AUC 0.792 & held & 0.7921\\
X8c.2 & margin AUC rises as $B$ falls ($\ge0.99$ at $B=1$, $\ge0.95$ at 10, $\ge0.88$ at 50) & failed & 0.831 / 0.820 / 0.817\\
X8c.3 & margin TPR at 0.1\% FPR $\ge0.80$ at $B=1$ & failed & 0.735\\
X8c.4 & loss-score AUC within $\pm0.03$ of margin AUC at every $B$ & failed & loss far stronger: 0.953 ($B=250$), 0.9998 ($B=1$)\\
X8c.5 & label-only AUC $\le0.52$, TPR at 0.1\% $\le0.02$ & held & 0.502--0.503; 0.004--0.006\\
X8c.6 & retained-member negatives give higher margin AUC than all others & held & 0.8311 vs 0.8310 ($B=1$) up to 0.810 vs 0.790\\
X8d.1 & pre-edit head-relearn $\ge90/90/65\%$; clean $\le5/70/5\%$; gap $\ge30$ for all triggers & held & $93.9/99.9/72.5$ vs $1.5/56.4/2.3$\\
X8d.2 & layer3 original-minus-clean gap $\ge20$ (BadNets, WaNet) & held & 76 / 39 points\\
X8d.3 & trigger-probe AUC at layer2 $\ge0.95$ for original and clean model & partly failed & Blended 1.000 both; BadNets clean 0.920\\
X8d.4 & re-created proj-pen (100\%): post-edit head-relearn $\le10\%$ / $\le40\%$ & held & $7.2$ / $28.2\%$\\
X8d.5 & re-created models within 5 points of X7's validation ASR & held & largest gap 4.2\\
X8d.6 & own trigger, step 10: edits $\ge50\%$, clean $\le10\%$ (BadNets, WaNet) & partly failed (regraded) & edits 88--99\% (held); clean 17.9 / 53.6\% (failed); the gap is at step 1 (86 vs 1.3\%)\\
X8d.7 & Blended: clean $\le20\%$ at step 10 and $\ge50\%$ at 100; proj-pen $\ge50\%$ at 10 & partly failed & clean 98.2\% at 10 (failed); 93.7\% at 100; proj-pen 100\%\\
X8d.8 & control trigger at step 10: original within 15 points of clean & partly failed & Blended held; BadNets 99.9 vs 48.0, WaNet 67.1 vs 48.2\\
X8d.9 & FT and SalUn relearn at least as fast as proj-pen (step 10) & failed & FT as fast on two; SalUn slower (BadNets 91.5 vs 99.0, WaNet 29.3 vs 92.7)\\
X8d.10 & detection AUROC $\ge0.95$ (BadNets, Blended), $\ge0.85$ (WaNet) & held & 0.995 / 1.000 / 0.984\\
X8d.11 & R1 recall $\ge80\%$ (BadNets, Blended) & held & 97.4 / 100\%\\
X8d.12 & detect-then-retrain R1 ASR $\le10\%$, clean within 0.5 of the clean retrain & held & $1.1$ / $0.2\%$; $92.9$ / $92.7\%$\\
X8d.13 & R1 beats the layer-selected edit on at most 1 trigger; relearns slower than every edit & partly failed & beats it on 2 (failed); slower relearn (held)\\
X8d.14 & R2 flags $\le5\%$ with recall $\ge85\%$ (BadNets, Blended) & held & 366 / 362 flagged, recall 90 / 89\% (but Blended ASR stays 89\%)\\
X8e.1 & ABL isolation precision $\ge80\%$ (BadNets, Blended) & failed & 0.2 / 14.0\%\\
X8e.2 & ABL ASR $\le10\%$ (BadNets), $\le20\%$ (Blended, WaNet) & failed & 95.1 / 21.5 / 78.4\%\\
X8e.3 & ABL clean accuracy within 2 points of the clean retrain & failed & 80.5--82.7 vs 92.8\%\\
X8e.4 & ABL not lower in ASR than R1 (BadNets, Blended) & held & 95.1 vs 1.1; 21.5 vs 0.2\\
X8e.5 & (after the main runs) ABL at 10\% poisons: precision $\ge90\%$, ASR $\le10\%$ & held & 100\%; 3.3\% (one seed)\\
X9a.1 & 5 seeds: layer-selected edit $\le10\%$ at 10\% known; retrain rem.\ and FT $\ge50\%$ & held & 1.5 / 7.8 / 5.5\%; 76.1--99.9\% (Blended interval [1.0, 14.6] crosses 10\%)\\
X9a.2 & proj-pen $\ge25\%$, proj-mid $\le5\%$ on BadNets at 10\% & held & 39.0 / 1.5\%\\
X9a.3 & $\ge80\%$ of 45 cells within 5 points of X7's 3-seed mean & held & 43 of 45\\
X9a.4 & upper end of the 95\% interval of the layer-selected edit $\le20\%$ & held & 1.7 / 14.6 / 7.3\%\\
X9a.5 & SalUn $\le2\%$ everywhere; its head-relearn $\ge75\%$ at 10\% & held & max 1.4\%; 95.0 / 99.9 / 79.4\%\\
X9a.6 & seeds 3--4 pre-edit probe as X8d.1; step-1 relearn $\ge50\%$ (proj-pen) and $\le5\%$ (clean) & held & 93.8 / 99.8 / 74.4 vs 1.4 / 59.2 / 2.5; step 1: 91.3 / 99.7 / 55.4 vs 1.2 / 1.4 / 1.6\\
X9a.7 & R1 $\le5\%$ (BadNets, Blended) and worse than the edit on WaNet & held & 1.0 / 0.2 / 17.1 vs 5.5\%\\
X9b.1 & implants (ASR $\ge80$--$95\%$) and clean accuracy thresholds in every setting & partly failed & all but CIFAR-100 BadNets (87.7 $<$ 90)\\
X9b.2 & retrain rem.\ and FT keep $\ge50\%$ at 25\% and 10\% known & held & 86.5--100\%\\
X9b.3 & layer-selected edit $\le10\%$ at 5\% poisons, $\le20\%$ at 0.5\% & partly failed & 0.5\%: 1.3 / 8.9 (interval [0.0, 28.8] crosses 20\%); 5\%: BadNets 1.3, Blended 44.5 (failed)\\
X9b.4 & layer-selected edit $\le15\%$ on CIFAR-100, SVHN, VGG & held & 0.2--8.8\% (VGG Blended interval [0.0, 35.7] crosses 15\%)\\
X9b.5 & the score picks mid for BadNets, pen for Blended ($\ge2$ of 3 seeds) everywhere & partly failed & 8 of 10 settings; not CIFAR-100 BadNets or SVHN Blended (both layers low there)\\
X9b.6 & SalUn $\le5\%$ at 100\%; its head-relearn $\ge50\%$ at 10\% & held & $\le0.3\%$; 86--100\%\\
X9b.7 & hidden signature: pre-edit gap $\ge30$ (BadNets); step-1 gap $\ge30$ (both triggers) & partly failed & pre-edit 88--96 (held); step 1 in 9 of 10, not CIFAR-100 BadNets (4.5 vs 0.1)\\
X9b.8 & R1 AUROC $\ge0.95$, ASR $\le10\%$ on BadNets everywhere; some trigger $>10\%$ at 5\% poisons & partly failed & BadNets held (0.2--1.2\%); at 5\%: 1.2 / 0.1\% (failed)\\
X9c.1 & in-sample $R^2\le t_Q/(1+t_Q)$ for every $k$ and seed & held & max $R^2$ minus bound $1.4\cdot10^{-26}$\\
X9c.2 & pre-edit $R^2\ge0.8$; bound within $1.5\times$ of $R^2$ pre-edit and for $k\le4$ & partly failed & 0.814, ratio 1.17 (held); $k\le4$: 1.70--2.06 (failed)\\
X9c.3 & loose bound $\ge0.3$ for $k\le8$; tight bound $\le0.05$ for every $k\ge1$ & partly failed & loose 0.50--0.70 (held); tight 0.25 / 0.22 / 0.07 at $k=1,2,4$\\
X9c.4 & held-out $R^2\le$ in-sample $+0.05$ for $k\ge1$ & held & at most $+0.004$\\
X9c.5 & proj$(k^\ast)$ forget $\le5\%$, retain within 1; head-relearn below the band, band $\ge70\%$; FT, SalUn above & partly failed & all held except band $\ge70\%$ (65.8\%)\\
X9c.6 & proj$(\delta)$: $R^2\le10^{-6}$ but forget accuracy $\ge20\%$ & partly failed (regraded) & $R^2\approx10^{-26}$ (held); forget accuracy 5.3\% (failed)\\
X9c.7 & $m\ge1{,}000$: $\|\delta\|^2$ within $2\times$ of $r/m$, $\gamma_1$ within 25\% of the MP edge; non-members $\|\delta_V\|^2\ge2\|\delta_F\|^2$ & partly failed & first two held (0.90--1.08; within 7\%); non-members 0.97--1.18$\times$ (failed)\\
X9c.8 & top-8 overlap $\le3\times8/r$ & held & 0.014--0.018 vs 0.016\\
X9c.9 & original loss-AUC $\ge0.55$ at every $m$; band in $[0.47, 0.53]$ & partly failed & band held; original 0.528 / 0.538 / 0.552\\
X9c.10 & proj$(k^\ast)$ per-example AUC within 0.05 of the original's ($m\ge1{,}000$), which is $\ge0.65$ & failed & 0.373 vs 0.611; 0.493 vs 0.686\\
X9c.11 & FT closer to 0.5 than proj at every $m$; band members in $[0.45, 0.55]$ & failed & only at $m=1{,}000$; members 0.41--0.59 at $m=100$\\
X10a.1 & exact on all 1,920 requests over 8 runs; final test predictions identical & held & 1,920/1,920, max $3.8\cdot10^{-12}$\\
X10a.2 & initial $\lambda^\star=1000$ on CIFAR-100 and Tiny-ImageNet & partly failed & 1000 / 300\\
X10a.3 & Select certificates on $L$ $\le10\%$; coarse check $\ge80\%$ on $U$ and $V$ in every run & partly failed & $L$ 0\% (held); 7 of 8 runs (one order: $U$ 67\%)\\
X10a.4 & worst-case augment certificate $<20\%$ of singles, 0\% of $L$ and $U$ & held & 9.1\% ($V$ only)\\
X10a.5 & relabelled fraction: C10 DINOv2 in $[0.10, 0.30]$, written $\ge0.40$, C100/Tiny $>0.20$ & held & 0.19--0.24; 0.45--0.87; 0.47 / 0.44\\
X10a.6 & flips per round: C10 DINOv2 $\le10$, written $\ge100$, C100/Tiny $\le50$ & partly failed & written held; one DINOv2 round 13.2; C100 97--161, Tiny 82--96\\
X10a.7 & latency: lean $\ge2\times$ cached on $U$; cached $\ge3\times$ rerun; lean $\ge5\times$ rerun on $L$, $U$; written lean/cached $<1.5\times$ & partly failed & first held; Tiny cached 2.9$\times$; lean on $L$ only $2.9$--$3.3\times$ on C100/Tiny; written 1.55 on one order\\
X10b.1 & exact deletion equals retraining to $\le10^{-9}$ & held & $\le5.5\cdot10^{-13}$\\
X10b.2 & U-LiRA original $\ge0.70$; exact deletion $0.50\pm0.02$ & held & 0.726 / 0.804 / 0.948; 0.500--0.502\\
X10b.3 & loss AUC $\ge0.99$ at $B=1$, $\ge0.90$ at $B=250$ & partly failed & $B=1$ held; $B=250$ CLIP 0.887\\
X10b.4 & margin AUC $\ge0.70$ and below the loss AUC everywhere & held & 0.790--0.981\\
X10b.5 & label-only AUC $\le0.53$, TPR at 0.1\% $\le0.02$ & partly failed & CIFAR-10 held; CIFAR-100 0.541, TPR 0.081\\
X10b.6 & CLIP below DINOv2-B in $B=250$ loss AUC and U-LiRA & held & 0.887 vs 0.948; 0.726 vs 0.804\\
X10b.7 & 95\% CI width $\le0.005$ for $B=250$ margin and loss AUC & held & 0.0003--0.0033\\
X12.1 & X2 written with the noise scale $s_0$: 240/240 exact, final test predictions identical & held & 240/240, max rel.\ $W$ diff $5.5\cdot10^{-12}$, 100\%\\
X12.2 & X10a written orders with $s_0$: 720/720 exact on certified, lean and cached paths & held & 720/720, $\le5.8\cdot10^{-12}$, predictions identical\\
X12.3 & X1 written with a data-free $\lambda$: within $10^{-9}$ of retraining after 1,000 deletions, identical predictions & held & $\le3.1\cdot10^{-11}$, 100\% at both $\rho$\\
X12.4 & $s_0$ accuracy within 3 points of $49.1$ / $45.4\%$ (before / after self-training) & held & $49.17$ / $45.50\%$\\
X12.5 & self-training lowers accuracy again & held & $49.17\to45.50\%$\\
X12.6 & initially selected $\lambda$ is an interior grid point & held & $\lambda^\star=100$\\
X12.7 & X1 written accuracy within 1 point of $65.9$ / $66.0\%$ & held & $65.87$ / $65.98\%$\\
X12.8 & sensitivity, $s=1$: 240/240 exact; accuracy within 3 points & held & 240/240; $49.67$ / $46.22\%$\\
\bottomrule
\end{longtable}}

\end{document}